\documentclass[twoside,11pt]{article}
\usepackage{blindtext}
\usepackage{jmlr2e}
\usepackage[utf8]{inputenc} 
\usepackage[T1]{fontenc}
\usepackage{url}            
\usepackage{booktabs}       
\usepackage{amsfonts}       
\usepackage{nicefrac}       
\usepackage{microtype}      
\usepackage{xcolor}         

\usepackage{enumerate}
\usepackage{wrapfig}
\usepackage{tikz}
\usetikzlibrary{positioning}
\usepackage{amsfonts,bbm,bbding,bm}
\usepackage{etoolbox}
\usepackage{color} 
\usepackage{enumitem}
\usepackage{dsfont}
\usepackage{tabularx}

\usepackage{amsmath,amsfonts,bbm,bbding,bm}
\usepackage{etoolbox}
\usepackage{subfigure}
\usepackage{xcolor}
\usepackage{booktabs} 
\usepackage{nicefrac}       
\usepackage{microtype} 
\usepackage{multirow}
\usepackage{mathtools}
\usepackage{color} 
\usepackage{enumitem}
\usepackage{dsfont}
\usepackage{tikz}

\newcommand{\Ball}{\mathbb{B}}
\newcommand{\PP}{\mathbb{P}}
\newcommand{\R}{\mathbb{R}}
\newcommand{\F}{\mathcal{F}}
\newcommand{\D}{\mathcal{D}}
\newcommand{\indi}{\mathds{1}}
\newtheorem{problem}{Problem}

\usepackage{lastpage}
\jmlrheading{27}{2026}{1-\pageref{LastPage}}{4/23; Revised
5/26}{6/26}{23-0511}{Jiancong Xiao, Yanbo Fan, Ruoyu Sun, and Zhi-Quan Luo}

\ShortHeadings{Adversarial Rademacher Complexity of Deep Neural Networks}{Xiao, Fan, Sun, and Luo}
\firstpageno{1}

\begin{document}  
\title{Adversarial Rademacher Complexity of Deep Neural Networks}

\author{\name Jiancong Xiao\email jiancongxiao@link.cuhk.edu.cn \\
       \addr
       The Chinese University of Hong Kong, Shenzhen\\
        Shenzhen, China
       \AND
       \name Yanbo Fan \email yanbofan@nju.edu.cn \\
       \addr
       Nanjing University\\
       Suzhou, China
       \AND
       \name Ruoyu Sun\thanks{Corresponding Author.} \email sunruoyu@cuhk.edu.cn \\
       \addr
       The Chinese University of Hong Kong, Shenzhen\\
        Shenzhen, China
       \AND
       \name Zhi-Quan Luo \email luozq@cuhk.edu.cn \\
       \addr
       The Chinese University of Hong Kong, Shenzhen\\
        Shenzhen, China
       }

\editor{Po-Ling Loh}
\maketitle

\begin{abstract}
Deep neural networks (DNNs) are highly vulnerable to adversarial attacks. Ideally, a robust model should perform well on both perturbed training data and unseen perturbed test data. While DNNs can fit perturbed training data, generalizing to perturbed test data remains a significant challenge. This motivates the study of generalization guarantees from a learning theory perspective. This paper focuses on adversarial Rademacher complexity (ARC), first introduced by \citet{khim2018adversarial} and \citet{yin2018rademacher}. Their work primarily addressed linear functions and highlighted the open question of how to bound ARC for neural networks. Since then, several attempts have been made, with the latest results applying ARC only to two-layer neural networks. The main challenge arises from the dynamic nature and unknown closed-form solution of adversarial examples. In this paper, we resolve this issue and provide the first bound on ARC for deep neural networks. Our bound is qualitatively comparable to Rademacher complexity bounds in similar settings. The key ingredient is a new concept we introduce, termed intermediate adversarial examples, along with a framework for calculating the covering number that is compatible with them. Finally, we present experiments to analyze poor robust generalization, demonstrating that the weight norm is a crucial factor influencing the robust generalization gap.
\end{abstract}

\begin{keywords}
  Rademacher Complexity, Adversarial Robustness, Generalization Bounds, Neural Networks
\end{keywords}

\section{Introduction}
\label{Intro}
Deep neural networks (DNNs) \citep{krizhevsky2012imagenet, hochreiter1997long} have achieved remarkable success in various machine learning tasks, including computer vision (CV) and natural language processing (NLP). However, they have been shown to be vulnerable to adversarial examples \citep{szegedy2013intriguing, goodfellow2014explaining}. More specifically, a well-trained model can perform poorly on slightly perturbed data samples. Incorporating perturbed samples into the training dataset can improve robustness in practice, but it does not always lead to satisfactory performance. One major issue arises from generalization: while training a model to fit perturbed training samples is relatively easy, such a model often fails to generalize well to adversarial examples in the test set. For instance, when applying ResNet to CIFAR-10, adversarial training can achieve nearly 100\% robust accuracy on the training set, yet only 47\% robust accuracy on the test set \citep{madry2017towards}. Recent works \citep{gowal2020uncovering, rebuffi2021fixing} have mitigated the overfitting issue, but it still has a $20\%$ robust generalization gap between robust test accuracy (approximately 60\%) and robust training accuracy (around 80\%). Therefore, it is interesting to provide a theoretical understanding of adversarially robust generalization. This paper focuses on Rademacher complexity.

In classical learning theory, the generalization gap can be bounded in terms of Rademacher complexity with high probability. Rademacher complexity is defined as\newline\begin{equation}\mathcal{R}_\mathcal{S}(\mathcal{H})=\mathbb{E}_\sigma \frac{1}{n}\bigg[\sup_{h\in\mathcal{H}}\sum_{i=1}^{n}\sigma_i h(x_i,y_i)\bigg],\end{equation}\newline where $\mathcal{S}=\{x_i,y_i\}_{i=1,\cdots,n}$ is the sample dataset with $n$ samples, $\mathcal{H}$ is the hypothesis function class, and $\sigma_i$ are i.i.d. Rademacher random variables, \emph{i.e.,} $\sigma_i$ takes values $1$ or $-1$ with equal probability. Techniques for deriving upper bounds on the Rademacher complexity of deep neural networks have been extensively studied, including layer-peeling \citep{neyshabur2015norm, golowich2018size} and covering number arguments \citep{bartlett2017spectrally}. For more details, see Section \ref{rel}.

\citet{khim2018adversarial} and \citet{yin2018rademacher} concurrently extended Rademacher complexity to adversarial settings. They demonstrated that the robust generalization gap can be bounded by the Rademacher complexity of the adversarial loss, defined as \(\tilde{h}(x_i, y_i) = \max_{x_i' \in \mathcal{B}(x_i)} h(x_i', y_i)\), where \(\mathcal{B}(x)\) is a norm ball around sample \(x\), and \(\tilde{\mathcal{H}}\) represents the hypothesis class of adversarial losses. This specific form of Rademacher complexity is referred to as adversarial Rademacher complexity. Their primary contribution was establishing bounds for linear function classes.

For neural networks, it may seem straightforward to extend methods for standard losses to adversarial losses. However, \citet{khim2018adversarial, yin2018rademacher} both pointed out that providing upper bounds in adversarial settings is significantly more challenging due to the presence of the $\max$ operator in adversarial loss. As a result, they relied on surrogate losses, leaving the following question for future work: \begin{center} \emph{How can the adversarial Rademacher complexity of deep neural networks be bounded?} \end{center} Since 2018, several attempts have been made to tackle this problem. \citet{awasthi2020adversarial} attempted to extend the bounds on adversarial Rademacher complexity from linear functions to two-layer neural networks. \citet{gao2021theoretical} proposed a more meaningful surrogate loss: the FGSM loss. Broadly, existing attempts to tackle this problem can be categorized into two main approaches.

\begin{table}[t]\small
    \caption{Comparison of our work with the two types of attempts on bounding Adversarial Rademacher complexity:
Type 1: Adversarial Loss in Shallow Networks \citep{yin2018rademacher,khim2018adversarial,awasthi2020adversarial}.
Type 2: Surrogate Loss in Deep Networks \citep{yin2018rademacher,khim2018adversarial,gao2021theoretical}. Our work distinguishes itself by providing the first bound for the Adversarial Rademacher Complexity of DNNs.}
    \centering
    \begin{tabular}{l l l p{2.7cm} p{4.1cm}}
    \toprule
    &Loss & Networks & Techniques & Limitation\\
    \midrule
Type 1 & \textbf{Adversarial Loss} &$\leq$ Two-Layer & Optimal Attack & Cannot be applied to DNNs\\
\multirow{2}{*}{Type 2}& \multirow{2}{*}{Surrogate Loss} & \multirow{2}{*}{\textbf{Multi-Layer}} & \multirow{2}{*}{Change Definition} & Cannot bound the robust generalization gap\\
  Ours &\textbf{Adversarial Loss} &\textbf{Multi-Layer} & Lemmas \ref{lemma:iae} \& \ref{lemma:induction} & - \\
  \bottomrule
    \end{tabular}
    \label{tab:my_label}
\end{table}
\paragraph{Type 1: Adversarial Loss in Shallow Networks.} The first approach focuses on obtaining closed-form solutions for the optimal adversarial examples $x^*$ and analyzing the adversarial loss, given by $\max_{x'} h(x',y) = h(x^*,y)$. \citet{khim2018adversarial, yin2018rademacher} introduced adversarial Rademacher complexity and provided bounds for linear functions using this method. \citet{awasthi2020adversarial} extended this analysis to two-layer neural networks, deriving bounds in this setting. However, in deeper networks, obtaining closed-form solutions becomes intractable, making it unclear how to extend this approach to multi-layer architectures and derive generalization bounds in deep neural networks.

\paragraph{Type 2: Surrogate Loss in Deep Networks.} This approach uses a surrogate loss $\hat{h}(x,y) \approx \tilde{h}(x,y) = \max_{x'} h(x',y)$ to bypass the main difficulty posed by the $\max$ operator, where the surrogate loss does not explicitly contain a $\max$ term. Examples of $\hat{h}(x,y)$ include tree-transformation loss \citep{khim2018adversarial}, SDP relaxation loss \citep{yin2018rademacher}, and FGSM loss \citep{gao2021theoretical}. However, this approach provides upper bounds for the Rademacher complexity of the surrogate loss rather than the adversarial loss, and thus cannot bound the robust generalization gap. For a more detailed discussion, see Appendix \ref{B2}.

In summary, these two types of attempts aim to eliminate the $\max$ operator using different approaches. The methods and their limitations are summarized in Table \ref{tab:my_label}. To our knowledge, the problem of bounding the adversarial Rademacher complexity of deep neural networks has remained unsolved since it was first raised in 2018. In this paper, we resolve this problem and provide the first bound for the adversarial Rademacher complexity of deep neural networks. Our approach is based on the covering number, which serves as an upper bound for Rademacher complexity. In adversarial settings, this problem becomes:

\begin{center}
    \emph{How to calculate the covering number of the adversarial hypothesis class?}
\end{center} 
The first challenge for this problem is that the closed-form expression of optimal adversarial examples is not known. To address this, we introduce a concept called intermediate adversarial examples, which allow us to bound the covering number of the linear function class without requiring access to the closed-form solution of the optimal adversarial example. Using this approach, we reproduce the bound in the linear setting. The formal definition of intermediate adversarial examples is provided later in Lemma~\ref{lemma:iae}.

The second challenge arises from the conflict between existing methods for calculating the covering number of DNNs and the dynamic nature of intermediate adversarial examples. Current techniques for computing the covering number of DNNs assume a static training set, whereas adversarial examples are model-dependent and evolve dynamically. To resolve this issue, we introduce a lemma called Layer-wise Induction for Adversarial Hypothesis Class, which is designed to be compatible with our intermediate adversarial examples. The formal statement of this lemma is provided later in Lemma~\ref{lemma:induction}. By combining these two techniques, we establish the first bound on the adversarial Rademacher complexity of DNNs.

\paragraph{Main Result.} For depth-$l$, width-$h$ fully-connected neural networks, assume that the weight matrices $W_1,W_2,\dots,W_l$ in each of the $l$ layers have Frobenius norms bounded by $M_1,\dots,M_l$, and all $n$ samples are bounded by $B$. Then, \[\text{Adversarial Rademacher Complexity} \leq\mathcal{O}\bigg(\frac{(B+\epsilon)h\sqrt{l\log l}\prod_{j=1}^l M_j}{\sqrt{n}}\bigg).\] We provide a comparison with existing bounds in similar settings. We show that our bound is comparable to (1) the upper bound for standard Rademacher complexity and (2) the upper bound for adversarial Rademacher complexity of two-layer neural networks \citep{awasthi2020adversarial}. Additionally, we provide a lower bound for adversarial Rademacher complexity and extend the results to multi-class classification settings. Finally, we study some empirical implications of our bounds. Our experiments indicate that the weight norm is positively correlated with the robust generalization gap. These findings contribute to a deeper theoretical understanding of adversarial robustness in deep learning models.

\section{Related Work}  
\label{rel}  
\paragraph{Adversarial Attacks and Defense.} Since 2013, it has been well established that deep neural networks trained using standard gradient descent are highly vulnerable to small perturbations in input data \citep{szegedy2013intriguing, goodfellow2014explaining, chen2017zoo, carlini2017towards, madry2017towards}. Research on improving the robustness of neural networks has followed two main directions. One line of work focuses on developing defense mechanisms to enhance model robustness against adversarial attacks \citep{wu2020adversarial, gowal2020uncovering}. Another line aims to design stronger adversarial attacks to evaluate and challenge existing defenses \citep{athalye2018obfuscated, tramer2020adaptive, chen2017zoo, xiao2022understanding}.
\paragraph{Robust Generalization.} Prior research has demonstrated that increasing the amount of training data can improve robust generalization \citep{schmidt2018adversarially, raghunathan2019adversarial, zhai2019adversarially}. Several works have analyzed generalization in adversarial settings through the lens of VC-dimension \citep{attias2021improved, montasser2019vc}. \citet{neyshabur2017pac} applied a PAC-Bayesian framework to derive generalization bounds for neural networks, which was later extended to adversarial settings by \citet{farniageneralizable, xiao2023pacbayesian}. \citet{sinha2017certifiable} examined robust generalization in the context of distributional robustness, while \citet{allen2020feature} explored it from the perspective of feature purification. Additionally, \citet{javanmard2020precise} studied generalization properties in the setting of linear regression.
\paragraph{Rademacher Complexity.} \citet{neyshabur2015norm} applied a layer-peeling technique to derive a generalization bound for depth-$l$ neural networks. Specifically, assuming that the Frobenius norms of the weight matrices $W_1, W_2, \dots, W_l$ are bounded by $M_1, \dots, M_l$, and that all $n$ input instances have $\ell_2$-norm bounded by $B$, they showed that the generalization gap between the population risk and the empirical risk is bounded with high probability by $\mathcal{O}(B 2^l \prod_{j=1}^l M_j / \sqrt{n})$. Additionally, \citet{bartlett2017spectrally} provided a spectral norm-based bound on the Rademacher complexity by controlling the covering number of the function class of deep neural networks. The relevant work on Adversarial Rademacher Complexity is discussed in the Introduction, with further details provided in Appendix \ref{appendix:B}. For a more detailed discussion, see Appendix \ref{B3}.
\section{Preliminaries}
\subsection{Generalization Gap and Rademacher Complexity}
\paragraph{Generalization Gap.} In the classical machine learning framework, we consider a function class $\mathcal{F}$ (e.g., linear functions, neural networks). The learning objective is to find a function $f \in \mathcal{F}$ that minimizes the population risk:
\[
R(f)=\mathbb{E}_{(x,y)\sim\mathcal{D}}[\ell(f(x),y)],
\]
where $\mathcal{D}$ denotes the underlying data distribution and $\ell(\cdot)$ is the loss function.
Since $\mathcal{D}$ is typically unknown, we minimize the empirical risk in practice. Given $n$ independent and identically distributed (i.i.d.) samples $\mathcal{S}=\left\{(x_1,y_1),\ldots,(x_n,y_n)\right\}$, the empirical risk is defined as:
\[
R_n(f)=\frac{1}{n}\sum_{i=1}^n\ell(f(x_i),y_i).
\]
The generalization gap is then defined as the difference between the population risk and the empirical risk:
\[
\text{Generalization Gap} := R(f) - R_n(f).
\]
Let the hypothesis class be defined as $\mathcal{H}=\left\{h\mid h(x,y)=\ell(f(x),y), f\in\mathcal{F}\right\}$, which connects the loss function to the function class. The Rademacher complexity framework leads to the following generalization bound.
\begin{proposition}[\citet{bartlett2002rademacher}]
    \label{thm:rad_vanilla}
Let the loss function $\ell(f(x),y)$ be bounded with range $[0,C]$. Then for any $\delta \in (0,1)$, with probability at least $1-\delta$, the following inequality holds for all $f \in \mathcal{F}$:
	\begin{equation*}
	R(f)\leq R_n(f)+2\mathcal{R}_\mathcal{S}(\mathcal{H})+3C\sqrt{\frac{\log\frac{2}{\delta}}{2n}}.
	\end{equation*}
\end{proposition}

\subsection{Robust Generalization Gap and Adversarial Rademacher Complexity}
\paragraph{Robust Generalization Gap.} In the context of adversarial robustness, we define the robust population risk and robust empirical risk as follows:
\[
\Tilde{R}(f)=\mathbb{E}_{(x,y)\sim \mathcal{D}}\max_{\|x'-x\|_p\leq\epsilon}\ell(f(x'),y)\quad \text{and}\ \quad
\Tilde{R}_n(f)=\frac{1}{n}\sum_{i=1}^n\max_{\|x_i'-x_i\|_p\leq\epsilon}\ell(f(x_i'),y_i).
\]
Throughout this paper, we focus on general $\ell_p$ attacks where $p\geq 1$. We denote by $\mathcal{B}(x)$ the general perturbation set around point $x$. For $\ell_p$ attacks, this set is defined as $\mathcal{B}(x)=\left\{x'\mid \|x'-x\|_p\leq\epsilon\right\}$. The robust generalization gap is then defined as:
\[
\text{Robust Generalization Gap} := \Tilde{R}(f)-\Tilde{R}_n(f).
\]
Let the adversarial loss be defined as $\Tilde{\ell}(f(x),y):=\max_{x'\in\mathcal{B}(x)}\ell(f(x'),y)$. We then define the adversarial hypothesis class as:
\begin{equation}
\label{eq:adv-hypo-class}
\Tilde{\mathcal{H}}=\left\{\Tilde{h}: \Tilde{h}(x,y)=\Tilde{\ell}(f(x),y), f\in\mathcal{F}\right\},
\end{equation}
and assume $0\in \Tilde{\mathcal{H}}$. Then, according to Proposition \ref{thm:rad_vanilla}, the robust generalization gap can be bounded by the Rademacher complexity of $\Tilde{\mathcal{H}}$. We have the following robust generalization bound.
\begin{proposition}
[\citet{yin2018rademacher}] 
	\label{prop2} 
 Let the loss function $\ell(f(x),y)$ be bounded with range $[0,C]$. Then for any $\delta \in (0,1)$, with probability at least $1-\delta$, the following inequality holds for all $f \in \mathcal{F}$:
	\begin{equation*}
	\Tilde{R}(f)\leq \Tilde{R}_n(f)+2\mathcal{R}_\mathcal{S}(\Tilde{\mathcal{H}})+3C\sqrt{\frac{\log\frac{2}{\delta}}{2n}}.
	\end{equation*}
\end{proposition}

\begin{definition}[Adversarial Rademacher Complexity] Following Proposition \ref{prop2}, we define the Adversarial Rademacher Complexity (ARC) as the Rademacher complexity of the adversarial hypothesis class $\tilde{\mathcal{H}}$:
\begin{equation*}
\mathcal{R}_\mathcal{S}(\tilde{\mathcal{H}})=\mathbb{E}_\sigma \frac{1}{n}\bigg[\sup_{\Tilde{h}\in\Tilde{\mathcal{H}}}\sum_{i=1}^{n}\sigma_i \Tilde{h}(x_i',y_i)\bigg]=\mathbb{E}_\sigma \frac{1}{n}\bigg[\sup_{h\in\mathcal{H}}\sum_{i=1}^{n}\sigma_i \max_{x_i'\in\mathcal{B}(x_i)}h(x_i',y_i)\bigg],
\end{equation*}
\end{definition}
The Rademacher complexity can be further upper bounded using the covering number, which we define as follows.
\begin{definition}[$\zeta$-cover]
\label{def:cov}
Let $\zeta>0$ and $(\mathcal{V},d(\cdot,\cdot))$ be a metric space, where $d(\cdot,\cdot)$ is a (pseudo)-metric.
A subset $\mathcal{C}\subset \mathcal{V}$ is called a $\zeta$-cover of $\mathcal{V}$ if for any $v\in \mathcal{V}$, there exists $v'\in\mathcal{C}$ such that $d(v,v')\leq\zeta$. The $\zeta$-covering number of $\mathcal{V}$, denoted as $\mathcal{N}(\mathcal{V},d(\cdot,\cdot),\zeta)$, is defined as the minimum cardinality $|\mathcal{C}|$ over all possible $\zeta$-covers.
\end{definition}

We now specialize this concept to hypothesis classes. Given a sample dataset $\mathcal{S}$, we define a pseudometric on $\mathcal{H}$ as: $\|h\|_\mathcal{S}^2=\frac{1}{n}\sum_{i=1}^{n}h( x_i,y_i)^2$. The $\zeta$-covering number of $\mathcal{H}$ is then defined as $\mathcal{N}(\mathcal{H},\|\cdot\|_\mathcal{S},\zeta)$. 

\paragraph{Function Class.} We consider depth-$l$, width-$h$ fully-connected neural networks,
\begin{equation}  
\label{hypo}  
    \mathcal{F} = \left\{ x \mapsto W_l \rho \left( W_{l-1} \rho \left( \cdots \rho \left( W_1 x \right) \cdots \right) \right) \mid \|W_j\| \leq M_j, \ j = 1, \dots, l \right\},
\end{equation}
where $\rho (\cdot)$ is an element-wise $L_\rho$-Lipschitz activation function and $\rho (0)=0$, $W_j$ are $h_j\times h_{j-1}$ matrices, for $j=1,\cdots,l$. $h_0$ equals to the input dimension $d$. Let $h=\max\left\{h_0,\cdots,h_l\right\}$ be the width of the neural networks. Denote the ($a,b$)-group norm $\|W\|_{a,b}$ as the $a$-norm of the $b$-norm of the rows of $W$. We consider two cases in Equation~\eqref{hypo}: the Frobenius norm and the \(\|\cdot\|_{1,\infty}\)-norm. We use the superscript $\mathrm{binary}$ to indicate the binary setting. 
In this case, functions $f\in\mathcal{F}^\mathrm{binary}$ have scalar outputs, i.e., $h_l=1$. 
Without this superscript, the notation refers to the general multiclass setting, where $f$ outputs a vector. The corresponding function classes are denoted as \(\mathcal{F}_2\) and \(\mathcal{F}_{1,\infty}\), respectively. Additionally, let the training data be $x_1,\cdots, x_n \in \mathbb{R}^d$. We assume that $\|X\|_{p,\infty}= B$, where $X$ is the data matrix whose $i$-th row is $x_i^\top$.

\section{Main Challenges in Bounding ARC}
\label{sec:difficulty}
In this section, we discuss the fundamental challenges encountered in bounding ARC and present our approach to addressing these challenges.

The main difficulty in bounding the ARC of DNNs comes from the fact that adversarial perturbations destroy
the recursive structure that underlies the existing analyses of Rademacher
complexity for DNNs.

To explain this point, recall that the existing approaches for bounding the
Rademacher complexity of deep neural networks mainly follow two related
routes. The first route directly peels the Rademacher complexity layer by layer.
Let \(\mathcal F_l\) denote the \(l\)-layer network class, this approach establishes a
recursive relationship of the form
\[
    \mathcal R_\mathcal{S}(\mathcal H_l)
    \quad \text{in terms of} \quad
    \mathcal R_\mathcal{S}(\mathcal H_{l-1}),
\]
where the last layer is peeled off and the remaining part is treated as an
\((l-1)\)-layer network class. The second route bounds the Rademacher
complexity through covering numbers. In this case, the key step is again
recursive: one constructs the covering number of the \(l\)-layer class, denoted as $\mathcal N_l$, from
the covering numbers of the previous layers, schematically,
\[
    \mathcal N_l
    \quad \text{in terms of} \quad
    \mathcal N_1,\ldots,\mathcal N_{l-1}.
\]
Although these two routes look different, both of them rely on the same
implicit fixed-sample structure. At each recursive step, the input feature set
to the current layer is treated as fixed once the previous layers have been
handled.

More concretely, consider an \(l\)-layer neural network and write
\[
    h_0(x)=x,
    \qquad
    h_j(x)=\sigma(W_jh_{j-1}(x)),
    \quad j=1,\ldots,l-1.
\]
In the standard, non-adversarial setting, the empirical process is evaluated
on the fixed samples \(\{x_i\}_{i=1}^n\). Thus, at the \(j\)-th layer, the
relevant feature set is
\[
    \{h_{j-1}(x_i):i=1,\ldots,n\}.
\]
This feature set is generated from the fixed training samples and the previous
layers. Hence, after the previous layers have been peeled or covered, the
analysis of the next layer can be performed conditionally on this empirical
feature set. This is the fixed-sample structure implicitly used by both the
direct Rademacher peeling method and the covering number method.

The adversarial setting breaks this structure. In ARC, the relevant input for \(x_i\) is selected by an inner maximization,
for example
\[
    x_i^*(f)
    \in
    \arg\max_{\|x-x_i\|\le \epsilon}
    \ell(f(x),y_i).
\]
Thus the effective sample is no longer the fixed sample
\(\{x_i\}_{i=1}^n\), but the network-dependent sample
\[
    \{x_i^*(f):i=1,\ldots,n\}.
\]
This network dependence prevents the standard recursive arguments from going
through.

For the direct Rademacher peeling route, the obstruction is the following.
A recursion from an \((l-1)\)-layer ARC to an
\(l\)-layer ARC would require comparing the
adversarial process of the \((l-1)\)-layer subnetwork with that of the
\(l\)-layer network. However, the adversarial examples associated with these
two objects are generally different:
\[
    x_i^*(W_{l-1},\cdots, W_1)
    \neq
    x_i^*(W_l,W_{l-1},\cdots, W_1).
\]
The former is selected according to an \((l-1)\)-layer network, while the
latter is selected according to the full \(l\)-layer network. Therefore, the
recursive step
\[
    \mathcal R_\mathcal{S}(\tilde{\mathcal H}_{l-1})
    \rightarrow
    \mathcal R_\mathcal{S}(\tilde{\mathcal H}_l)
\]
does not have the same fixed samples on both sides. The sample on
which the \((l-1)\)-layer quantity is evaluated is not the sample that appears
inside the \(l\)-layer adversarial process. This mismatch is precisely what
prevents a direct layer-by-layer peeling argument.

For the covering number route, the obstruction appears in two places: one is
the comparison between two nearby networks, and the other is again the
recursive step. Let \(f\) and \(f'\) be two networks.

First, in the standard setting, one controls their empirical discrepancy on
the same fixed samples:
\[
    \bigl|f(x_i)-f'(x_i)\bigr|.
\]
In the adversarial setting, however, the two adversarial examples
\[
    x_i^*(f)=x_i^*(W_l,\cdots,W_1)
    \ \text{and}\
    x_i^*(f')=x_i^*(W_l',\cdots,W_1')
\]
are different in general. Therefore, the relevant comparison is no longer
between two networks evaluated on the same input, but between two networks
evaluated on different network-dependent adversarial examples.

Second, in the standard setting, this common sample set allows the
covering number to be calculated recursively across layers. In the
adversarial setting, however, the covering numbers of different layers are
evaluated on different adversarial examples, since the adversarial examples
depend on the network being covered. Consequently, similar to the direct
Rademacher-complexity route, the usual layerwise covering recursion
\[
    \mathcal N_1,\ldots,\mathcal N_{l-1}
    \longrightarrow
    \mathcal N_l
\]
does not directly go through.

This motivates us to construct an adversarial example that can play the role of a common reference point in the recursive analysis. The goal is to define layerwise adversarial examples
that are not fixed in the same sense as the original samples in the standard
setting, but are sufficiently stable and structured to support the desired
recursions. To this end, we introduce intermediate adversarial examples. They serve as replacements for the fixed intermediate feature sets in the
non-adversarial theory. They allow us to track adversarially selected inputs
through the network layer by layer and thereby recover a recursive structure
for the adversarial Rademacher complexity analysis.

\subsection{Intermediate Adversarial Examples}
\label{sec:challenge1}
In this section, we use binary classification with a one-dimensional linear function as a simple example to illustrate how intermediate adversarial examples are defined and used to bound the ARC. Following \citet{yin2018rademacher} and \citet{ awasthi2020adversarial}, we define the loss function as \(\ell(f(x), y) = \phi(y f(x))\), where \(\phi\) is a non-increasing function.
Then 
\[
\max_{x'} \ell \left( f \left( x' \right), y \right) = \phi \left( \min_{x'} y f \left( x' \right) \right).
\]
Assume that the function $\phi$ is $L_\phi$-Lipschitz, by Talagrand’s
Lemma \citep{ledoux2013probability}, we have $\mathcal{R}_\mathcal{S}(\Tilde{\mathcal{H}})\leq L_\phi \mathcal{R}_\mathcal{S}(\tilde{\mathcal{F}}^\mathrm{binary})$, where we define the adversarial hypothesis class as
\begin{equation}  
\label{hypo2}  
    \tilde{\mathcal{F}}^\mathrm{binary} = \left\{ \tilde{f} : (x,y) \mapsto \inf_{\left\| x - x' \right\|_p \leq \epsilon} y f \left( x' \right) \mid f \in \mathcal{F}^\mathrm{binary} \right\}.  
\end{equation}
Let \(f_w(x): \mathcal{X} \rightarrow \mathbb{R}\) with \(|w| \le M\) and $\mathcal{X}\subset\mathbb{R}$, and define its adversarial function as  
\[
g_w(x) \triangleq \min_{x' \in [x - \epsilon,\, x + \epsilon]} f_w(x').
\]  
For simplicity, suppose we have only one sample \(x\) with \(|x| = B\). The problem is as follows:

\begin{problem}
\label{p1}
Bound the size of an $\zeta$-cover $\mathcal{N}(\tilde{\mathcal{F}}^\mathrm{binary},\|\cdot\|_\mathcal{S},\zeta)$ of the adversarial hypothesis class
$\tilde{\mathcal{F}}^\mathrm{binary} = \left\{ g(x) \triangleq \min_{x' \in [x - \epsilon, x + \epsilon]} f_w(\cdot) : |w| \le M \right\}$ given sample $x$.
Here, the $\zeta$-cover is a set of functions
whose distance to any function in $\tilde{\mathcal{F}}^\mathrm{binary}$ is no more than $\zeta$.
\end{problem}
To help better understand Problem \ref{p1},
we consider a simpler problem of a standard function class
$\mathcal{F}^\mathrm{binary} = \left\{ f_w(\cdot): \mathbb{R} \rightarrow \mathbb{R} \mid |w| \le M \right\}$.

\begin{problem}
\label{p2}
Bound the size of an $\zeta$-cover $\mathcal{N}(\mathcal{F}^\mathrm{binary},\|\cdot\|_\mathcal{S},\zeta)$ of the standard function class $\mathcal{F}^\mathrm{binary} = \left\{ f_w(\cdot): \mathbb{R} \rightarrow \mathbb{R} \mid |w| \le M \right\}$ given sample $x$.
\end{problem}

The idea is to relate a cover of the function class to a cover of the
parameter region. Consider the one-dimensional linear function $f_w(x)=wx, w\in[-M,M].$ For two parameters \(w_1,w_2\in[-M,M]\), we have
\[
    |f_{w_1}(x)-f_{w_2}(x)|
    =
    |(w_1-w_2)x|
    \leq
    |w_1-w_2|\,|x|.
\]
Therefore, if \(|x|\leq B\) and $|w_1-w_2|\leq \epsilon_w:=\frac{\zeta}{B}$, then $|f_{w_1}(x)-f_{w_2}(x)|\leq \zeta$.
Thus, an \(\epsilon_w\)-cover of the parameter interval \([-M,M]\) induces a
\(\zeta\)-cover of the function class \(\mathcal F^\mathrm{binary}\) on the domain
\(\{x:|x|\leq B\}\). It remains to bound the size of such a parameter cover. For example,
\[
    \left\{M, M-2\epsilon_w, M-4\epsilon_w,\ldots\right\}
\]
is a \(2\epsilon_w\)-spaced grid over \([-M,M]\), whose size is at most $\frac{2M}{2\epsilon_w}
    =
    \frac{M}{\epsilon_w}
    =
    \frac{MB}{\zeta}$. Hence, for the one-dimensional linear class, the covering number satisfies
\[
    \mathcal N(\mathcal F^\mathrm{binary},\zeta)
    \leq
    \frac{MB}{\zeta}.
\]

Now we return to Problem \ref{p1} and show how to define and use intermediate adversarial examples to solve this problem. Given an original example $x$, let 
\[
x_1^*=\arg\inf_{\|x-x'\|\leq\epsilon} w_1 x',\quad\text{and}\quad x_2^*=\arg\inf_{\|x-x'\|\leq\epsilon} w_2x'.
\]
The intended idea is that $x_1^*$ and $x_2^*$ are two different adversarial examples derived from the same original sample $x$, constructed against $w_1$ and $w_2$, respectively. Let
\[
\bar{x}=\begin{cases}
     x_2^* & if\quad w_1 x_1^*\geq w_2 x_2^* \\
     x_1^* & if \quad w_1 x_1^*<w_2 x_2^* .
\end{cases}.
\]
If $w_1 x_1^*\geq w_2 x_2^*$, we have $w_1x_1^*-w_2x_2^*\leq w_1x_2^*-w_2x_2^*=w_1 \bar{x}-w_2\bar{x}$. If $w_1 x_1^*< w_2 x_2^*$, we have $w_2x_2^*-w_1x_1^*\leq w_2x_1^*-w_1x_1^*=w_2 \bar{x}-w_1\bar{x}$. Combine these two inequalities, we have
\begin{equation}  
    \label{eq:linear_perturb}  
    \left| w_1 x_1^* - w_2 x_2^* \right| \leq \left| w_1 \bar{x} - w_2 \bar{x} \right| \leq \left| w_1 - w_2 \right| \left| \bar{x} \right| \leq \epsilon_w \left( B + \epsilon \right).  
\end{equation}
The choice of \(\bar{x}\) in the two-dimensional case is illustrated in Figure~\ref{fig:iae}. Consider the original sample \(x = (0,0)\). Given \(w_1\) and \(w_2\), the points \(x^*_1\) and \(x^*_2\) lie on the boundary of the \(\epsilon\)-ball in the directions opposite to \(w_1\) and \(w_2\), respectively. If we assume that the magnitude of \(w_1\) is larger than that of \(w_2\), then \(w_1^\intercal x^*_1 < w_2^\intercal x^*_2\) (In the two-dimensional case, \(wx\) becomes the inner product between \(w\) and \(x\)). Accordingly, we set \(\bar{x} = x^*_1\).

\begin{figure}[ht]
\centering
\begin{tikzpicture}[scale=2.0]
\node[fill,circle,inner sep=1pt,label=above:$x$] (X) at (0,0) {};
\draw[thick] (0,0) circle (1.0);
\draw[->,thick] (0,0) -- (1.4,0.5) node[midway,above] {$w_1$};
\draw[->,thick] (0,0) -- (0.4,1.2) node[midway,above] {$w_2$};
\node[fill,circle,inner sep=1pt,label=right:$x_1^*$] (X1) at (-0.9421,-0.336) {};

\node[fill,circle,inner sep=1pt,label=right:$x_2^*$] (X2) at (-0.3614,-0.949) {};

\node[fill,circle,inner sep=1pt,label=above:$\bar{x}$] (Xbar) at (-0.9421,-0.336) {};

\draw[dashed] (0,0) -- (-0.9421,-0.336);

\draw[dashed] (0,0) -- (-0.3614,-0.949);

\node at (0.8,0.0) {$\epsilon$};

\end{tikzpicture}
\caption{A schematic illustration of the choice of \(\bar{x}\). The points \(x_1^*\) and \(x_2^*\) within the ball \(\|x' - x\|\le \epsilon\) are the minimizers of \(w_1^\intercal x'\) and \(w_2^\intercal x'\), respectively. The point \(\bar{x}\) is chosen as either \(x_1^*\) or \(x_2^*\), depending on which of \(w_1^\intercal x'\) and \(w_2^\intercal x'\) is smaller.
} 
\label{fig:iae}
\end{figure}
Therefore, the $\zeta$-cover of $\tilde{\mathcal{F}}^\mathrm{binary}$
  is no more than   $ M/\epsilon_w  = M(B+\epsilon) / \zeta $.

\paragraph{Remark 1.} This approach recovers the upper bound of ARC for linear functions, as established in \citet{khim2018adversarial,yin2018rademacher}. The main advantage of this approach is that it provides a bridge for computing the distance between two adversarial functions without requiring the closed-form solution of adversarial examples. Consequently, this method shows potential for extension to multi-layer neural networks. However, we will demonstrate an additional challenge prevents the direct application of this approach to multi-layer neural networks in the following subsection. Nevertheless, our proof reveals that the definition of $\bar{x}$ is a crucial step. We refer to this as the \emph{intermediate adversarial example} and present the corresponding lemma for general functions below.

\begin{lemma}[Intermediate Adversarial Example]
\label{lemma:iae} Given $(x,y)$ and perturbation set $\mathcal{B}(x)$. Suppose $\mathcal{B}(x)$ is compact. For all $\tilde{h}_1,\tilde{h}_2\in\tilde{\mathcal{H}}$ with their standard counterparts $h_1,h_2\in\mathcal{H}$, there exists an adversarial example $x'(\tilde{h}_1,\tilde{h}_2)\in \mathcal{B}(x)$, s.t.
\begin{equation*}  
    \left| \tilde{h}_1(x,y) - \tilde{h}_2(x,y) \right|  
    \leq \left| h_1 \left( x' \left( \tilde{h}_1, \tilde{h}_2 \right), y \right)  
    - h_2 \left( x' \left( \tilde{h}_1, \tilde{h}_2 \right), y \right) \right|.  
\end{equation*}
We refer to this adversarial example $x'(\tilde{h}_1,\tilde{h}_2)\in \mathcal{B}(x)$ as intermediate adversarial example.
\end{lemma}
\subsection{Layer-wise Induction with Intermediate Adversarial Examples}
We now return to the general multidimensional, multilayer, multiclass setting. While our approach successfully bounds the ARC for linear functions using intermediate adversarial examples, extending this method to DNNs presents inherent challenges. The primary difficulty arises from a fundamental conflict between the intermediate adversarial example approach and established methods for bounding DNN Rademacher complexity, such as layer peeling and covering number techniques.
The conflict stems from the dynamic nature of intermediate adversarial examples: the point $\bar{x}$ varies as we move from $(l-1)$-layer to $l$-layer networks. This variability contradicts a key requirement of traditional methods, which rely on fixed inputs at each layer.
We use the covering number approach by \citet{bartlett2017spectrally} to illustrate this challenge. Let $X_i$ denote the fixed output of the $i$-th layer. The covering number of the hypothesis class $\mathcal{H}$ is derived through induction, expressing the bound in Eq. \eqref{eq:stddecomp} as a sum of covering numbers over matrix space of $W_jx_{j-1}$.
\begin{equation}  
\label{eq:stddecomp}  
    \ln \mathcal{N} \left( \mathcal{H}, \|\cdot\|_\mathcal{S}, \zeta \right)  
    \leq \sum_{j=1}^l \sup_{\left( W_1, \dots, W_{j-1} \right)}  
    \ln \mathcal{N} \left( \left\{ W_j x_{j-1} : \left\| W_j^\top \right\|_{2,1} \leq a_j \right\}, \|\cdot\|, \delta_j \right).  
\end{equation}
for some $\zeta$. In the adversarial setting, however, $x_{j-1}$ is not fixed; it depends on both the weights of the preceding layers $(W_1,\ldots,W_{j-1})$ and the weights of subsequent layers $(W_j,\ldots,W_l)$. This interdependence between layers prevents the direct application of traditional covering number bounds to the adversarial setting. The same issue arises in the layer peeling approach, as detailed in Appendix \ref{appendix:B}. To address this challenge, we propose an alternative decomposition based on Lemma \ref{lemma:iae}, which bounds the covering number of the adversarial hypothesis class using the covering number of the weight space.
\begin{lemma}[Layer-wise Induction for Adversarial Hypothesis Class]
\label{lemma:induction}
Let $(\delta_1, \ldots ,\delta_l )$ be given, along with Lipschitz activation function $\rho$ (where $\rho$ is $L_\rho$-Lipschitz and $\rho(0)=0$). Let the loss function $\ell(f(x),y)$ be $L_\phi$-Lipschitz with respect to the first argument, i.e.,
$|\ell(u,y)-\ell(v,y)|\leq L_\phi\|u-v\|_2$. Let the function class be $\mathcal{F}_2$ or $\mathcal{F}_{1,\infty}$ and the adversarial hypothesis class be defined in \eqref{eq:adv-hypo-class}. Define
\begin{equation}  
    \zeta = L_{\phi} \sum_{j=1}^l L_\rho^{l-1} \frac{\prod_{k=1}^l M_k}{M_j} \max\left\{ 1, d^{1 - \frac{1}{r} - \frac{1}{p}} \right\} \left( \| X \|_{p,\infty} + \epsilon \right) \delta_j,  
\end{equation}
where $r=2$ for Frobenius norm and $r=1$ for $(1,\infty)$-norm. Then:
\label{lem:induction_of_layer}
\begin{equation*}  
    \begin{aligned}  
        \ln \left( \mathcal{N} \left( \tilde{\mathcal{H}}, \|\cdot\|_\mathcal{S}, \zeta \right) \right)  
        \leq \sum_{j=1}^l \ln \left( \mathcal{N} \left( \left\{ W_j \mid \|W_j\| \leq M_j \right\}, \|\cdot\|, \delta_j \right) \right).  
    \end{aligned}  
\end{equation*}
\end{lemma}
The proof is based on Lemma~\ref{lemma:iae}, which provides the bridge for the layer-wise induction given in Appendix~\ref{lem:induction_of_layer}. Since the upper bound is expressed as a sum of covering numbers over the weight spaces $W_j$ rather than the weight-input products $W_jx_{j-1}$, it effectively resolves the issue of input dependency on subsequent layer weights. 
 
\section{Bounds for Adversarial Rademacher Complexity}
\label{sec:arc-bounds}
\subsection{Binary Classification}
We begin by establishing bounds for the ARC in the binary classification setting.
\begin{theorem}[Frobenius Norm Bound]
	\label{ARCmulti}
	Consider the function class $\mathcal{F}_2^{\mathrm{binary}}$, and its corresponding adversarial function class $\tilde{\mathcal{H}}$ in Eq. \eqref{eq:adv-hypo-class}. The ARC of deep neural networks, $\mathcal{R}_\mathcal{S}(\tilde{\mathcal{H}})$, satisfies
\begin{equation*}  
\begin{aligned}  
\mathcal{R}_\mathcal{S}(\tilde{\mathcal{H}})  
\leq \frac{24 L_\phi}{\sqrt{n}} \max\left\{ 1, d^{\frac{1}{2} - \frac{1}{p}} \right\} \left( \|X\|_{p,\infty} + \epsilon \right) L_\rho^{l-1}  
\sqrt{\sum_{j=1}^l h_j h_{j-1} \log(3l)} \prod_{j=1}^l M_j.  
\end{aligned}  
\end{equation*}
Furthermore, under the assumptions that $L_\phi=1$, $L_\rho=1$, $p\leq 2$, $\|X\|_{p,\infty}= B$, and $h=\max\left\{h_0,\cdots,h_l\right\}$, we have
\begin{equation}  
\label{bound}  
\mathcal{R}_\mathcal{S}(\tilde{\mathcal{H}})  
\leq \mathcal{O} \left( \frac{(B+\epsilon) h \sqrt{l \log(l)} \prod_{j=1}^l M_j}{\sqrt{n}} \right).  
\end{equation}
\end{theorem}
The proof is based on bounding the Rademacher complexity using the covering number, which is known as Dudley’s integral. Specifically, we proceed as follows:  
\begin{itemize}
    \item We first establish an upper bound on the distance between two adversarial functions using Lemma~\ref{lemma:iae} (Intermediate Adversarial Example).  
    \item Next, we apply Lemma~\ref{lemma:induction} to relate the covering number of the adversarial hypothesis class to the covering number of the weight norm.  
    \item This reduces the problem to bounding the covering number of a norm ball, a well-established result in mathematical analysis. 
\end{itemize}
The complete proof is provided in Appendix~\ref{appendix:A}.

\begin{theorem}[$(1,\infty)$-norm Bound]
	\label{ARCinfty}
	Consider the function class $\mathcal{F}_{1,\infty}^{\mathrm{binary}}$, and its corresponding adversarial function class $\tilde{\mathcal{H}}$ in Eq. \eqref{eq:adv-hypo-class}. The ARC of deep neural networks, $\mathcal{R}_\mathcal{S}(\tilde{\mathcal{H}})$, satisfies
\begin{equation*}  
\begin{aligned}  
\mathcal{R}_\mathcal{S}(\tilde{\mathcal{H}})  
\leq \frac{24}{\sqrt{n}} \left( \left\| X \right\|_{p,\infty} + \epsilon \right) L_\rho^{l-1}  
\sqrt{\sum_{j=1}^l h_j h_{j-1} \log \left( 3l \right)} \prod_{j=1}^l M_j.  
\end{aligned}  
\end{equation*}
\end{theorem}
In the case of the \((1,\infty)\)-norm, the bound is similar to that of the Frobenius norm, except for the additional term \(\max\left\{1, d^{1/2 - 1/p} \right\}\). Therefore, for all \( p \geq 1 \), the \((1,\infty)\)-norm bound maintains the same order as in Eq. (\ref{bound}).

We compare our bound to the bounds in similar settings. Specifically, we compare our bound with the covering number bounds for (standard) Rademacher complexity \citep{bartlett2017spectrally} and the bound of ARC in two-layer cases.
\paragraph{Covering Number Bound for Standard Rademacher complexity.}  
The work of \citet{bartlett2017spectrally} used a covering number argument to show that the generalization gap is bounded by  
\[
\tilde{\mathcal{O}} \left( \frac{B \prod_{j=1}^l \left\| W_j \right\|}{\sqrt{n}} \left( \sum_{j=1}^l \frac{\left\| W_j \right\|_{2,1}^{2/3}}{\left\| W_j \right\|^{2/3}} \right)^{3/2} \right).
\]
Our bound differs in two key aspects. First, it includes an additional dependence on \(\epsilon\), which is unavoidable in adversarial settings. Second, as discussed in \citet{neyshabur2017pac} and \citet{golowich2018size}, the term $\bigg(\sum_{j=1}^l\frac{\|W_j\|_{2,1}^{2/3}}{\|W_j\|^{2/3}}\bigg)$ admits the following bounds:
\[
l^{\frac{3}{2}} \leq \left( \sum_{j=1}^l \frac{\left\| W_j \right\|_{2,1}^{2/3}}{\left\| W_j \right\|^{2/3}} \right)^{3/2} \leq l^{\frac{3}{2}} h.
\]
On the other hand, our bound exhibits a dependence on network size of \(\mathcal{O}(\sqrt{l\log(l)}h)\). The main difference arises from the need for two distinct approaches to perform layer-wise induction in standard and adversarial settings. Compared to the upper bound on network size dependence in existing standard results, our bound maintains a comparable dependence on depth \(l\) and width \(h\).

\paragraph{Bound for ARC in Two-Layer Cases.}  
The work of \citet{awasthi2020adversarial} established that the ARC is bounded by  
\begin{equation*}  
\mathcal{O} \left( \frac{(B+\epsilon) \sqrt{h_1 d} \sqrt{\log n} M_1 M_2}{\sqrt{n}} \right)  
\end{equation*}
in the two-layer setting. Applying our bound to the two-layer case, we obtain  
\[
\mathcal{O} \left( \frac{(B+\epsilon) \sqrt{h_1 d} M_1 M_2}{\sqrt{n}} \right),
\]
which is strictly tighter. Notably, under the same conditions for other factors, our bound exhibits a lower dependence on the sample size \(n\).

\begin{theorem}[Lower Bound]  
\label{lower}  
Consider the function class \(\mathcal{F}_2^{\mathrm{binary}}\). Let \(\tilde{\mathcal{F}}_2^{\mathrm{binary}}\) denote its corresponding adversarial function class as defined in Eq.~\eqref{hypo2}. There exists an activation function and a dataset \(S\) such that the ARC of deep neural networks satisfies  
\begin{equation*}  
\mathcal{R}_\mathcal{S}(\tilde{\mathcal{F}}_2^{\mathrm{binary}})  
\geq \Omega \left( \frac{(B+\epsilon) \prod_{j=1}^l M_j}{\sqrt{n}} \right).  
\end{equation*}
\end{theorem}
The proof is based on reducing the problem of lower bounding the ARC of DNNs to that of linear hypothesis classes, and it is provided in Appendix~\ref{appendix:A}. For function classes with the \((1,\infty)\)-norm, ARC admits the same lower bound. From this bound, we observe a gap in depth \( l \) and width \( h \) between the upper and lower bounds, while the other terms remain unavoidable. In the next section, we extend the ARC analysis to multi-class classification.

\subsection{Multi-Class Classification}
\label{margin}
The setting for multi-class classification follows \citep{bartlett2002rademacher}. In a K-class classification problem, let $\mathcal{Y}=\left\{1,2,\cdots,K\right\}$. The functions in the hypothesis class $\mathcal{F}$ map $\mathcal{X}$ to $\mathbb{R}^K$, the $k$-th output of $f$ is the score of $f(x)$ assigned to the $k$-th class.

Define the margin operator $M(f(x),y)=[f(x)]_y-\max_{y'\neq y}[f(x)]_{y'}$, where $[f(x)]_y$ denotes the $y$-th output of $f(x)$. The function makes a correct prediction if and only if $M(f(x),y)> 0$. We consider a particular loss function $\ell (f(x), y) = \phi_\gamma (M(f(x), y))$, where $\gamma > 0$ and $\phi_\gamma:\R \rightarrow [0,1]$ is the ramp loss:
\begin{equation*}\label{eq:ramp}
\phi_\gamma(t) = \begin{cases}
1 & t \le 0 \\
1-\frac{t}{\gamma} & 0 < t < \gamma \\
0 & t \ge \gamma.
\end{cases}
\end{equation*}
$\phi_\gamma(t) \in [0,1]$ and $\phi_\gamma(\cdot)$ is $1/\gamma$-Lipschitz. The loss function $\ell (f(x), y)$ satisfies:
\begin{equation*}  
\indi\left( y \neq \arg\max_{y' \in [K]} \left[ f(x) \right]_{y'} \right)  
\leq \ell \left( f(x), y \right)  
\leq \indi \left( \left[ f(x) \right]_y \leq \gamma + \max_{y' \neq y} \left[ f(x) \right]_{y'} \right).  
\end{equation*}
Define the function class $\ell_{\F} := \left\{(x, y) \mapsto \phi_\gamma (M(f(x), y)) : f\in\F\right\}$. In adversarial training, the adversarial hypothesis class is defined as
\begin{equation}  
\label{eq:multi_function_class}  
    \tilde{\mathcal{H}} := \left\{ (x, y) \mapsto \max_{x' \in \mathcal{B}(x)} \phi_\gamma \left( M \left( f(x'), y \right) \right) : f \in \F \right\},
\end{equation}
and assume $0\in \Tilde{\mathcal{H}}$. Then, the following generalization bound holds.
\begin{corollary}[\citep{yin2018rademacher}]\label{cor:multi_adv}
Consider the above adversarial multi-class classification setting. For any fixed $\gamma >0$, we have with probability at least $1-\delta$, for all $f \in \F$,
\begin{equation*}  
    \begin{aligned}  
        & \PP_{(x, y) \sim \D} \left\{ \exists x' \in \Ball^p_x(\epsilon) ~\text{s.t.}~ y \neq \arg\max_{y' \in [K]} \left[ f(x') \right]_{y'} \right\} \\  
        \leq & \frac{1}{n} \sum_{i=1}^n \indi \left( \exists x_i' \in \Ball^p_{x_i} (\epsilon) ~\text{s.t.}~ \left[ f( x_i' ) \right]_{y_i} \leq \gamma + \max_{y' \neq y} \left[ f( x_i' ) \right]_{y'} \right) \\  
        & + 2\mathcal{R}_{S}(\tilde{\mathcal{H}}) + 3\sqrt{\frac{\log \frac{2}{\delta}}{2n}}.  
    \end{aligned}  
\end{equation*}
\end{corollary}
Using the same idea as in binary settings, we can calculate the covering number of the adversarial hypothesis class via intermediate adversarial examples. Then, we have the following bound for ARC.
\begin{theorem}
\label{thm:multi}
	Given the function class $\mathcal{F}_2$, and the corresponding adversarial hypothesis class $\tilde{\mathcal{H}}$ in Eq. \eqref{eq:multi_function_class}, the ARC of deep neural networks $\mathcal{R}_\mathcal{S}(\tilde{\mathcal{H}})$ satisfies
\begin{equation}  
\begin{aligned}  
\mathcal{R}_\mathcal{S}(\tilde{\mathcal{H}})  
\leq \frac{48}{\gamma \sqrt{n}} \max \left\{ 1, d^{\frac{1}{2} - \frac{1}{p}} \right\} \left( \|X\|_{p,\infty} + \epsilon \right) L_\rho^{l-1}  
\sqrt{\sum_{j=1}^l h_j h_{j-1} \log(3l)} \prod_{j=1}^l M_j.  
\end{aligned}  
\end{equation}
\end{theorem}
The proof is based on the fact that $\phi_\gamma(M(\cdot,y))$ is $2/\gamma$-Lipschitz, which is provided in Appendix \ref{appendix:A}. Notably, our bound does not introduce an additional coordinate-wise multiplicative factor in the number of classes $K$. 
This should not be interpreted as saying that the bound is independent of $K$. 
In the multiclass setting, the output dimension is $h_l=K$, and the bound also depends on the last-layer norm constraint $M_l$. 
Thus, the number of classes can affect the bound through the architecture and the norm constraints. 
The point is that our covering-number argument controls the multiclass adversarial loss class directly, rather than decomposing it into $K$ scalar coordinate classes.

\section{Experiments}
\label{experiments}
\subsection{Comparing Standard and Adversarial Rademacher Complexity}
We now examine the relationship between the bounds for standard and adversarial Rademacher complexity. We begin by recalling the upper bound for standard Rademacher complexity from \citet{golowich2018size}:  
\begin{equation}  
\label{eq:compare}  
\mathcal{R}_\mathcal{S}(\mathcal{H})  
\leq \mathcal{O} \left( \frac{B \sqrt{l} \prod_{j=1}^l M_j}{\gamma \sqrt{n}} \right).  
\end{equation}
Next, we categorize the factors in the bounds into two groups.
\paragraph{Algorithm-Independent Factors.}  
The bounds include five algorithm-independent factors: the number of samples \(n\), depth \(l\), width \(h\), sample size \(B\), and perturbation intensity \(\epsilon\). For notational convenience, we define \(C_{std} = B\sqrt{l}/\sqrt{n}\) and \(C_{adv} = (B+\epsilon)h\sqrt{l\log l}/\sqrt{n}\) as the constants for standard and adversarial Rademacher complexity, respectively. By definition, \(C_{adv} > C_{std}\).

\paragraph{Algorithm-Dependent Factors.} There are two remaining terms: the product of upper bounds on matrix norms, \(\prod_{j=1}^l M_j\), and the margin, \(\gamma\). By definition, these terms are independent of the algorithm. However, they are implicitly algorithm-dependent \citep{bartlett2017spectrally,neyshabur2017pac,neyshabur2017exploring}, with more discussion provided in Appendix~\ref{optgamma}. The bound is universal and holds for all neural networks with weights satisfying \(\|W_j\| \leq M_j\) for \(j = 1, \dots, l\). Conversely, once a neural network is trained with specific weight norms \(\|W_j\|\), we can set \(M_j = \|W_j\|\) for all \(j\), ensuring that the bound applies specifically to the trained network and is the tightest possible bound for it. Similarly, \(\gamma\) is set to the margin of the trained network. Thus, the two algorithm-dependent factors in the bound are the product of weight norms and the margin. We define the ratio \( W_{std} := \prod_{j=1}^l \|W_j\| / \gamma \) for standard training and \( W_{adv} := \prod_{j=1}^l \|W_j\| / \gamma \) for adversarial training. Our experimental results, presented in the next subsection and Appendix \ref{appendix:C}, consistently show that \( W_{adv} > W_{std} \).

\paragraph{Generalization Gap Analysis.} Let $\mathcal{E}(\cdot)$ denote the standard generalization gap and $\tilde{\mathcal{E}}(\cdot)$ represent the robust generalization gap. We use $f_{std}$ and $f_{adv}$ to denote models trained using standard and adversarial training, respectively. Our analysis aims to understand why adversarially trained models exhibit substantially larger robust generalization gaps compared to the standard generalization gaps of normally trained models (i.e., why $\tilde{\mathcal{E}}(f_{adv})>\mathcal{E}(f_{std})$). While prior work \citep{zhang2021understanding} has established that Rademacher complexity is large in these settings, we can still analyze the relationship between robust generalization gaps and our identified factors through the standard and ARC bounds:

\begin{equation*}
	\tilde{\mathcal{E}}(f_{adv})\propto C_{adv}W_{adv}\quad and \quad \mathcal{E}(f_{std})\propto C_{std}W_{std}.
\end{equation*}

Notably, the bounds apply universally to any model. We can analyze the standard Rademacher complexity bound for adversarially trained models, i.e., $C_{std}W_{adv}$, and vice versa. To isolate the individual effects of factors $C_{adv}$ and $W_{adv}$, we examine two additional generalization gaps: the robust generalization gap of standard-trained models ($\tilde{\mathcal{E}}(f_{std})$) and the standard generalization gap of adversarially-trained models ($\mathcal{E}(f_{adv})$). These gaps help decompose the contributions of each factor:
\begin{equation*}
	\tilde{\mathcal{E}}(f_{std})\propto C_{adv}W_{std}\quad and \quad \mathcal{E}(f_{adv})\propto C_{std}W_{adv}.
\end{equation*}

In the previous section, we identified the normalized product of weight norms $\prod_{j=1}^l\|W_j\|/\gamma$ as a key algorithm-dependent factor in the ARC bounds. To empirically validate our theoretical analysis, we conducted extensive experiments comparing these terms between standard and adversarial training settings. Since our bounds generalize to convolutional neural networks, we evaluated VGG architectures \citep{simonyan2014very} on both CIFAR-10 and CIFAR-100 datasets \citep{krizhevsky2009learning}. Our analysis encompasses 88 trained models, with additional experimental results provided in Appendix \ref{appendix:C}.

\paragraph{Training Protocol.} We employed SGD optimization with a three-stage learning rate schedule: 0.1 for the first 100 epochs, 0.01 for the next 50 epochs, and 0.001 for the final 50 epochs. Weight decay was primarily set to $5\times 10^{-4}$, which was empirically determined to be optimal for robust accuracy, though we explored other values in ablation studies. For adversarial training, we implemented $\ell_\infty$ PGD \citep{madry2017towards} with $\epsilon=8/255$ perturbation intensity, using 20 steps during training and 40 steps during testing, with a step size of $2/255$ for inner maximization.
\paragraph{Margin Computation.} Following \citet{neyshabur2017exploring}, we defined margins differently for standard and adversarial training. For standard training, the margin was calculated as the 5th percentile of $f(x_i)[y_i]-\max_{y\neq y_i} f(x)[y]$ across all training points. For adversarial training, we used the 5th percentile of margins computed on PGD-adversarial examples. Since both the standard-trained and adversarially-trained models achieved 100\% training accuracy, the margins of all samples are positive. This ensures that the 5th percentile of margins is also positive. the  Detailed ablation studies on percentile selection are presented in Appendix \ref{C3}.

\begin{figure}[htpb]
	\centering
	\scalebox{0.9}{
		\subfigure[]{
			\begin{minipage}[htp]{0.49\linewidth}
				\centering
				\includegraphics[width=2.4in]{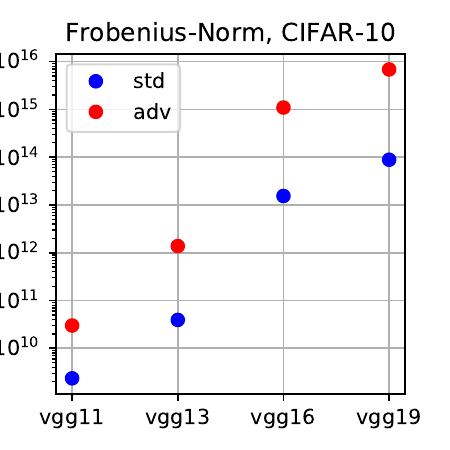}
			\end{minipage}%
		}
		\subfigure[]{
			\begin{minipage}[htp]{0.49\linewidth}
				\centering
				\includegraphics[width=2.4in]{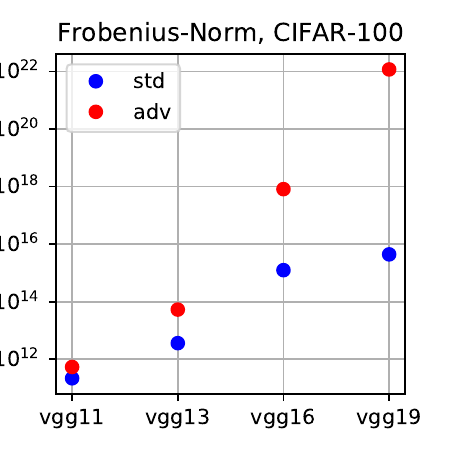}
			\end{minipage}
	}}
	\vskip -0.15in
	\caption{Comparison of Frobenius norms between standard and adversarial training models on CIFAR-10 and CIFAR-100 datasets.}
	\label{fig:norm}
	\vskip -0.1in
\end{figure}

\paragraph{Higher Weight Norms in Adversarially-Trained Models.} Figure \ref{fig:norm} compares weight norms between standard and adversarial training across VGG architectures on both CIFAR-10 and CIFAR-100 datasets\footnote{While larger models naturally exhibit higher weight norms due to their increased parameter count, our focus is on the relative difference between adversarially-trained and standard-trained models.}. Using a logarithmic scale for visualization, our results consistently demonstrate that adversarially-trained models have larger weight norms than their standard-trained counterparts ($W_{adv}\geq W_{std}$). Additional ablation studies in Appendix \ref{appendix:C} further confirm this relationship across different experimental conditions.
\paragraph{Analysis of Standard and Robust Generalization Gaps.} Table \ref{table1} presents standard and robust generalization gaps for both training approaches, using VGG-19 on CIFAR-10 as our primary example. Standard-trained models exhibit small standard generalization gaps ($\mathcal{E}(f_{std})=10.45\%$), while adversarially-trained models show larger standard generalization gaps ($\mathcal{E}(f_{adv})=26.34\%$). This increased gap aligns with the known phenomenon that adversarial training typically compromises standard generalization, possibly due to overfitting to adversarial examples.
The robust generalization gap presents a striking contrast: standard-trained models show minimal robust generalization gaps ($\tilde{\mathcal{E}}(f_{std})=0$), but this is not indicative of good performance. Rather, it reflects uniformly poor robustness, with both training and test robust accuracy approaching 0\%. Conversely, adversarially-trained models exhibit large robust generalization gaps ($\tilde{\mathcal{E}}(f_{adv})=58.90\%$). This substantial gap in robust generalization is a key phenomenon that we aim to analyze and explain.

\begin{table}\footnotesize
	\centering
	\caption{Comparison of four generalization gaps for VGG-19 trained on CIFAR-10: standard and robust gaps for both training methods. Note: $\tilde{\mathcal{E}}(f_{std})=0\%$ reflects complete model failure (100\% training error), while other cases achieve near-zero training errors.}
	\begin{tabular}{ccccc}
		\toprule
		&\multicolumn{2}{c}{Standard-trained models} & \multicolumn{2}{c}{Adversarially-trained models}\\
		\midrule
		Types of Generalization Gaps & Standard & Robust & Standard & Robust\\
		Training Errors & 0\%& 100\% & 0\% & 0.02\%\\
		Test Errors& 10.45\%& 100\% &26.34\%& 58.92\%\\
		Generalization Gaps & $\mathcal{E}(f_{std})$=10.45\%& $\tilde{\mathcal{E}}(f_{std})$=0\% & $\mathcal{E}(f_{adv})$=26.34\% & $\tilde{\mathcal{E}}(f_{adv})$=58.90\%\\
		\bottomrule
	\end{tabular}
	\label{table1}
\end{table}
\paragraph{Interpreting Zero Robust Generalization Gap in Standard Training.} Table \ref{table1} reveals that $\tilde{\mathcal{E}}(f_{std})=0\%$ represents a degenerate case where standard-trained models achieve 100\% robust training error, indicating complete failure to fit any adversarial examples in the training set. This renders both the generalization gap and its corresponding Rademacher complexity bound $\tilde{\mathcal{E}}(f_{std})\leq\mathcal{O}(C_{adv}W_{std}/\sqrt{n})$ trivial. In contrast, the other three cases achieve near-zero training errors, providing meaningful generalization gaps. We focus our analysis on understanding why $\tilde{\mathcal{E}}(f_{adv})>\mathcal{E}(f_{std})$ by examining the relationship $\tilde{\mathcal{E}}(f_{adv})>\mathcal{E}(f_{adv})>\mathcal{E}(f_{std})$.

\paragraph{Impact of $C_{adv}$ on Generalization Gaps.} Comparing generalization gaps for adversarially-trained models, we observe that the robust generalization gap significantly exceeds the standard generalization gap ($\tilde{\mathcal{E}}(f_{adv})=58.90\% > \mathcal{E}(f_{adv})=26.34\%$). Using Rademacher complexity bounds as approximations for these gaps, we can express this relationship as $\tilde{\mathcal{E}}(f_{adv})\propto C_{adv}W_{adv}$ and $\mathcal{E}(f_{adv})\propto C_{std}W_{adv}$. This suggests that $C_{adv}$ directly contributes to the increased robust generalization gap, as it scales with the perturbation intensity $\epsilon$.

\paragraph{Impact of $W_{adv}$ on Standard Generalization.} When comparing standard generalization gaps, we observe that adversarially-trained models exhibit poorer generalization compared to standard training ($\mathcal{E}(f_{adv})=26.34\% > \mathcal{E}(f_{std})=10.45\%$). This widely observed degradation in standard generalization can be understood through Rademacher complexity bounds. Using these bounds as approximations ($\mathcal{E}(f_{adv})\propto C_{std}W_{adv}$ and $\mathcal{E}(f_{std})\propto C_{std}W_{std}$), we can attribute the increased generalization gap to larger weight norms in adversarially-trained models ($W_{adv}$), demonstrating its positive correlation with generalization degradation.

The relationship between generalization gaps ($\tilde{\mathcal{E}}(f_{adv})>\mathcal{E}(f_{adv})>\mathcal{E}(f_{std})$) can be characterized by the corresponding complexity terms: $C_{adv}W_{adv}> C_{std}W_{adv}>C_{std}W_{std}$. This analysis reveals that robust generalization challenges stem from two distinct sources: (1) an algorithm-independent component $C_{adv}$, which is inherent to the minimax nature of adversarial training and thus unavoidable, and (2) an algorithm-dependent component $W_{adv}$, reflecting increased weight norms in adversarially-trained models, which might be addressable through improved training techniques.

\paragraph{Role of Weight Decay.} Our analysis suggests that controlling weight norms could improve robust generalization. In Appendix \ref{c4}, we investigate the effects of incrementally increasing weight decay. While larger weight decay values reduce weight norms and improve generalization, they also degrade training performance, revealing a fundamental trade-off between training accuracy and generalization. At weight decay of $10^{-2}$, training fails completely, though notably, even in this regime, adversarially-trained models maintain larger weight norms than standard-trained models.

\paragraph{Neural Network Representation Capacity.} We hypothesize that the increased weight norms in adversarial training stem from fundamental representation requirements: neural networks with small weight norms appear insufficient to fit adversarial examples in the training set, forcing the optimization to converge to solutions with larger weight norms. However, as our analysis shows, these large-norm solutions typically exhibit poor generalization properties, leading to robust overfitting. While this hypothesis aligns with our observations, comprehensive validation would require large-scale experimentation beyond our current scope.
\section{Conclusion}

\paragraph{Limitation.} The main limitation is that norm-based bounds tend to be excessively large in practical scenarios. As shown in Figure~\ref{fig:norm}, the bounds for VGG networks exceed $10^9$ in experiments on the CIFAR-10 dataset. The key challenge is how to obtain tighter norm-based bounds in real-world settings, not only for adversarial robustness but also in standard scenarios. This remains an open problem.

We present the first bounds on adversarial Rademacher complexity for deep neural networks, providing new theoretical insights into robust generalization. Our analysis reveals that robust generalization challenges arise from two distinct sources: an algorithm-independent factor inherent to the adversarial setting, and an algorithm-dependent factor related to neural network weight norms. Through extensive empirical validation, we establish clear correlations between these factors and robust generalization performance. These findings open new directions for both theoretical research in understanding adversarial training and practical improvements in robust generalization methods.
\clearpage
\acks{
We thank the action editor and the anonymous reviewers for their thoughtful comments and constructive suggestions. This work was supported by Hetao Shenzhen-Hong Kong Science and Technology Innovation Cooperation Zone Project (No.HZQSWS-KCCYB-2024016); Guangdong Provincial Key Laboratory of Mathematical Foundations for Artificial Intelligence (2023B1212010001).}

		\bibliography{main.bib}
\clearpage

\appendix
\setcounter{theorem}{0}

\section{Proofs of Technical Results}
\label{appendix:A}
\subsection{Proof of Lemma \ref{lemma:iae}}
\begin{proof}

Since $\mathcal{B}(x)$ is compact and for all $h\in\mathcal{H}$, $h(\cdot,y)$ is continuous, $\arg\max_{x'\in\mathcal{B}(x)}h(x',y)$ is attainable. Let 
$$x(\tilde{h}_1)=\arg\max_{x'\in\mathcal{B}(x)}h_1(x',y),\quad x(\tilde{h}_2)=\arg\max_{x'\in\mathcal{B}(x)}h_2(x',y).$$
By the property of $\arg\max$, we have $$h_1(x(\tilde{h}_1),y)-h_2(x(\tilde{h}_2),y)\leq h_1(x(\tilde{h}_1),y)-h_2(x(\tilde{h}_1),y)$$and
$$h_2(x(\tilde{h}_2),y)-h_1(x(\tilde{h}_1),y)\leq h_2(x(\tilde{h}_2),y)-h_1(x(\tilde{h}_2),y).$$ 
Let
\begin{equation}
\label{eq:iae}
    \bar{x}(\tilde{h}_1,\tilde{h}_2)=\begin{cases}
        x(\tilde{h}_1),\ \text{if}\  h_1(x(\tilde{h}_1),y)\geq h_2(x(\tilde{h}_2),y)\\
        x(\tilde{h}_2),\ \text{if}\  h_1(x(\tilde{h}_1),y)<h_2(x(\tilde{h}_2),y).
    \end{cases}
\end{equation}
\textbf{Case 1:} If $h_1(x(\tilde{h}_1),y)\geq h_2(x(\tilde{h}_2),y)$, then
\[
0 \leq h_1(x(\tilde{h}_1),y)- h_2(x(\tilde{h}_2),y) \leq h_1(x(\tilde{h}_1),y)-h_2(x(\tilde{h}_1),y).
\]
Since both $h_1(x(\tilde{h}_1),y)- h_2(x(\tilde{h}_2),y)$ and $h_1(x(\tilde{h}_1),y)-h_2(x(\tilde{h}_1),y)$ are nonnegative, it follows that 
\[
\left|h_1(x(\tilde{h}_1),y)- h_2(x(\tilde{h}_2),y)\right| \leq \left|h_1(x(\tilde{h}_1),y)-h_2(x(\tilde{h}_1),y)\right|.\]
In this case, $\bar{x}(\tilde{h}_1,\tilde{h}_2)=x(\tilde{h}_1)$, thus
\begin{equation*}
    \left|\tilde{h}_1(x,y)-\tilde{h}_2(x,y)\right|\leq\left|h_1(\bar{x}(\tilde{h}_1,\tilde{h}_2),y)-h_2(\bar{x}(\tilde{h}_1,\tilde{h}_2),y)\right|.
\end{equation*}
\textbf{Case 2:} If $h_1(x(\tilde{h}_1),y)< h_2(x(\tilde{h}_2),y)$, then
\[
0 \leq h_2(x(\tilde{h}_2),y)-h_1(x(\tilde{h}_1),y) \leq h_2(x(\tilde{h}_2),y)-h_1(x(\tilde{h}_2),y).
\]
Since both $h_2(x(\tilde{h}_2),y)-h_1(x(\tilde{h}_1),y)$ and $h_2(x(\tilde{h}_2),y)-h_1(x(\tilde{h}_2),y)$ are nonnegative, it follows that 
\[
\left|h_2(x(\tilde{h}_2),y)-h_1(x(\tilde{h}_1),y)\right| \leq \left|h_2(x(\tilde{h}_2),y)-h_1(x(\tilde{h}_2),y)\right|.\]
In this case, $\bar{x}(\tilde{h}_1,\tilde{h}_2)=x(\tilde{h}_2)$, thus
\begin{equation*}
    \left|\tilde{h}_2(x,y)-\tilde{h}_1(x,y)\right|\leq\left|h_2(\bar{x}(\tilde{h}_1,\tilde{h}_2),y)-h_1(\bar{x}(\tilde{h}_1,\tilde{h}_2),y)\right|.
\end{equation*}
Therefore, in either case 1 or case 2, we conclude that
\begin{equation*}
    \left|\tilde{h}_1(x,y)-\tilde{h}_2(x,y)\right|\leq\left|h_1(\bar{x}(\tilde{h}_1,\tilde{h}_2),y)-h_2(\bar{x}(\tilde{h}_1,\tilde{h}_2),y)\right|.
\end{equation*}
The expression in Eq. (\ref{eq:iae}) is the intermediate adversarial examples.
\end{proof}
\subsection{Proof of Lemma \ref{lem:induction_of_layer}}
The proof of Lemma \ref{lem:induction_of_layer} requires the following Lemma.
\begin{lemma}[\cite{awasthi2020adversarial}, cf. Lemma 1]  
\label{L2}  
If \( x_i^* \in \{ x_i' \mid \| x_i - x_i' \|_p \leq \epsilon \} \), then  
\begin{equation*}  
\| x_i^* \|_{r^*} \leq \max\{1, d^{1 - \frac{1}{r} - \frac{1}{p}}\} (\| X \|_{p,\infty} + \epsilon).  
\end{equation*}  
\end{lemma}  

\begin{proof}  
If \( p \geq r^* \), applying Hölder's inequality with \( 1/r^* = 1/p + 1/s \), we obtain  
\begin{equation*}  
\| x_i^* \|_{r^*}  
\leq \sup \|\mathbf{1} \|_s \| x_i^* \|_p  
= \|\mathbf{1} \|_s \| x_i^* \|_p  
= d^{\frac{1}{s}} \| x_i^* \|_p  
= d^{1 - \frac{1}{r} - \frac{1}{p}} \| x_i^* \|_p.  
\end{equation*}  
Equality holds when all entries are equal. If \( p < r^* \), we have  
\begin{equation*}  
\| x_i^* \|_{r^*} \leq \| x_i^* \|_p.  
\end{equation*}  
Equality holds when one entry equals one, and all others are zero. Thus,  
\begin{equation*}  
\begin{aligned}  
\| x_i^* \|_{r^*}  
&\leq \max\{1, d^{1 - \frac{1}{r} - \frac{1}{p}}\} \| x_i^* \|_p \\  
&\leq \max\{1, d^{1 - \frac{1}{r} - \frac{1}{p}}\} (\| x_i \|_p + \| x_i - x_i^* \|_p) \\  
&\leq \max\{1, d^{1 - \frac{1}{r} - \frac{1}{p}}\} (\| X \|_{p,\infty} + \epsilon).  
\end{aligned}  
\end{equation*}  
\end{proof}
\begin{lemma}  
\label{L3}  
Let \( A \) be an \( m \times k \) matrix and \( b \) be an \( n \)-dimensional vector. Then,  
\begin{equation*}  
\| A b \|_2 \leq \| A \|_F \| b \|_2.  
\end{equation*}  
\end{lemma}  

\begin{proof}  
Let \( A_i \) denote the rows of \( A \) for \( i = 1, \dots, m \). Then,  
\begin{equation*}  
\| A b \|_2 = \sqrt{\sum_{i=1}^{m} (A_i b)^2}  
\leq \sqrt{\sum_{i=1}^{m} \| A_i \|_2^2 \| b \|_2^2}  
= \sqrt{\sum_{i=1}^{m} \| A_i \|_2^2} \cdot \sqrt{\| b \|_2^2}  
= \| A \|_F \| b \|_2.  
\end{equation*}  
\end{proof}  

\begin{lemma}  
\label{L4}  
Let \( A \) be an \( m \times k \) matrix and \( b \) be an \( n \)-dimensional vector. Then,  
\begin{equation*}  
\| A b \|_\infty \leq \| A \|_{1,\infty} \| b \|_\infty.  
\end{equation*}  
\end{lemma}  

\begin{proof}  
Let \( A_i \) denote the rows of \( A \) for \( j = 1, \dots, m \). Then,  
\begin{equation*}  
\| A b \|_\infty = \max | A_i b |  
\leq \max \| A_i \|_1 \| b \|_\infty  
= \| A \|_{1,\infty} \| b \|_\infty.  
\end{equation*}  
\end{proof}  
Now, we move the the proof of Lemma~\ref{lem:induction_of_layer}.
\begin{proof}
In Frobenius norm case, let $r=2$ and $\mathcal{C}_j$ denote $\delta_j$-covers of the set $\{\|W_j\|_F\leq M_j\}$, for $j=1,2,\cdots,l$. Define
\begin{eqnarray*}  
        \mathcal{F}^c = \left\{ f^c : x \mapsto W_l^c \rho \left( W_{l-1}^c \rho \left( \cdots \rho \left( W_1^c x \right) \cdots \right) \right) , W_j^{c} \in \mathcal{C}_j, \ j = 1,2, \dots, l \right\};  
\end{eqnarray*}
In $(1, \infty)$-norm case, let $r=1$ and $\mathcal{C}_j^m$ be $\delta_j$-covers of $\{\|W_j^m\|_1\leq M_j\}$, $j=1,2,\cdots,l$, $m=1,\cdots,h_j$, where $W_j^m$ is the $m^{th}$ row of $W_j^m$. Let
\begin{eqnarray*}  
        \mathcal{F}^c = \left\{ f^c : x \mapsto W_l^c \rho \left( W_{l-1}^c \rho \left( \cdots \rho \left( W_1^c x \right) \cdots \right) \right),  
        W_j^{cm} \in \mathcal{C}_j^m, \ m = 1, \dots, h_j, \ j = 1,2, \dots, l \right\}.  
\end{eqnarray*}
Notably, the definition of $\mathcal{C}_j$ and $\mathcal{C}_j^m$ are different, particularly in dimension. Then, the following discussion holds for both $\mathcal{F}_2$ and $\mathcal{F}_{1,\infty}$. Define the adversarial hypothesis class as
\begin{equation*}  
\Tilde{\mathcal{H}}^c = \left\{ \Tilde{h} : \Tilde{h}(x,y) = \Tilde{\ell} \left( f(x), y \right), f \in \mathcal{F}^c \right\}.  
\end{equation*}
For any $\tilde{h}\in\tilde{\mathcal{H}}$, we aim to determine the smallest distance to $\tilde{\mathcal{H}^c}$, , which involves computing
\begin{eqnarray*}  
\max_{\tilde{h} \in \tilde{\mathcal{H}}} \min_{\tilde{h}^c \in \tilde{\mathcal{H}}^c} \left\| \tilde{h} - \tilde{h}^c \right\|_S.  
\end{eqnarray*}
$\forall ( x_i,y_i), i=1,\cdots,n$, given $\tilde{h}$ and $\tilde{h}^c$, by Lemma \ref{lemma:iae}, there exist an intermediate adversarial example $\bar{x}_i$, such that,
\begin{equation*}  
    \left| \tilde{h}(x_i,y_i) - \tilde{h}^c(x_i,y_i) \right| \leq \left| h(\bar{x}_i,y_i) - h^c(\bar{x}_i,y_i) \right|.  
\end{equation*}  
Since the loss function \(\ell(f(x), y)\) is \(L_\phi\)-Lipschitz with respect to the first argument,  
\begin{equation*}  
    \left| \tilde{h}(x_i,y_i) - \tilde{h}^c(x_i,y_i) \right| \leq L_\phi \| f(\bar{x}_i) - f^c(\bar{x}_i) \|.  
\end{equation*}
Define $g_b^a(\cdot)$ as
\begin{equation*}  
g_b^a(\bar{x}) = W_b \rho \left( W_{b-1} \rho \left( \cdots W_{a+1} \rho \left( W_a^c \cdots \rho \left( W_1^c \bar{x} \right) \cdots \right) \right) \right).  
\end{equation*}
In words, for the layers $b\geq j>a$ in $g_b^a(\cdot)$, the weight is $W_j$, for the layers $a\geq j\geq 1$ in $g_b^a(\cdot)$, the weight is $W_j^c$. Then we have $f(\bar{ x}_i)=g_l^0(\bar{ x}_i)$, $f^c(\bar{ x}_i)=g_l^l(\bar{ x}_i)$. We can decompose 
\begin{eqnarray}  
\label{eq:decom2}  
        \left| f(\bar{x}_i) - f^c(\bar{x}_i) \right|
    & = & \left| g_l^0(\bar{x}_i) - g_l^l(\bar{x}_i) \right| \nonumber\\  
    & = & \left| g_l^0(\bar{x}_i) - g_l^1(\bar{x}_i) + \cdots + g_l^{l-1}(\bar{x}_i) - g_l^l(\bar{x}_i) \right| \nonumber\\  
    & \leq & \left| g_l^0(\bar{x}_i) - g_l^1(\bar{x}_i) \right| + \cdots + \left| g_l^{l-1}(\bar{x}_i) - g_l^l(\bar{x}_i) \right|.  
\end{eqnarray}
To bound the gap $|f(\bar{ x}_i)-f^c(\bar{ x}_i)|$, we first calculate $|g_l^{j-1}(\bar{ x}_i)-g_l^{j}(\bar{ x}_i)|$ for $j=1,\cdots,l$.
\begin{eqnarray*}  
        \left| g_l^{j-1}(\bar{x}_i) - g_l^j(\bar{x}_i) \right| 
        &=& \left| W_l \rho \left( g_{l-1}^{j-1}(\bar{x}_i) \right) - W_l \rho \left( g_{l-1}^j(\bar{x}_i) \right) \right| \nonumber\\  
        &\overset{(\romannumeral1)}{\leq}& \| W_l \| \left\| \rho \left( g_{l-1}^{j-1}(\bar{x}_i) \right) - \rho \left( g_{l-1}^j(\bar{x}_i) \right) \right\|_{r^*} \nonumber\\  
        &\overset{(\romannumeral2)}{\leq}& L_\rho M_l \left\| g_{l-1}^{j-1}(\bar{x}_i) - g_{l-1}^j(\bar{x}_i) \right\|_{r^*} \nonumber\\  
        &\overset{(\romannumeral3)}{=}& L_\rho M_l \left\| W_{l-1} \rho \left( g_{l-2}^{j-1}(\bar{x}_i) \right) - W_{l-1} \rho \left( g_{l-2}^j(\bar{x}_i) \right) \right\|_{r^*} \nonumber\\  
        &\leq& \cdots \nonumber\\  
        &\leq& L_\rho^{l-j} \prod_{k=j+1}^l M_k \left\| W_j \rho \left( g_{j-1}^{j-1}(\bar{x}_i) \right) - W_j^c \rho \left( g_{j-1}^{j-1}(\bar{x}_i) \right) \right\|_{r^*}.  
\end{eqnarray*}  
where (\romannumeral1) is due to Lemma \ref{L3}, (\romannumeral2) is due to the bound of $\|W_j\|$ and the Lipschitz of $\rho(\cdot)$, (\romannumeral3) is because of the definition of $g_b^a(\bar{ x})$. Then
\begin{eqnarray}  
\label{eq:Lto1}  
\left| g_l^{j-1}(\bar{x}_i) - g_l^j(\bar{x}_i) \right|  
         & \leq & L_\rho^{l-j} \prod_{k=j+1}^l M_k \left\| W_j \rho \left( g_{j-1}^{j-1}(\bar{x}_i) \right) - W_j^c \rho \left( g_{j-1}^{j-1}(\bar{x}_i) \right) \right\|_{r^*} \nonumber\\  
        &=& L_\rho^{l-j} \prod_{k=j+1}^l M_k \left\| \left( W_j - W_j^c \right) \rho \left( g_{j-1}^{j-1}(\bar{x}_i) \right) \right\|_{r^*} \nonumber\\  
        &\overset{(\romannumeral1)}{\leq}& L_\rho^{l-j} \prod_{k=j+1}^l M_k \left\| W_j - W_j^c \right\| \left\| \rho \left( g_{j-1}^{j-1}(\bar{x}_i) \right) \right\|_{r^*} \nonumber\\  
        &\overset{(\romannumeral2)}{\leq}& L_\rho^{l-j} \prod_{k=j+1}^l M_k \delta_j \left\| \rho \left( g_{j-1}^{j-1}(\bar{x}_i) \right) \right\|_{r^*}.
\end{eqnarray}
where inequality (\romannumeral1) is due to Lemma \ref{L3}, inequality (\romannumeral2) is due to Lemma \ref{L3} and inequality (\romannumeral3) is due to  the assumption that $\|W_j-W_j^c\|\leq\delta_j$. It is lefted to bound $\|\rho(g_{j-1}^{j-1}(\bar{ x}_i)))\|_{r^*} $, we have
\begin{eqnarray}  
\label{eq:lto2}  
\left\| \rho \left( g_{j-1}^{j-1}(\bar{x}_i) \right) \right\|_{r^*} 
        &=& \left\| \rho \left( g_{j-1}^{j-1}(\bar{x}_i) \right) - \rho(0) \right\|_{r^*} \nonumber\\  
        &\leq& L_\rho \left\| g_{j-1}^{j-1}(\bar{x}_i) \right\|_{r^*} \nonumber\\  
        &=& L_\rho \left\| W_{j-1}^c \rho \left( g_{j-2}^{j-2}(\bar{x}_i) \right) \right\|_{r^*} \nonumber\\  
        &\leq& L_\rho \left\| W_{j-1}^c \right\| \left\| \rho \left( g_{j-2}^{j-2}(\bar{x}_i) \right) \right\|_{r^*} \nonumber\\  
        &\leq& L_\rho M_{j-1} \left\| \rho \left( g_{j-2}^{j-2}(\bar{x}_i) \right) \right\|_{r^*} \nonumber\\  
        &\leq& \cdots \nonumber\\  
        &\leq& L_\rho^{j-1} \prod_{k=1}^{j-1} M_k \max \left\{ 1, d^{1- \frac{1}{r} - \frac{1}{p}} \right\} \left( \| x \|_{p,\infty} + \epsilon \right).  
\end{eqnarray}
combining Eq. (\ref{eq:Lto1}) and (\ref{eq:lto2}), we have 
\begin{eqnarray}  
\label{eq:de2}  
\left| g_l^{j-1}(\bar{x}_i) - g_l^j(\bar{x}_i) \right| 
        &\leq& L_\rho^{l-1} \frac{\prod_{k=1}^l M_k}{M_j} \delta_j \max \left\{ 1, d^{1- \frac{1}{r} - \frac{1}{p}} \right\} \left( \| x \|_{p,\infty} + \epsilon \right) \nonumber\\  
        &=& \frac{D \delta_j}{2 M_j},
\end{eqnarray}
where 
\[
D=2L_\rho^{l-1} \prod_{k=1}^l M_k \max \left\{ 1, d^{1- \frac{1}{r} - \frac{1}{p}} \right\} \left( \| x \|_{p,\infty} + \epsilon \right).
\]
Therefore, combining Eq. (\ref{eq:decom2}) and (\ref{eq:de2}), we have 
\begin{eqnarray}  
\left| f(\bar{x}_i) - f^c(\bar{x}_i) \right|
    &\leq& \left| g_l^0(\bar{x}_i) - g_l^1(\bar{x}_i) \right| + \cdots + \left| g_l^{l-1}(\bar{x}_i) - g_l^l(\bar{x}_i) \right| \nonumber\\  
    &\leq& \sum_{j=1}^l \frac{D \delta_j}{2 M_j}. \nonumber 
\end{eqnarray}
Then
\begin{eqnarray*}  
\max_{\tilde{f} \in \tilde{\mathcal{F}}} \min_{\tilde{f}^c \in \tilde{\mathcal{F}}^c} \left\| \tilde{f} - \tilde{f}^c \right\|_S  
\leq \sum_{j=1}^l \frac{D \delta_j}{2 M_j}.  
\end{eqnarray*}
Let $\delta_j=2M_j\zeta/L_\phi lD$, $j=1,\cdots,l$, we have \[
\max_{\tilde{h} \in \tilde{\mathcal{H}}} \min_{\tilde{h}^c \in \tilde{\mathcal{H}}^c} \left\| \tilde{h} - \tilde{h}^c \right\|_S  
    \leq L_\phi \sum_{j=1}^l \frac{D \delta_j}{2 M_j}  
    \leq \zeta.
\]
We then calculate the $\zeta$-covering number $\mathcal{N}(\tilde{\mathcal{H}},\|\cdot\|_S,\zeta)$.
Because $\tilde{\mathcal{H}}^c$ is a $\zeta$-cover of $\tilde{\mathcal{H}}
$. The cardinality of $\tilde{\mathcal{H}}^c$ is 
\begin{eqnarray*}  
        \mathcal{N} \left( \tilde{\mathcal{H}}, \|\cdot\|_S, \zeta \right)  
    &=& \left| \tilde{\mathcal{H}}^c \right| \nonumber\\  
        &=& \begin{cases}  
\prod_{j=1}^l \left| \mathcal{C}_j \right| \quad & \text{if } r = 2; \\  
\prod_{j=1}^l \prod_{m=1}^{h_j} \left| \mathcal{C}_j^m \right| \quad & \text{if } r = 1.  
        \end{cases} \\  
        &=& \prod_{j=1}^l \mathcal{N} \left( \left\{ W_j \mid \|W_j\| \leq M_j \right\}, \|\cdot\|, \delta_j \right).  
\end{eqnarray*}
Therefore,
\begin{equation*}  
    \begin{aligned}  
        \ln \left( \mathcal{N} \left( \tilde{\mathcal{H}}, \|\cdot\|_\mathcal{S}, \zeta \right) \right)  
        \leq \sum_{j=1}^l \ln \left( \mathcal{N} \left( \left\{ W_j \mid \|W_j\| \leq M_j \right\}, \|\cdot\|, \delta_j \right) \right).  
    \end{aligned}  
\end{equation*}
\end{proof}

\subsection{Proof of Theorem~\ref{ARCmulti} and \ref{ARCinfty}}
 Before we provide the proof, we first introduce the Dudley's integral.
\begin{proposition}[Dudley's integral]
\label{thm:chaining}
	The Rademacher complexity $\mathcal{R}_S(\mathcal{F}^\mathrm{binary})$ satisfies
	\begin{eqnarray*}
	\mathcal{R}_S(\mathcal{F}^\mathrm{binary})\leq\inf_{\delta\geq 0}\bigg[8\delta+\frac{12}{\sqrt{n}}\int_{\delta}^{\text{Diam}/2}\sqrt{\log\mathcal{N}(\mathcal{F},\|\cdot\|_S,\zeta)}d\zeta\bigg].
	\end{eqnarray*}
\end{proposition}
This proposition is a well-established result in statistical learning theory, with detailed proofs available in standard references such as \citet{wainwright2019high}. Using this relationship between covering numbers and Rademacher complexity, we can derive upper bounds on the Rademacher complexity of function class $\mathcal{F}^\mathrm{binary}$ from its covering number bounds.

\begin{lemma}[Covering number of norm-balls (Lemma 6.27 in \citet{mohri2018foundations}]
\label{lemma:normball}
Let $\mathcal{B}$ be a $\ell_p$ norm ball with radius $W$. Let $d( x_1, x_2)=\| x_1- x_2\|_p$. Define the $\zeta$-covering number of $B$ as $\mathcal{N}(\mathcal{B},d(\cdot,\cdot),\zeta)$, we have
\begin{eqnarray*}  
\mathcal{N} \left( \mathcal{B}, d(\cdot,\cdot), \zeta \right) \leq \left( \frac{3W}{\zeta} \right)^d.  
\end{eqnarray*}
\end{lemma}
In the case of Frobenius norm ball of $m\times k$ matrices, we have the dimension $d=m\times k$ and 
\begin{eqnarray*}  
\mathcal{N} \left( \mathcal{B}, \|\cdot\|_F, \zeta \right)  
\leq  \left( \frac{3W}{\zeta} \right)^{m \times k}.  
\end{eqnarray*}
Now we move to the proof of Theorem~\ref{ARCmulti} and \ref{ARCinfty}.
\begin{proof} We first consider the Lipschitz constant of the loss function $\ell(f(x), y)=\phi(yf(x))$ in binary settings. Since 
\[
\left| \phi \left( y f(x_1) \right) - \phi \left( y f(x_2) \right) \right|  
\leq L_\phi \left| y f(x_1) - y f(x_2) \right|  
= L_\phi \left| f(x_1) - f(x_2) \right|,
\]
the loss function $\ell(f(x), y)=\phi(yf(x))$ is $L_\phi$-Lipschitz with respect to the first argument. Based on Lemma~\ref{lem:induction_of_layer}, define
\begin{equation*}  
    \zeta = L_\phi \sum_{j=1}^l L_\rho^{l-1} \frac{\prod_{k=1}^l M_k}{M_j} \max \left\{ 1, d^{1 - \frac{1}{r} - \frac{1}{p}} \right\} \left( \| x \|_{p,\infty} + \epsilon \right) \delta_j.  
\end{equation*}
where $r=2$ for Frobenius norm, which corresponds to Theorem \ref{ARCmulti}, and $r=1$ for $(1,\infty)$-norm, and which corresponds to Theorem \ref{ARCinfty}. Then:
\begin{eqnarray*}  
        \ln \left( \mathcal{N} \left( \tilde{\mathcal{H}}, \|\cdot\|_\mathcal{S}, \zeta \right) \right)  
        &\leq& \sum_{j=1}^l \ln \left( \mathcal{N} \left( \left\{ W_j \mid \|W_j\| \leq M_j \right\}, \|\cdot\|, \delta_j \right) \right) \nonumber\\  
        &=& \sum_{j=1}^l \ln \left| \mathcal{C}_j \right| \nonumber\\  
        &\overset{(\romannumeral1)}{\leq}& \sum_{j=1}^l \ln \left( \frac{3M_j}{\delta_j} \right)^{h_j h_{j-1}} \nonumber\\  
        &=& \ln \left( \frac{3L_\phi lD}{2\zeta} \right) \sum_{j=1}^l h_j h_{j-1}.  
\end{eqnarray*}
where inequality (\romannumeral1)) is due to Lemma \ref{lemma:normball}. 

Next, we consider the diameter of $\tilde{\mathcal{H}}$. Since $0\in\tilde{\mathcal{H}}$, for all $x,y$, there exists a function $f'\in\mathcal{F}$, such that $\tilde{h}'(x,y)=\max_{x'\in\mathcal{B}(x)}\ell(f'(x'),y)=0$. $\forall ( x_i,y_i), i=1,\cdots,n$, given $\tilde{h}$ and $\tilde{h}'$, by Lemma \ref{lemma:iae}, there exist an intermediate adversarial example $\bar{x}_i$, such that
\begin{equation*}  
   \left| \tilde{h}(x_i,y_i) \right| =\left| \tilde{h}(x_i,y_i) - \tilde{h}'(x_i,y_i) \right| \leq \left| h(\bar{x}_i,y_i) - h'(\bar{x}_i,y_i) \right|.  
\end{equation*}  
Since the loss function \(\ell(f(x), y)\) is \(L_\phi\)-Lipschitz with respect to the first argument,  
\begin{equation*}  
\begin{aligned}
    \left| \tilde{h}(x_i,y_i) - \tilde{h}'(x_i,y_i) \right| \leq & L_\phi \left| f(\bar{x}_i) - f'(\bar{x}_i) \right|\\
\leq & L_\phi \left(
\left| f(\bar{x}_i)\right| + \left|f'(\bar{x}_i) \right|\right)\\
\leq & 2L_\rho^{l} \prod_{k=1}^{l} M_k \max \left\{ 1, d^{1- \frac{1}{r} - \frac{1}{p}} \right\} \left( \| x \|_{p,\infty} + \epsilon \right)
    \end{aligned}
\end{equation*}
where the last inequality is due to Equation~\eqref{eq:lto2}. Therefore, the diameter of $\tilde{\mathcal{H}}$ is no larger than
\[
4L_\phi L_\rho^{l} \prod_{k=1}^{l} M_k \max \left\{ 1, d^{1- \frac{1}{r} - \frac{1}{p}} \right\} \left( \| x \|_{p,\infty} + \epsilon \right),
\]
which we denoted as $D_{\tilde{\mathcal{H}}}$.
By Dudley's integral, we have
\begin{eqnarray*}  
\mathcal{R}_S(\tilde{\mathcal{H}})
	&\leq& \inf_{\delta \geq 0} \bigg[ 8\delta + \frac{12}{\sqrt{n}} \int_{\delta}^{D_{\tilde{\mathcal{H}}}/2} \sqrt{\log \mathcal{N} (\mathcal{H}, \|\cdot\|_S, \zeta)} d\zeta \bigg] \nonumber\\  
	&\leq& \inf_{\delta \geq 0} \bigg[ 8\delta + \frac{12}{\sqrt{n}} \int_{\delta}^{D_{\tilde{\mathcal{H}}}/2} \sqrt{\left( \sum_{j=1}^l h_j h_{j-1} \right) \log(3L_\phi lD/2\zeta)} d\zeta \bigg] \nonumber\\  
\end{eqnarray*}
Let $\zeta'=2\zeta/D_{\tilde{\mathcal{H}}}$. By definition $D_{\tilde{\mathcal{H}}}=2 L_\phi D$, then $\zeta'=2\zeta/D_{\tilde{\mathcal{H}}}=\zeta/(L_\phi D)$. Therefore, 
\begin{eqnarray}
\label{eq:result1}&&\inf_{\delta \geq 0} \bigg[ 8\delta + \frac{12}{\sqrt{n}} \int_{\delta}^{D_{\tilde{\mathcal{H}}}/2} \sqrt{\left( \sum_{j=1}^l h_j h_{j-1} \right) \log(3L_\phi lD/2\zeta)} d\zeta \bigg] \nonumber\\  
&=&\inf_{\delta \geq 0} \bigg[ 8\delta + \frac{12}{\sqrt{n}} \int_{2\delta/D_{\tilde{\mathcal{H}}}}^{1} \sqrt{\left( \sum_{j=1}^l h_j h_{j-1} \right) \log(3L_\phi lD/2L_\phi D\zeta')} dL_\phi D\zeta' \bigg] \nonumber\\  
	&=& \inf_{\delta \geq 0} \bigg[ 8\delta + \frac{12L_\phi D \sqrt{\sum_{j=1}^l h_j h_{j-1}}}{\sqrt{n}} \int_{2\delta / D_{\tilde{\mathcal{H}}}}^{1} \sqrt{\log(3l/2\zeta')} d\zeta' \bigg].  
\end{eqnarray}
Let $\delta\rightarrow 0$. Then, we evaluate the integration
\[
\int_{0}^{1} \sqrt{\log\left(\frac{3l}{2\zeta'}\right)} \, d\zeta'.
\]
By Cauchy--Schwarz,
\[
\int_{0}^{1} \sqrt{\log\left(\frac{3l}{2\zeta'}\right)} \, d\zeta'
\;\le\; \sqrt{\int_{0}^{1} \log\left(\frac{3l}{2\zeta'}\right) \, d\zeta'}.
\]
A direct computation gives
\[
\int_{0}^{1} \log\left(\frac{3l}{2\zeta'}\right) \, d\zeta'
= \log\frac{3l}{2} \int_{0}^{1} d\zeta' - \int_{0}^{1} \log\zeta' \, d\zeta'
= \log\frac{3l}{2} + 1,
\]
since $\int_{0}^{1} \log \zeta' \, d\zeta' = -1$. Therefore
\[
\int_{0}^{1} \sqrt{\log\left(\frac{3l}{2\zeta'}\right)} \, d\zeta'
\;\le\; \sqrt{\log\frac{3l}{2} + 1}.
\]
Finally, because $l \ge 2$ implies $\log(3l) \ge \log 6$, we have
\[
\sqrt{\log\frac{3l}{2} + 1}
= \sqrt{\log(3l) + (1-\log 2)}
\;\le\; 2\sqrt{\log(3l)}.
\]

Plugging it to Eq. (\ref{eq:result1}), we have 
\begin{equation*}
\mathcal{R}_S(\tilde{\mathcal{H}})  
\leq \frac{24}{\sqrt{n}} \max \left\{ 1, d^{1- \frac{1}{r} - \frac{1}{p}} \right\} \left( \| x \|_{p,\infty} + \epsilon \right) L_\rho^{l-1}  
\sqrt{\sum_{j=1}^l h_j h_{j-1} \log(3l)} \prod_{j=1}^l M_j. 
\end{equation*}
If $r=1$, the bound reduces to the Frobenius norm bound in Theorem~\ref{ARCmulti}. If $r=1$, the bound reduces to the $(1,\infty)$-norm bound in Theorem~\ref{ARCinfty}. 
\end{proof}

\subsection{Proof of Theorem~\ref{lower}}
Let \(\rho(\cdot)\) be the identity activation function. The following discussion hods for both $\ell_2$-norm and $\ell_{1,\infty}$ norm cases. The proof of the theorem is based on constructing a linear network. By the definition of Rademacher complexity, if \(\mathcal{H}'\) is a subset of \(\mathcal{H}\), then  

\begin{eqnarray*}  
\mathcal{R}_\mathcal{S}(\mathcal{H}') &=& \mathbb{E}_\sigma \frac{1}{n} \left[ \sup_{h \in \mathcal{H}'} \sum_{i=1}^{n} \sigma_i h( x_i, y_i) \right] \\  
&\leq& \mathbb{E}_\sigma \frac{1}{n} \left[ \sup_{h \in \mathcal{H}} \sum_{i=1}^{n} \sigma_i h( x_i, y_i) \right]\\ &=& \mathcal{R}_\mathcal{S}(\mathcal{H}).  
\end{eqnarray*}

This inequality follows directly from the fact that restricting the hypothesis class cannot increase the supremum in the definition of Rademacher complexity.

Therefore, it suffices to lower bound the complexity of \(\tilde{\mathcal{F}}^{\mathrm{binary}'}\) under a specific distribution \(\mathcal{D}\), where \(\tilde{\mathcal{F}}^{\mathrm{binary}'}\) is a subset of \(\tilde{\mathcal{F}}^\mathrm{binary}\). We define  

\begin{eqnarray*}  
\tilde{\mathcal{F}}^{\mathrm{binary}'} = \left\{ x \mapsto \inf_{\| x' - x\|_p \leq \epsilon} y M_l \cdot M_2 w^T x \mid w \in \mathbb{R}^q, \|w\|_2 \leq M_1 \right\}.  
\end{eqnarray*}  
This formulation constrains the function class while maintaining a meaningful lower bound on its complexity.

We first prove that \(\tilde{\mathcal{F}}^{\mathrm{binary}'}\) is a subset of \(\tilde{\mathcal{F}}^{\mathrm{binary}}\). In \(\tilde{\mathcal{F}}^{\mathrm{binary}}\), we set the activation function \(\rho(\cdot)\) to be the identity mapping. Define  

\begin{eqnarray}  
\label{eq:addconstraint}  
W_1 &=&  
\begin{bmatrix}  
w \\  
0 \\  
\vdots \\  
0  
\end{bmatrix} \in \mathbb{R}^{h_1 \times h_0}, \quad  
W_j =  
\begin{bmatrix}  
M_j & 0 & \cdots & 0 \\  
0 & 0 & \cdots & 0 \\  
\vdots & \vdots & & \vdots \\  
0 & 0 & \cdots & 0  
\end{bmatrix} \in \mathbb{R}^{h_j \times h_{j-1}}, \quad j = 2, \dots, l.  
\end{eqnarray}  

Since \(\|W_j\| \leq M_j\), imposing the additional constraint in Eq. \eqref{eq:addconstraint} on \(\tilde{\mathcal{F}}^{\mathrm{binary}}\) reduces it to \(\tilde{\mathcal{F}}^{\mathrm{binary}'}\), confirming that \(\tilde{\mathcal{F}}^{\mathrm{binary}'}\) is a subset of \(\tilde{\mathcal{F}}^{\mathrm{binary}}\).  

To proceed, we need to establish a lower bound for the adversarial Rademacher complexity of linear hypothesis classes. This result follows from the work of \citet{yin2018rademacher, awasthi2020adversarial}, which we state below.
\setcounter{theorem}{3}
\begin{proposition}
Given the function class 
$$\mathcal{G}=\{ x\rightarrow y w^T x| w\in\mathbb{R}^q, \|w\|_r\leq W\}$$ 
and 
$$\tilde{\mathcal{G}}=\{ x\rightarrow \inf_{\| x'- x\|_r\leq\epsilon}yw^T x| w\in\mathbb{R}^q, \|w\|_r\leq W\},$$ 
the adversarial Rademacher complexity $\mathcal{R}_S(\tilde{\mathcal{G}})$ satisfies
\begin{eqnarray*}
\mathcal{R}_S(\tilde{\mathcal{G}})\geq\max\bigg\{\mathcal{R}_S(\mathcal{G}),\frac{\epsilon\max\{1,d^{1-\frac{1}{r}-\frac{1}{p}}\} W}{2\sqrt{n}}\bigg\}.
\end{eqnarray*}
\end{proposition}
The standard Rademacher complexity of linear hypothesis classes can be expressed as\begin{equation*}\mathcal{R}_S(\mathcal{G})=\frac{W}{m}\mathbb{E}_\sigma\|\sum_{i=1}^n\sigma_i x_i\|_{r*}.\end{equation*}
Let $\| x_i\|=B$ with equal entries for $i=1,\cdots,n$, by Lemma \ref{L2},
we have
\begin{eqnarray*}\mathcal{R}_S(\mathcal{G})\geq\frac{W}{n}\mathbb{E}_\sigma|\sum_{i=1}^n\sigma_i|\max\{1,d^{1-\frac{1}{r}-\frac{1}{p}}\}B.\end{eqnarray*}
By Khintchine’s inequality, we know that there exists a universal constant $c>0$ such that
\begin{eqnarray*}\mathbb{E}_\sigma|\sum_{i=1}^n\sigma_i|\geq c\sqrt{n}.\end{eqnarray*}
Then, we have
\begin{eqnarray*}\mathcal{R}_S(\mathcal{G})\geq\frac{cW}{\sqrt{n}}\max\{1,d^{1-\frac{1}{r}-\frac{1}{p}}\}B.\end{eqnarray*}
As discussed in \citet{awasthi2020rademacher}, this lower bound is in tight in terms of dependence on sample size $n$ and dimension $d$.
Therefore, 
\begin{eqnarray*}
\mathcal{R}_S(\tilde{\mathcal{G}})&\geq&\max\bigg\{\mathcal{R}_S(\mathcal{G}),\frac{\epsilon\max\{1,d^{1-\frac{1}{r}-\frac{1}{p}}\} W}{2\sqrt{n}}\bigg\}\\
&\geq&\frac{1}{1+2c}\mathcal{R}_S(\mathcal{G})+\frac{2c}{1+2c}\times\frac{\epsilon\max\{1,d^{1-\frac{1}{r}-\frac{1}{p}}\} W}{2\sqrt{n}}\\
&\geq&\frac{c}{1+2c}\bigg(\frac{(B+\epsilon)\max\{1,d^{1-\frac{1}{r}-\frac{1}{p}}\} W}{\sqrt{n}}\bigg).
\end{eqnarray*}

Let $W=\prod_{j=1}^l M_j$, we have
\begin{eqnarray*}
\mathcal{R}_S(\tilde{\mathcal{F}}^{\mathrm{binary}})
\geq\Omega\bigg(\frac{\max\{1,d^{1-\frac{1}{r}-\frac{1}{p}}\}(B+\epsilon)\prod_{j=1}^l M_j}{\sqrt{n}}\bigg),
\end{eqnarray*}
where $r=2$ for frobenius norm bound and $r=1$ for $\|\cdot\|_{1,\infty}$-norm bound. 
\subsection{Proof of Theorem \ref{thm:multi}}

\begin{proof}
For any $a,b\in\mathbb{R}^K$, we have
\begin{equation*}
\begin{aligned}
\left|M(a,y)-M(b,y)\right|
&=
\left|a_y-b_y-\left(\max_{y'\neq y}a_{y'}-\max_{y'\neq y}b_{y'}\right)\right|  \\
&\leq |a_y-b_y|+
\left|\max_{y'\neq y}a_{y'}-\max_{y'\neq y}b_{y'}\right|  \\
&\leq 2\|a-b\|_2 .
\end{aligned}
\end{equation*}
Since $\phi_\gamma$ is $1/\gamma$-Lipschitz, the loss function
$\ell(f(x),y)=\phi_\gamma(M(f(x),y))$ is $2/\gamma$-Lipschitz with respect
to the first argument.

Now we apply Lemma~\ref{lemma:induction} to the adversarial hypothesis class
$\tilde{\mathcal{H}}$ in Eq.~\eqref{eq:multi_function_class} with
$L_\phi=2/\gamma$ and $r=2$. For any $(\delta_1,\ldots,\delta_l)$, define
\begin{equation*}
\zeta
=
\frac{2}{\gamma}
\sum_{j=1}^l
L_\rho^{l-1}
\frac{\prod_{k=1}^l M_k}{M_j}
\max\left\{1,d^{\frac{1}{2}-\frac{1}{p}}\right\}
\left(\|X\|_{p,\infty}+\epsilon\right)\delta_j .
\end{equation*}
Then
\begin{equation*}
\ln \mathcal{N}\left(\tilde{\mathcal{H}},\|\cdot\|_\mathcal{S},\zeta\right)
\leq
\sum_{j=1}^l
\ln \mathcal{N}\left(
\left\{W_j:\|W_j\|_F\leq M_j\right\},
\|\cdot\|_F,\delta_j
\right).
\end{equation*}
By the standard covering number bound for Euclidean balls,
\begin{equation*}
\ln \mathcal{N}\left(
\left\{W_j:\|W_j\|_F\leq M_j\right\},
\|\cdot\|_F,\delta_j
\right)
\leq
h_jh_{j-1}\log\left(\frac{3M_j}{\delta_j}\right).
\end{equation*}
Following the same choice of $\delta_j$ and the same Dudley integral calculation
as in the proof of Theorem~\ref{ARCmulti}, we obtain
\begin{equation*}
\begin{aligned}
\mathcal{R}_\mathcal{S}(\tilde{\mathcal{H}})
\leq
\frac{48}{\gamma\sqrt{n}}
\max\left\{1,d^{\frac{1}{2}-\frac{1}{p}}\right\}
\left(\|X\|_{p,\infty}+\epsilon\right)
L_\rho^{l-1}
\sqrt{\sum_{j=1}^l h_jh_{j-1}\log(3l)}
\prod_{j=1}^l M_j .
\end{aligned}
\end{equation*}
This completes the proof.
\end{proof}
\section{Discussion on Existing Methods for Rademacher Complexity}
\label{appendix:B}
In this section, we review existing approaches for calculating Rademacher complexity, examine related work in the field, and highlight the challenges in analyzing adversarial Rademacher complexity.

\subsection{Existing Bounds for Standard Rademacher Complexity}
\label{exist}
\paragraph{Layer Peeling Technique.} The Rademacher complexity of multi-layer neural networks is primarily calculated using the `layer peeling' technique \citep{neyshabur2015norm}. For a function class $\mathcal{F}$ and function $g$, we define the composition $g\circ\mathcal{F}$ as $\{g\circ f| f\in\mathcal{F}\}$. Talagrand's Lemma establishes that $\mathcal{R}_\mathcal{S}(g\circ f)\leq L_g \mathcal{R}_\mathcal{S}(\mathcal{F})$. Applying this result to neural networks, we can show that $\mathcal{R}_\mathcal{S}(\mathcal{F}_{l})\leq 2L_\rho M_j \mathcal{R}_\mathcal{S}(\mathcal{F}_{l-1})$, where $\mathcal{F}_{l}$ represents the function class of $l$-layer neural networks. Since the Rademacher complexity of a linear function class is bounded by $\mathcal{O}(BM_1/\sqrt{n})$, induction yields an upper bound of $\mathcal{O}(B2^l L_\rho^{l-1}\prod_{j=1}^lM_j/\sqrt{n})$. For activation functions like ReLU where $L_\rho=1$, we can simplify this bound by eliminating the $L_\rho$ term.

\citet{golowich2018size} achieves an improved bound by reducing the depth dependence from $2^l$ to $\sqrt{l}$. Their key insight involves reformulating the Rademacher complexity expression $\mathbb{E}_\sigma[\cdot]$ as $\mathbb{E}_\sigma\exp\ln[\cdot]$. This transformation allows for layer peeling to be performed within the $\ln(\cdot)$ function, effectively containing the exponential $2^l$ term inside the logarithm and yielding the improved $\sqrt{l}$ dependence.

\paragraph{Covering Number.} \citet{bartlett2017spectrally} established a generalization gap bound using covering numbers:
\begin{eqnarray*}
\tilde{\mathcal{O}}\bigg(\frac{B\prod_{j=1}^l\|W_j\|}{\sqrt{n}}\bigg(\sum_{j=1}^l\frac{\|W_j\|_{2,1}^{2/3}}{\|W_j\|^{2/3}}\bigg)^{3/2}\bigg),
\end{eqnarray*}
where $\|\cdot\|$ denotes the spectral norm. Their proof employs layer-wise induction: for each layer $j$, let $W_j$ represent the layer's weight matrix and $X_j$ denote the network's output after processing through layers 1 to $j-1$. The bound is derived by inductively computing the matrix covering number $\mathcal{N}(\{W_jX_j\},\|\cdot\|_2,\epsilon)$ for each layer.

\subsection{Existing Bounds for Rademacher Complexity on Surrogate Loss}
\label{B2}
For linear models, ARC bounds can be derived directly from its definition \citep{khim2018adversarial,yin2018rademacher}. However, extending these analyses to multi-layer networks presents additional challenges. We begin by examining approaches that utilize surrogate loss functions.

\paragraph{Tree Transformation Loss.}  \citet{khim2018adversarial} introduced a tree transformation $T$ and demonstrated that $\max_{\|x-x'\|\leq \epsilon}\ell(f( x),y)\leq\ell(Tf( x),y)$. This leads to an upper bound on the adversarial population risk. Specifically, for any $\delta\in(0,1)$:
\begin{eqnarray*}
\tilde{R}(f)\leq R(Tf)\leq R_n(Tf)+2L\mathcal{R}_S(T\circ\mathcal{F})+3\sqrt{\frac{\log\frac{2}{\delta}}{2n}},
\end{eqnarray*}

where $R_n(Tf)$ represents the empirical risk of the transformed function, $\mathcal{R}_S(T\circ\mathcal{F})$ is the Rademacher complexity of the transformed function class, and $L$ is the Lipschitz constant. This result bounds the robust population risk in terms of empirical risk and the standard Rademacher complexity of the transformed function class $T\circ\mathcal{F}$. While $\mathcal{R}_S(T\circ\mathcal{F})$ serves as an approximation of adversarial Rademacher complexity, this bound has two key limitations: the empirical risk term $R_n(Tf)$ differs from the objective used in practical adversarial training, and the bound does not directly characterize the robust generalization gap between training and test performance.

\paragraph{SDP Relaxation Surrogate Loss.}
In \citep{yin2018rademacher}, the authors introduced an SDP surrogate loss to approximate the adversarial loss for two-layer neural networks. This surrogate loss is defined as:
\begin{eqnarray*}
\hat{\ell}(f( x),y)=\phi_\gamma\bigg(M(f( x),y)-\frac{\epsilon}{2}\max_{k\in [K],z=\pm 1}\max_{P\succeq0, diag(P)\leq 1}\langle zQ(w_{2,k},W_1),P\rangle\bigg).
\end{eqnarray*}
Using this formulation, the adversarial Rademacher complexity can be approximated by computing the Rademacher complexity of this surrogate loss function. However, this approach shares the same limitations as the previously discussed method, as it still does not directly characterize the robust generalization gap between training and test performance.

\paragraph{FGSM Attack Loss.}
\citet{gao2021theoretical} analyzed Rademacher complexity in the adversarial setting by focusing on Fast Gradient Sign Method (FGSM) adversarial examples to handle the $\max$ operation in the adversarial loss. Under specific gradient assumptions, they derived an upper bound for the adversarial Rademacher complexity using the loss $\ell(f(x_\text{FGSM}),y)$. Their analysis requires that $|\nabla \ell(f(x),y)|\geq\kappa$ holds for all $x$ in the domain, where $\kappa$ appears in the denominator of their final bound. This assumption proves problematic for two reasons: first, it is a strong condition that may not hold in practice, and second, the bound becomes unbounded as $\kappa$ approaches zero. Moreover, since their approach modifies the original loss function, it cannot provide guarantees on the robust generalization gap.

\subsection{Layer Peeling Technique for ARC}
\label{B3}
We first briefly introduce the layer peeling technique in standard settings.
\begin{eqnarray*}  
\mathcal{R}_\mathcal{S}(\mathcal{H}) &=& \mathbb{E}_\sigma \frac{1}{n} \left[ \sup_{h \in \mathcal{H}} \sum_{i=1}^{n} \sigma_i h( x_i) \right] \\  
&=& \mathbb{E}_\sigma \frac{1}{n} \left[ \sup_{h' \in \mathcal{H}_{l-1}, \|W_l\| \leq M_l} \sum_{i=1}^{n} \sigma_i W_l \rho(h'( x_i)) \right] \\  
&\leq& M_l \mathbb{E}_\sigma \frac{1}{n} \left[ \sup_{h' \in \mathcal{H}_{l-1}} \left\| \sum_{i=1}^{n} \sigma_i \rho(h'( x_i)) \right\| \right] \\  
&\leq& 2 M_l L_\rho \mathbb{E}_\sigma \frac{1}{n} \left[ \sup_{h' \in \mathcal{H}_{l-1}} \sum_{i=1}^{n} \sigma_i h'( x_i) \right] \\  
&=& 2 M_l L_\rho \mathcal{R}_\mathcal{S}(\mathcal{H}_{l-1}).  
\end{eqnarray*}  
In adversarial settings, if we directly apply the layer peeling technique, we have  
\begin{eqnarray*}  
\mathcal{R}_\mathcal{S}(\tilde{\mathcal{H}}) &=& \mathbb{E}_\sigma \frac{1}{n} \left[ \sup_{h \in \mathcal{H}} \sum_{i=1}^{n} \sigma_i \max_{\| x_i - x_i' \| \leq \epsilon} h( x_i') \right] \\  
&=& \mathbb{E}_\sigma \frac{1}{n} \left[ \sup_{h' \in \mathcal{H}_{l-1}, \|W_l\| \leq M_l} \sum_{i=1}^{n} \sigma_i W_l \rho(h'( x_i^*(h))) \right] \\  
&\leq& M_l \mathbb{E}_\sigma \frac{1}{n} \left[ \sup_{h' \in \mathcal{H}_{l-1}} \left\| \sum_{i=1}^{n} \sigma_i \rho(h'( x_i^*(h))) \right\| \right] \\  
&\leq& 2 M_l L_\rho \mathbb{E}_\sigma \frac{1}{n} \left[ \sup_{h' \in \mathcal{H}_{l-1}} \sum_{i=1}^{n} \sigma_i h'( x_i^*(h)) \right] \\  
&\neq& 2 M_l L_\rho \mathbb{E}_\sigma \frac{1}{n} \left[ \sup_{h' \in \mathcal{H}_{l-1}} \sum_{i=1}^{n} \sigma_i h'( x_i^*(h')) \right] \\  
&=& 2 M_l L_\rho \mathcal{R}_\mathcal{S}(\tilde{\mathcal{H}}_{l-1}).  
\end{eqnarray*}

where $x_i^*(h)$ and $x_i^*(h')$ denote the optimal adversarial examples for $l$-layer and $(l-1)$-layer neural networks, respectively. Since $x_i^*(h) \neq x_i^*(h')$ in general, the optimal adversarial examples differ between architectures of different depths, which prevents the direct extension of layer peeling techniques to the adversarial setting.

\subsection{Comparison of Adversarial Generalization Bounds}

\paragraph{VC-Dimension Bounds.} 
The VC dimension is a fundamental tool in statistical learning theory for bounding generalization gaps. Several works, including \citet{cullina2018pac}, \citet{montasser2019vc}, and \citet{attias2021improved}, have extended this framework to the adversarial setting. However, these approaches fail to provide computable bounds on the adversarial generalization gap, as we explain below. Let $\mathcal{H}$ denote the hypothesis class (for example, the set of neural networks with a fixed architecture).

\citet{cullina2018pac} introduced the concept of adversarial VC dimension (AVC) and established bounds on the adversarial generalization gap in terms of $\text{AVC}(\mathcal{H})$. However, they did not provide methods to compute the AVC for neural networks, thus leaving their bounds non-computable in practice.

\citet{montasser2019vc} took a different approach by defining an adversarial function class $\mathcal{L}_{\mathcal{H}}^{\mathcal{U}}$, where $\mathcal{L}$ represents the loss function and $\mathcal{U}$ denotes the uncertainty set. While their bound on the adversarial generalization gap using $\mathcal{L}_{\mathcal{H}}^{\mathcal{U}}$ differs from the $\text{AVC}(\mathcal{H})$ approach of \citet{cullina2018pac}, they similarly did not provide a method to compute their bound, rendering it non-computable in practice.

\citet{attias2021improved} analyze the case where the perturbation set $U(x)$ is finite, containing exactly $k$ possible adversarial examples for each sample $x$. Under this assumption, they bound the adversarial generalization gap by:
\begin{eqnarray*}  
\mathcal{O} \left( \frac{1}{\zeta^2} \left( \sqrt{k \cdot VC(\mathcal{H})} \log \left( \frac{3}{2} + a \right) k \cdot VC(\mathcal{H}) \right) + \log \frac{1}{\delta} \right).  
\end{eqnarray*}
Since $\text{VC}(\mathcal{H})$ can be upper-bounded by the number of network parameters, this result provides a computable bound, improving upon previous approaches. However, this computability comes with a significant limitation: the bound depends on $k$, the number of allowed perturbed samples, which deviates from the standard notion of adversarial generalization where $U(x)$ is an infinite set. In contrast, our bound applies to the original adversarial generalization framework where $U(x)$ is infinite ($k=+\infty$).

\paragraph{Adversarial Generalization in Alternative Settings.} Several studies have explored adversarial generalization in specific model architectures. \citet{javanmard2020precise} analyzed generalization properties in linear regression, while multiple researchers \citep{taheri2020asymptotic,javanmard2020precise,dan2020sharp} investigated adversarial generalization using Gaussian mixture models. Studies on uniform stability in adversarial training \citep{xing2021on,xiao2022stability,xiao2022adaptive,xiao2022smoothedsgdmax,pmlr-v235-xiao24e} suggest that poor generalization may result from the non-smooth nature of adversarial loss functions.

\paragraph{Related Work on Regularization and Robustness.} 
The connection between robustness and regularization has been widely studied in the literature. 
A line of work has explored Jacobian regularization, where robustness is promoted by directly penalizing the sensitivity of model outputs to input perturbations \citep{jakubovitz2018improving, yu2018interpreting, ross2018improving, finlay2021scaleable}. 
Another line has investigated weight regularization, where robustness emerges from penalizing the norm of the model parameters \citep{Blanchet2019robust, sinha2017certifiable, rahimian2022frameworks}. 
Since weight decay is equivalent to $\ell_2$ regularization, our results naturally relate to this literature, suggesting that weight decay not only controls model capacity but may also contribute to adversarial robustness. 
We highlight these connections to situate our contribution within the broader context of robustness through regularization. Related work by Xing and collaborators investigates adversarial robustness through robust statistics, generalization theory, training dynamics, and minimax analysis, covering topics such as robust linear regression, the generalization behavior of adversarial training, phase transitions from clean to adversarial training, and the benefit of unlabeled data for robustness \citep{xing2021adversarially,xing2021generalization,xing2022phase,xing2022unlabeled}.

\paragraph{Certified Robustness.} Research on certified robustness focuses on guaranteeing model performance within norm-constrained neighborhoods of training data. \citet{cohen2019certified} developed certification methods through random smoothing, while \citet{lecuyer2019certified} approached certification through differential privacy frameworks.

\paragraph{Geometric Perspectives on Adversarial Examples.} Several works have examined the geometric properties of adversarial examples \citep{gilmer2018adversarial,khoury2018geometry}. The off-manifold hypothesis, first proposed by \citet{szegedy2013intriguing}, suggests that adversarial examples deviate from the underlying data manifold. Supporting this view, \citet{song2017pixeldefend} demonstrated using generative models that adversarial examples typically occupy low-probability regions of the data distribution. Similarly, \citet{ma2018characterizing} employed Local Intrinsic Dimensionality (LID) to show that adversarial subspaces are both low-probability and distinct from the data submanifold.

\section{Additional Experiments}
\label{appendix:C}
In this section, we present additional experimental results to further validate our theoretical findings and explore their implications.

\subsection{Experiments on VGG Architectures}
\begin{figure}[htbp]
\scalebox{0.9}{
			\begin{minipage}[htp]{0.01\linewidth}
			\centering
			\rotatebox{90}{\text{VGG-11}}
		\end{minipage}
		\subfigure[]{
			\begin{minipage}[htp]{0.26\linewidth}
				\centering
				\includegraphics[width=1.5in]{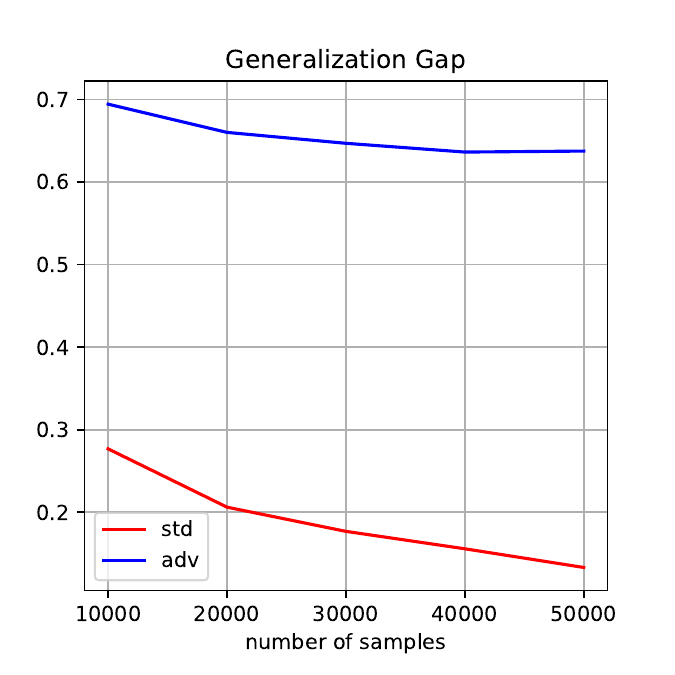}
			\end{minipage}%
		}
		\subfigure[]{
			\begin{minipage}[htp]{0.26\linewidth}
				\centering
				\includegraphics[width=1.5in]{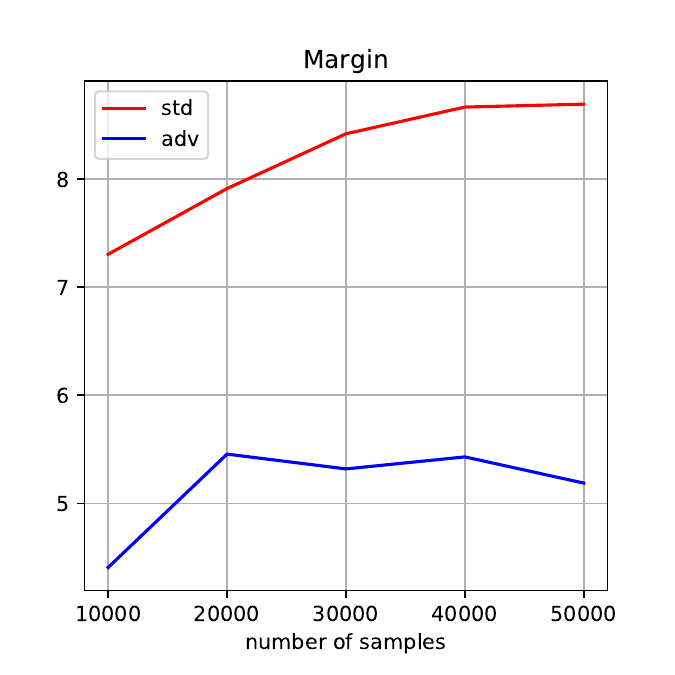}
			\end{minipage}
		}
		\subfigure[]{
			\begin{minipage}[htp]{0.26\linewidth}
				\centering
				\includegraphics[width=1.5in]{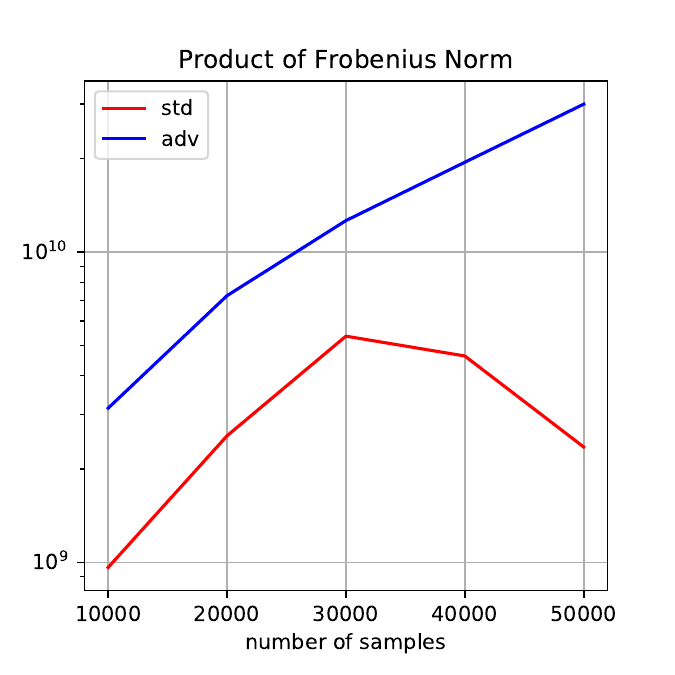}
			\end{minipage}
		}
		\subfigure[]{
			\begin{minipage}[htp]{0.26\linewidth}
				\centering
				\includegraphics[width=1.5in]{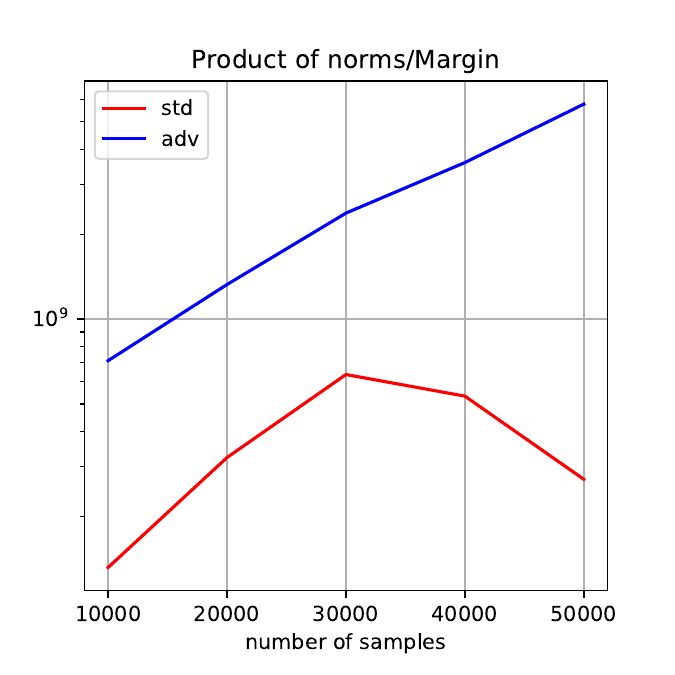}
			\end{minipage}
	}}
	\scalebox{0.9}{
			\begin{minipage}[htp]{0.01\linewidth}
			\centering
			\rotatebox{90}{\text{VGG-13}}
		\end{minipage}
		\subfigure[]{
			\begin{minipage}[htp]{0.26\linewidth}
				\centering
				\includegraphics[width=1.5in]{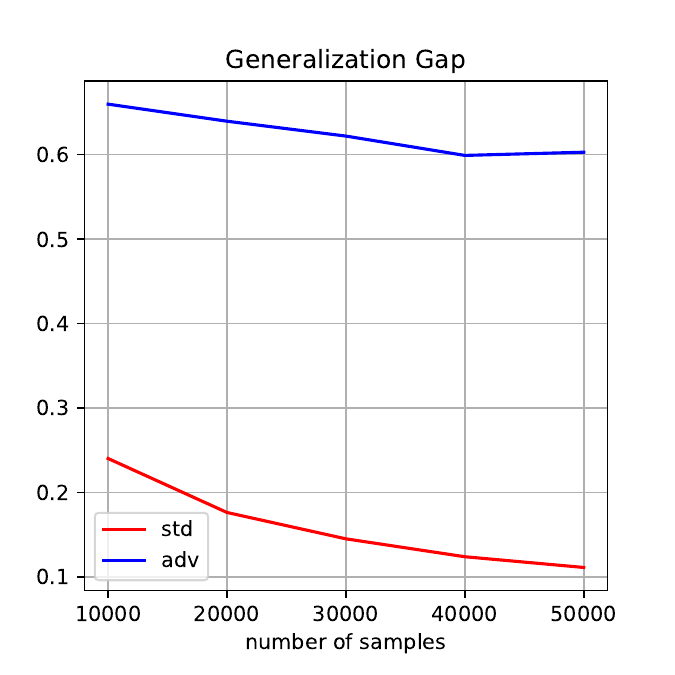}
			\end{minipage}
		}
		\subfigure[]{
			\begin{minipage}[htp]{0.26\linewidth}
				\centering
				\includegraphics[width=1.5in]{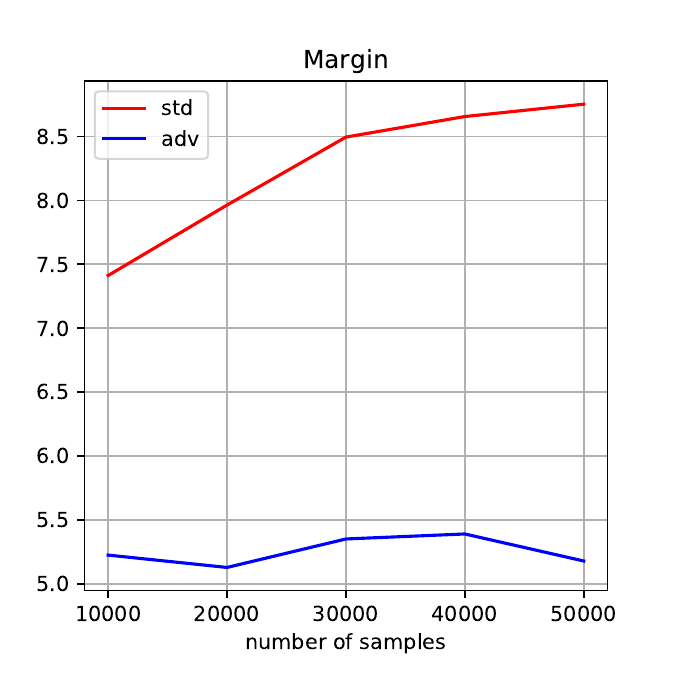}
			\end{minipage}
		}
		\subfigure[]{
			\begin{minipage}[htp]{0.26\linewidth}
				\centering
				\includegraphics[width=1.5in]{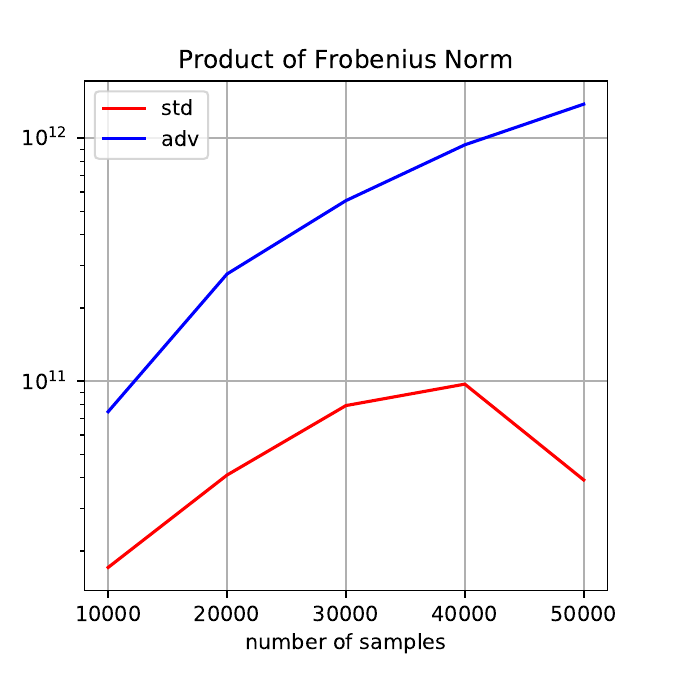}
			\end{minipage}
		}
		\subfigure[]{
			\begin{minipage}[htp]{0.26\linewidth}
				\centering
				\includegraphics[width=1.5in]{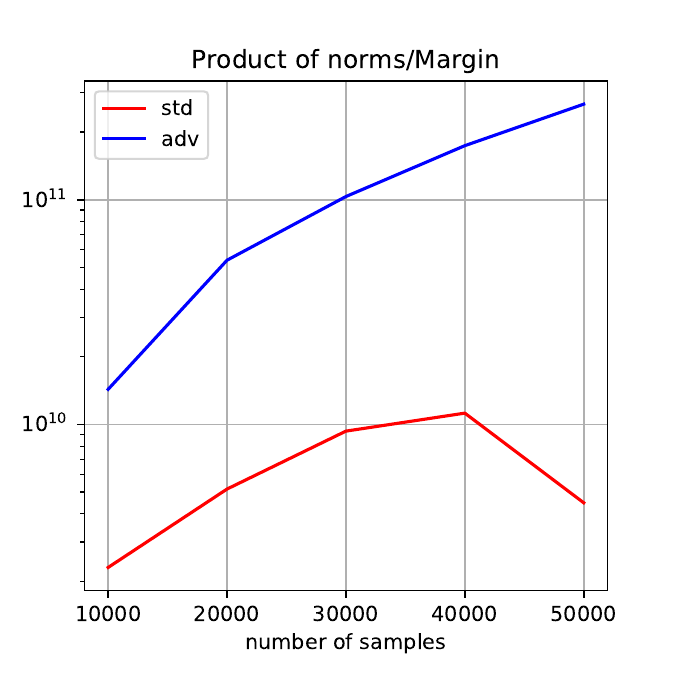}
			\end{minipage}
	}}
	\centering
	\vskip -0.15in
	\caption{Comparison of Frobenius norm products across VGG architectures. Results from standard training (red lines) and adversarial training (blue lines) for VGG-11 (top row) and VGG-13 (bottom row). The plots show: (a,e) Generalization gap, (b,f) Training set margin $\gamma$, (c,g) Product of layer-wise Frobenius norms $\prod_{j=1}^l\|W_j\|_F$, and (d,h) Ratio of Frobenius norm product to margin $\prod_{j=1}^l\|W_j\|_F/\gamma$.}
	\label{fig:vgg11_13}
	\vskip -0.1in
\end{figure}

\begin{figure}[htbp]
	\centering
\scalebox{0.9}{
			\begin{minipage}[htp]{0.01\linewidth}
			\centering
			\rotatebox{90}{\text{VGG-16}}
		\end{minipage}
		\subfigure[]{
			\begin{minipage}[htp]{0.26\linewidth}
				\centering
				\includegraphics[width=1.5in]{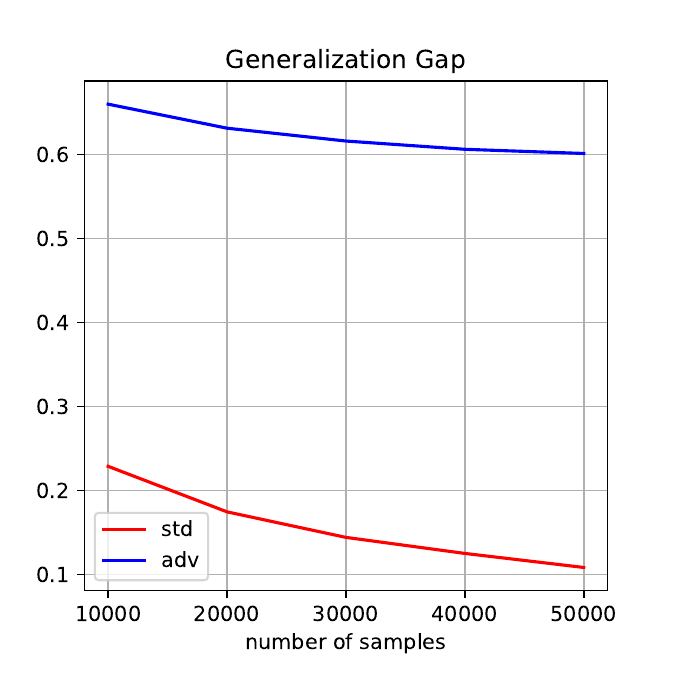}
			\end{minipage}%
		}
		\subfigure[]{
			\begin{minipage}[htp]{0.26\linewidth}
				\centering
				\includegraphics[width=1.5in]{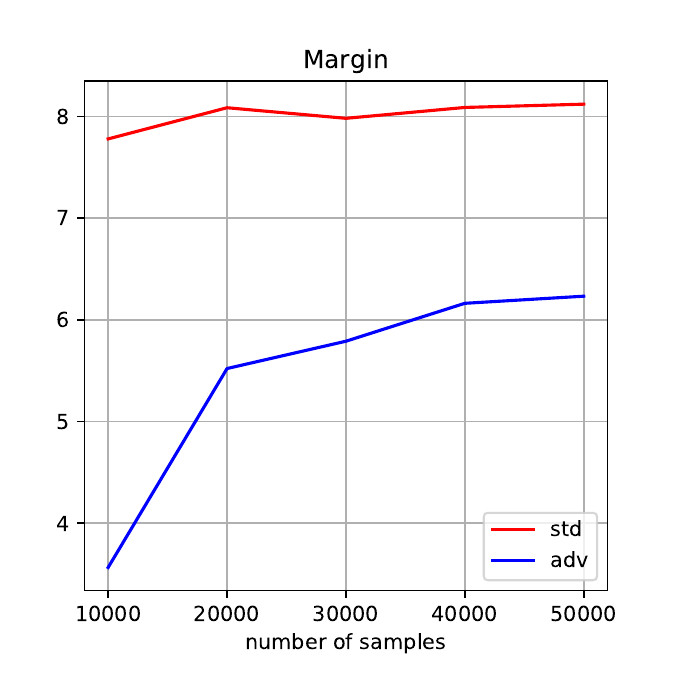}
			\end{minipage}
		}
		\subfigure[]{
			\begin{minipage}[htp]{0.26\linewidth}
				\centering
				\includegraphics[width=1.5in]{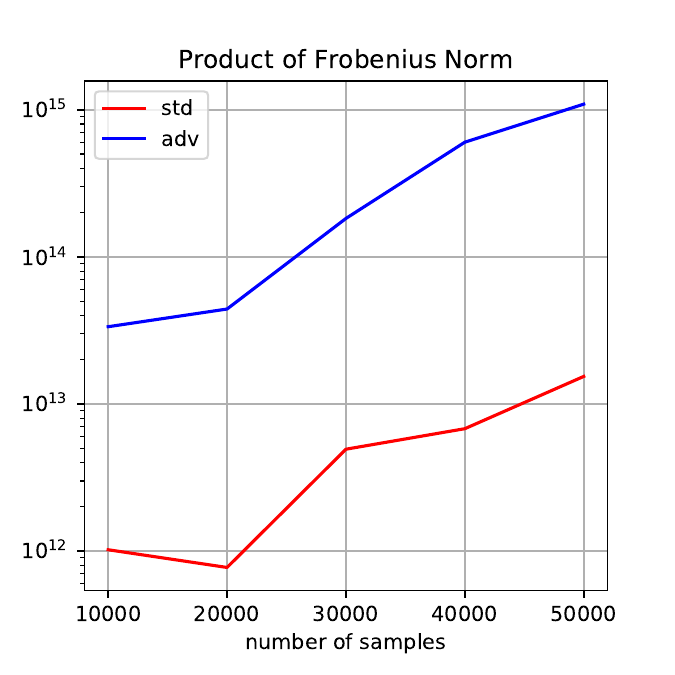}
			\end{minipage}
		}
		\subfigure[]{
			\begin{minipage}[htp]{0.26\linewidth}
				\centering
				\includegraphics[width=1.5in]{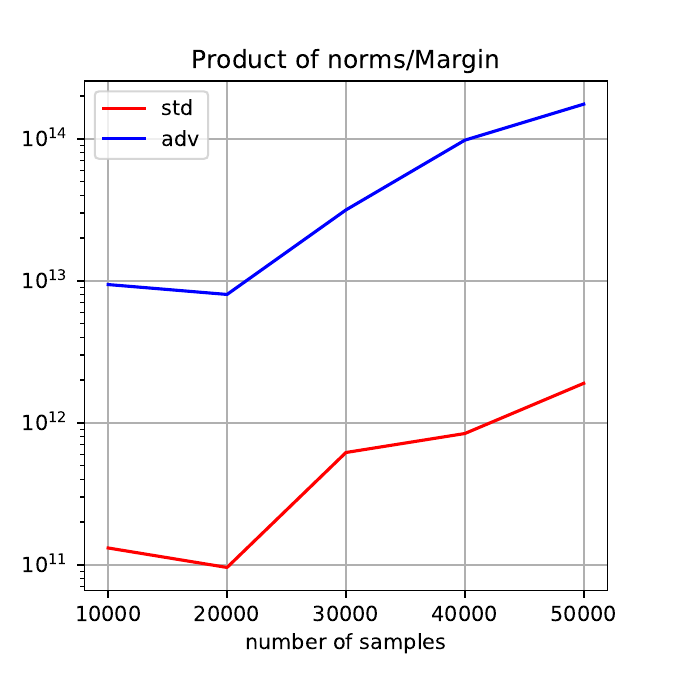}
			\end{minipage}
	}}
	
	\scalebox{0.9}{
			\begin{minipage}[htp]{0.01\linewidth}
			\centering
			\rotatebox{90}{\text{VGG-19}}
		\end{minipage}
		\subfigure[]{
			\begin{minipage}[htp]{0.26\linewidth}
				\centering
				\includegraphics[width=1.5in]{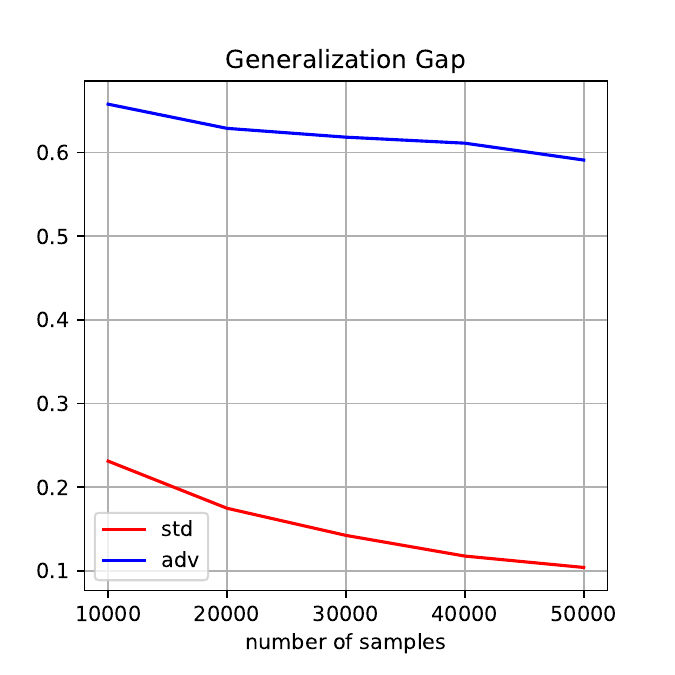}
			\end{minipage}%
		}
		\subfigure[]{
			\begin{minipage}[htp]{0.26\linewidth}
				\centering
				\includegraphics[width=1.5in]{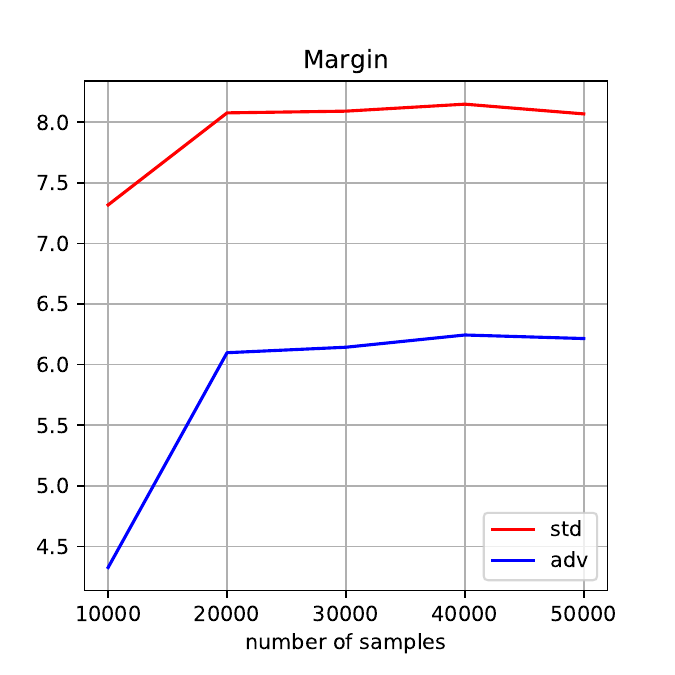}
			\end{minipage}
		}
		\subfigure[]{
			\begin{minipage}[htp]{0.26\linewidth}
				\centering
				\includegraphics[width=1.5in]{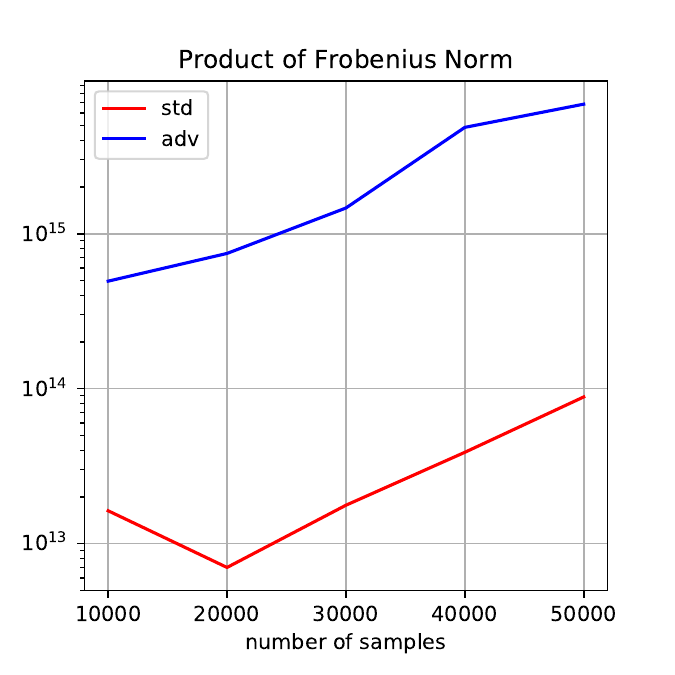}
			\end{minipage}
		}
		\subfigure[]{
			\begin{minipage}[htp]{0.26\linewidth}
				\centering
				\includegraphics[width=1.5in]{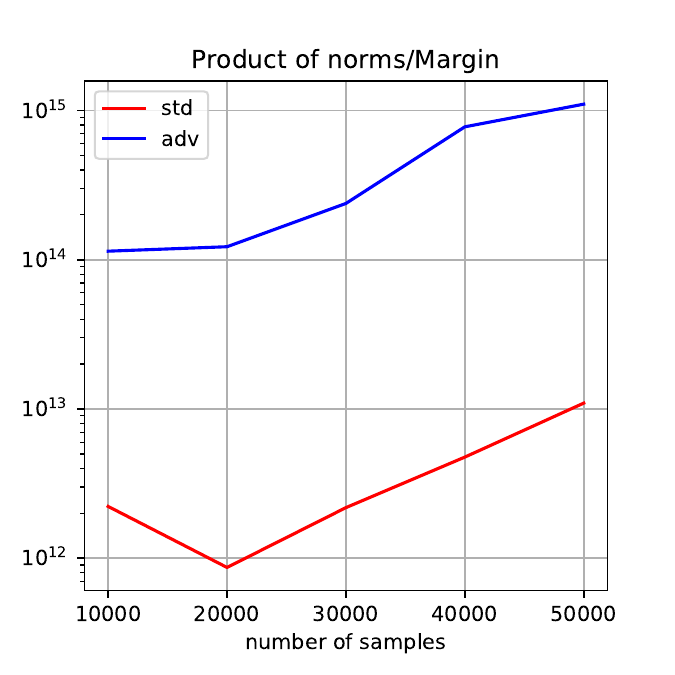}
			\end{minipage}
	}}
	\centering
	\vskip -0.15in
\caption{CIFAR-10 experimental results comparing standard training (red lines) and adversarial training (blue lines) for VGG-16 (top row) and VGG-19 (bottom row). The plots show: (a,e) Generalization gap, (b,f) Training set margin, (c,g) Product of layer-wise Frobenius norms $\prod_{j=1}^l\|W_j\|_F$, and (d,h) Ratio of Frobenius norm product to margin $\prod_{j=1}^l\|W_j\|_F/\gamma$.}
\label{fig:fro}
\end{figure}

Figure \ref{fig:vgg11_13} presents experimental results for VGG-11 and VGG-13 architectures, while Figure \ref{fig:fro} demonstrates findings for VGG-16 and VGG-19. Our analysis reveals a substantial difference in the product of Frobenius norms between standard and adversarial training methods, which corresponds to poor generalization performance in the adversarial setting.

\begin{figure}[htbp]
	\centering

\scalebox{0.9}{
		\subfigure[]{
			\begin{minipage}[htp]{0.26\linewidth}
				\centering
				\includegraphics[width=1.5in]{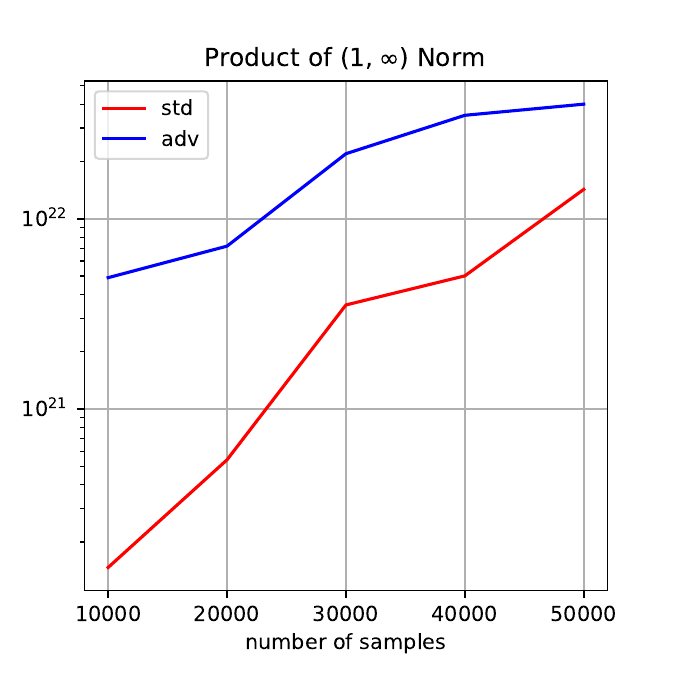}
			\end{minipage}%
		}
		\subfigure[]{
			\begin{minipage}[htp]{0.26\linewidth}
				\centering
				\includegraphics[width=1.5in]{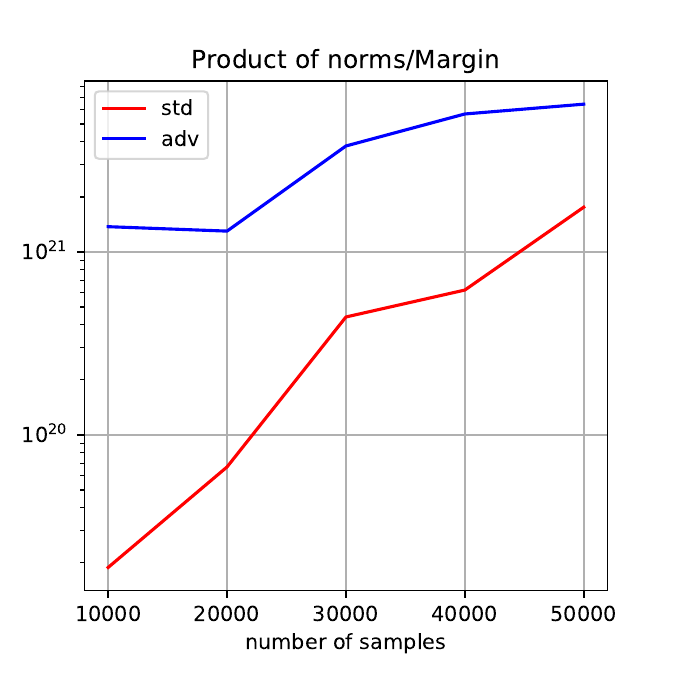}
			\end{minipage}
		}
		\subfigure[]{
			\begin{minipage}[htp]{0.26\linewidth}
				\centering
				\includegraphics[width=1.5in]{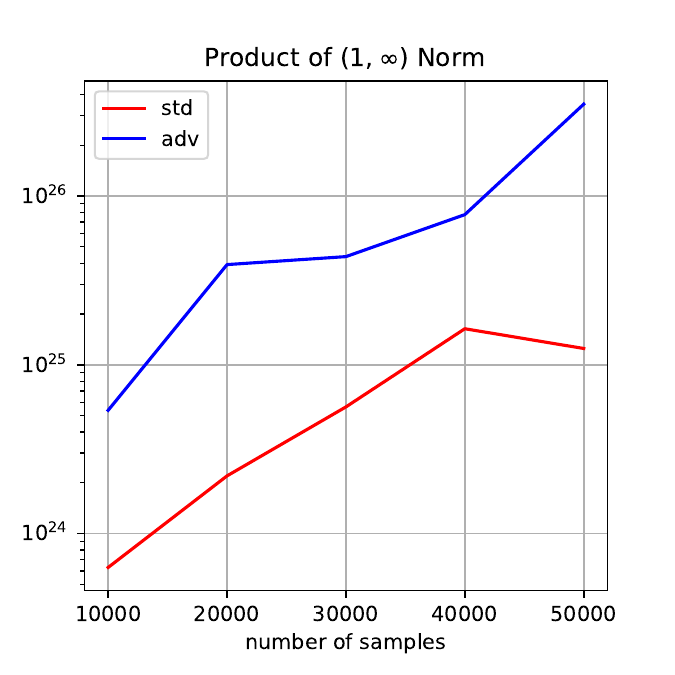}
			\end{minipage}
		}
		\subfigure[]{
			\begin{minipage}[htp]{0.26\linewidth}
				\centering
				\includegraphics[width=1.5in]{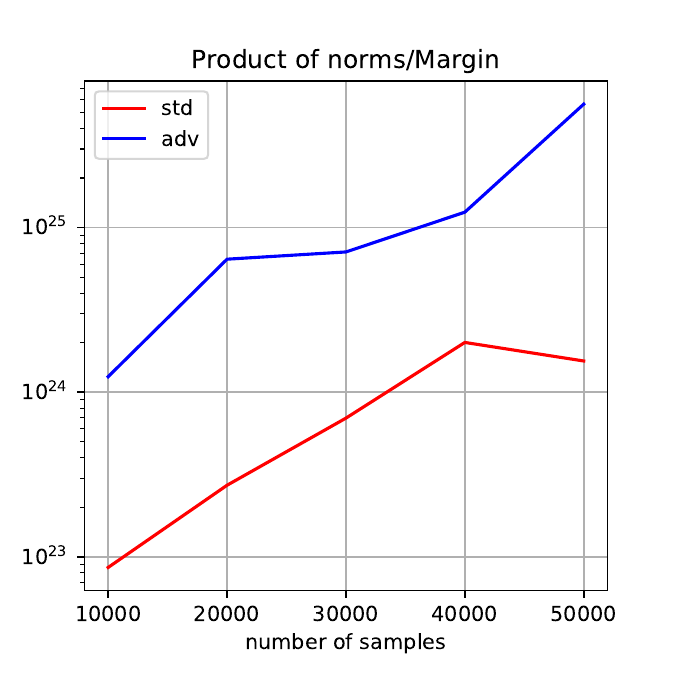}
			\end{minipage}
	}}
	
	\centering
	\vskip -0.15in
	\caption{Comparison of $\|\cdot\|_{1,\infty}$-norm products on CIFAR-10 between standard training (red lines) and adversarial training (blue lines) for: VGG-16 (a,b) and VGG-19 (c,d). The plots show: (a,c) Product of layer-wise $\|\cdot\|_{1,\infty}$-norms $\prod_{j=1}^l\|W_j\|_{1,\infty}$, and (b,d) Ratio of norm product to margin $\prod_{j=1}^l\|W_j\|_{1,\infty}/\gamma$.}
	\label{fig:linf}
	\vskip -0.1in
\end{figure}

\paragraph{$\|\cdot\|_{1,\infty}$-Norm Bounds.} The $\|\cdot\|_{1,\infty}$-norm bounds are shown in Figure \ref{fig:linf}. Similar the the Frobenius norm bounds,The gap of $\prod_{j=1}^l\|W_j\|_{1,\infty}$ between adversarial training and standard training are large. But the magnitude of $\prod_{j=1}^l\|W_j\|_{1,\infty}$ is larger than the magnitude of $\prod_{j=1}^l\|W_j\|_F$.

\subsection{Ablation Study on Margin Distribution}
\subsubsection{Discussion of optimizing over Gamma}
\label{optgamma}
By definition, these terms are
independent of the algorithm. However, they are implicitly algorithm-dependent 
\citep{bartlett2017spectrally,neyshabur2017pac}. The optimization over the margin parameter $\gamma$ is standard in empirical evaluations of
margin-based generalization bounds. For example, \citet{bartlett2017spectrally} derive a
spectrally-normalized margin bound based on Rademacher-complexity arguments and
empirically evaluate the bound as a function of the normalized margin. The experiment is reported in Figure 1 in their paper. Similarly,
\citet{neyshabur2017exploring} empirically compute and compare several generalization bounds,
including Rademacher-complexity-based norm bounds and spectrally-normalized margin bounds. Their
public implementation\footnote{\texttt{https://github.com/bneyshabur/generalization-bounds.}} reports the following bounds:
\begin{table}[t]
\centering
\small
\caption{Generalization bounds computed in \citet{neyshabur2017exploring}.}
\label{tab:gamma_optimization}
\begin{tabularx}{\textwidth}{l X l}
\toprule
\textbf{Bound} & \textbf{Reference} & \textbf{Type} \\
\midrule
$L_{1,\infty}$ bound
&
\citet{bartlett2002rademacher}, with depth dependence from \citet{golowich2018size}
&
Rademacher \\

$L_{3,1.5}$ bound
&
\citet{neyshabur2015norm}, with depth dependence from \citet{golowich2018size}
&
Rademacher \\

Frobenius bound
&
\citet{neyshabur2015norm}, with depth dependence from \citet{golowich2018size}
&
Rademacher \\

$\mathrm{Spec}_{L_1}$ bound
&
\citet{bartlett2017spectrally}
&
Rademacher margin \\

$\mathrm{Spec}_{\mathrm{Fro}}$ bound
&
\citet{neyshabur2017pac}
&
PAC-Bayesian margin \\
\bottomrule
\end{tabularx}
\end{table}

\subsubsection{Results}
\begin{figure}[htbp]
	
	\scalebox{0.9}{
			\begin{minipage}[htp]{0.01\linewidth}
			\centering
			\rotatebox{90}{\text{$1^{th}$-percentile}}
		\end{minipage}
		\subfigure[]{
			\begin{minipage}[htp]{0.26\linewidth}
				\centering
				\includegraphics[width=1.5in]{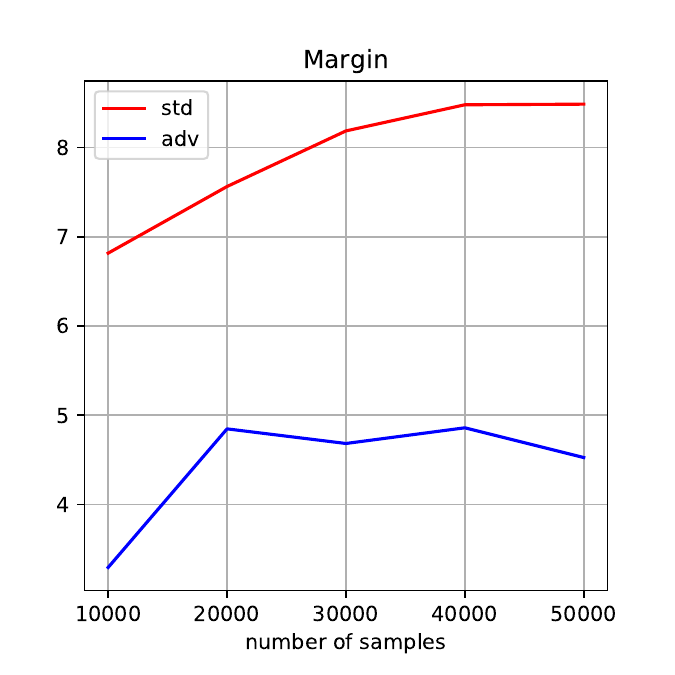}
			\end{minipage}%
		}
		\subfigure[]{
			\begin{minipage}[htp]{0.26\linewidth}
				\centering
				\includegraphics[width=1.5in]{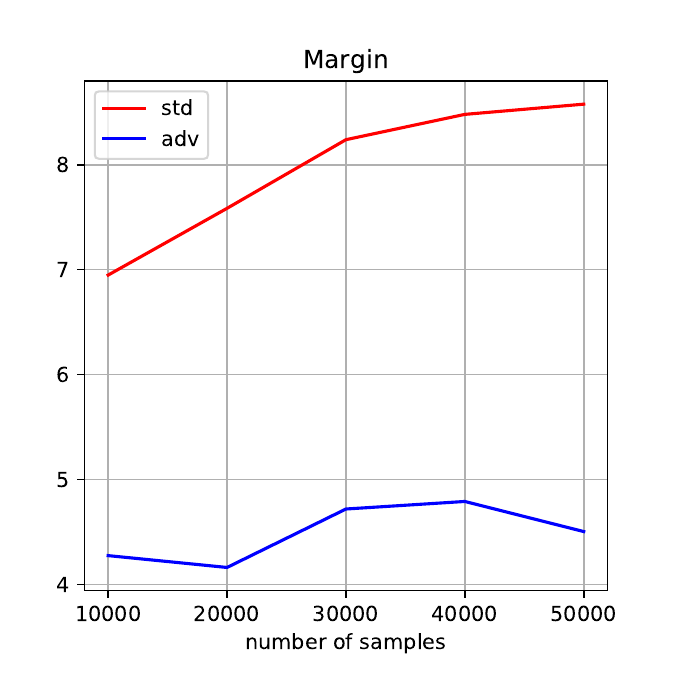}
			\end{minipage}
		}
		\subfigure[]{
			\begin{minipage}[htp]{0.26\linewidth}
				\centering
				\includegraphics[width=1.5in]{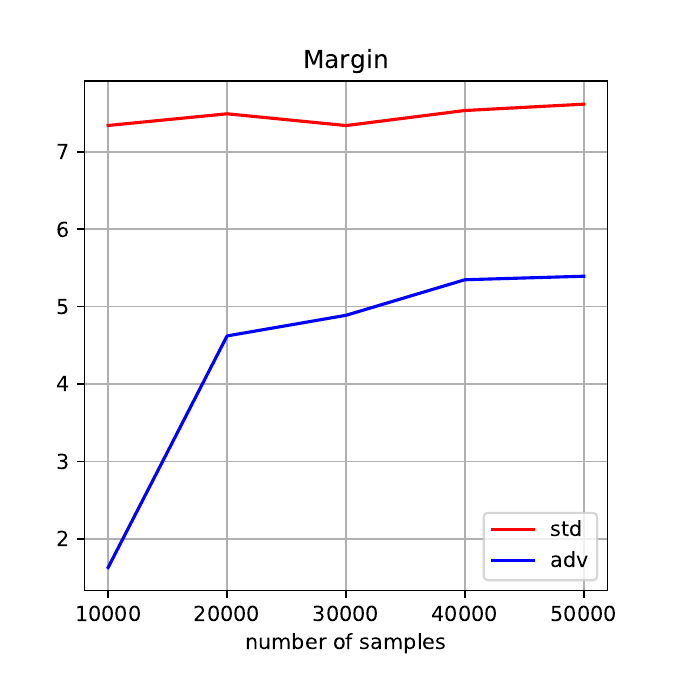}
			\end{minipage}
		}
		\subfigure[]{
			\begin{minipage}[htp]{0.26\linewidth}
				\centering
				\includegraphics[width=1.5in]{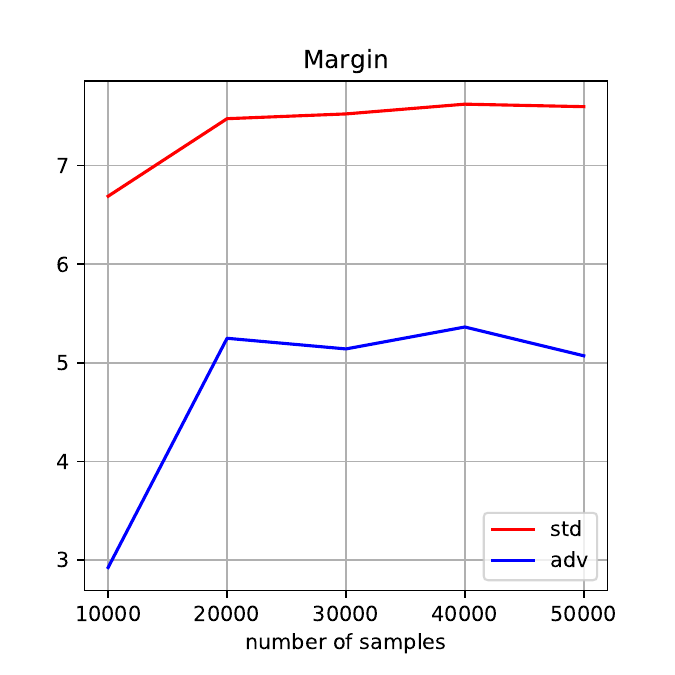}
			\end{minipage}
	}}
	
		\scalebox{0.9}{
			\begin{minipage}[htp]{0.01\linewidth}
			\centering
			\rotatebox{90}{\text{$3^{th}$-percentile}}
		\end{minipage}
		\subfigure[]{
			\begin{minipage}[htp]{0.26\linewidth}
				\centering
				\includegraphics[width=1.5in]{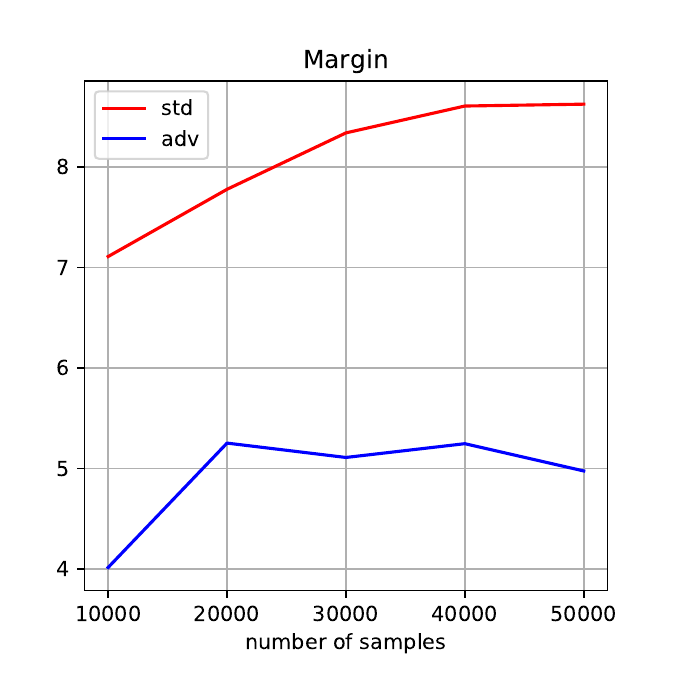}
			\end{minipage}%
		}
		\subfigure[]{
			\begin{minipage}[htp]{0.26\linewidth}
				\centering
				\includegraphics[width=1.5in]{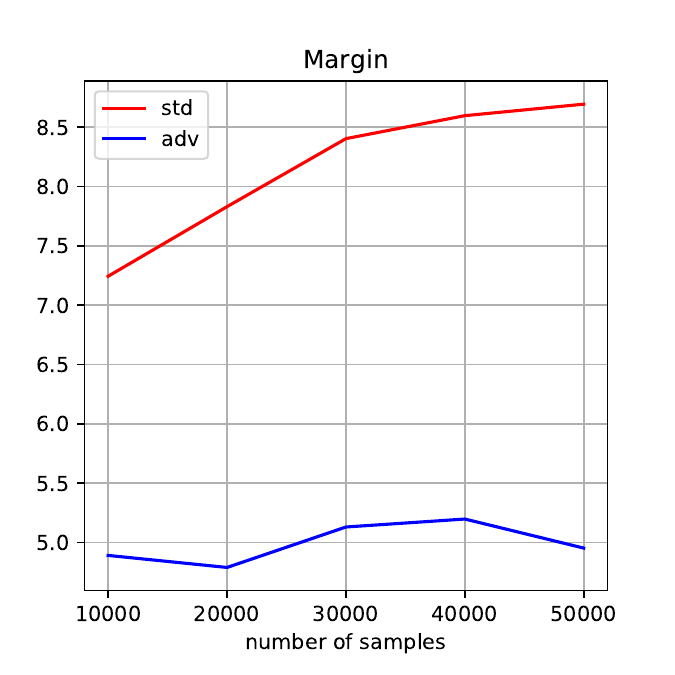}
			\end{minipage}
		}
		\subfigure[]{
			\begin{minipage}[htp]{0.26\linewidth}
				\centering
				\includegraphics[width=1.5in]{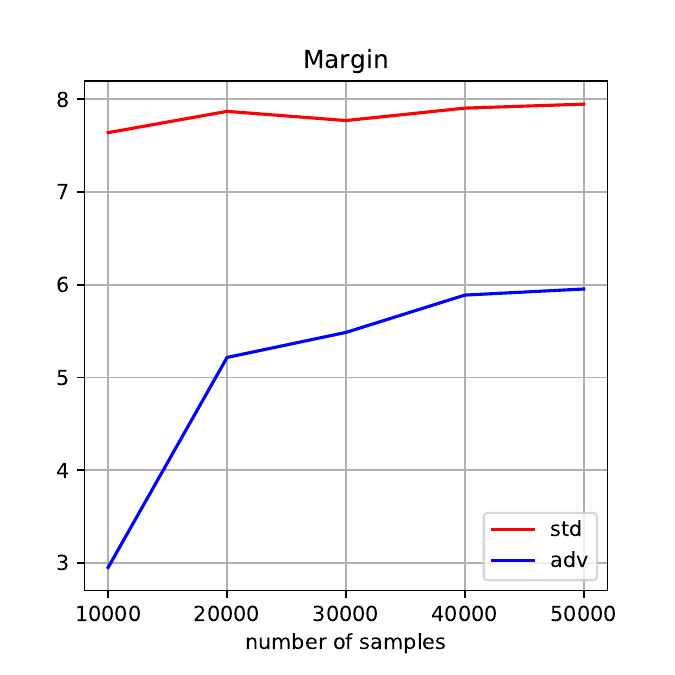}
			\end{minipage}
		}
		\subfigure[]{
			\begin{minipage}[htp]{0.26\linewidth}
				\centering
				\includegraphics[width=1.5in]{margin_vgg19_1.pdf}
			\end{minipage}
	}}
	
			\scalebox{0.9}{
			\begin{minipage}[htp]{0.01\linewidth}
			\centering
			\rotatebox{90}{\text{$5^{th}$-percentile}}
		\end{minipage}
		\subfigure[]{
			\begin{minipage}[htp]{0.26\linewidth}
				\centering
				\includegraphics[width=1.5in]{vgg11_3.pdf}
			\end{minipage}%
		}
		\subfigure[]{
			\begin{minipage}[htp]{0.26\linewidth}
				\centering
				\includegraphics[width=1.5in]{vgg13_3.pdf}
			\end{minipage}
		}
		\subfigure[]{
			\begin{minipage}[htp]{0.26\linewidth}
				\centering
				\includegraphics[width=1.5in]{vgg16_3.pdf}
			\end{minipage}
		}
		\subfigure[]{
			\begin{minipage}[htp]{0.26\linewidth}
				\centering
				\includegraphics[width=1.5in]{vgg19_3.pdf}
			\end{minipage}
	}}
	\centering
	\caption{Margin analysis across VGG architectures: Results from VGG-11 (first column), VGG-13 (second column), VGG-16 (third column), and VGG-19 (fourth column).}
	\label{fig:margins}
\end{figure}

Figure \ref{fig:margins} illustrates the margin distributions at the 1st, 3rd, and 5th percentiles of the training dataset. As the robust training accuracy reaches 100\%, the choice of percentile does not significantly impact our analysis. Across all percentiles, standard training consistently achieves larger margins compared to adversarial training. Since margins appear in the denominator of the Rademacher complexity upper bound, these smaller margins in adversarial training contribute, albeit modestly, to its poorer generalization performance.

\subsection{Experiments on CIFAR-100}
\label{C3}
\paragraph{Performance Analysis.} Table \ref{tab:my_jabel} presents comparative results between standard and adversarial training on CIFAR-100 using VGG-16 and VGG-19 architectures. Our experiments reveal that the standard CIFAR-100 training set size of 50,000 samples is insufficient to effectively train VGG networks to acceptable performance levels. This limitation makes it challenging to analyze weight norm trends through CIFAR-100 experiments. Nevertheless, we compare the product of weight norms between standard and adversarial training methods to gain insights into their relative behaviors.

\begin{figure}[htbp]

\scalebox{0.9}{
			\begin{minipage}[htp]{0.01\linewidth}
			\centering
			\rotatebox{90}{\text{VGG-16}}
		\end{minipage}
		\subfigure[]{
			\begin{minipage}[htp]{0.26\linewidth}
				\centering
				\includegraphics[width=1.5in]{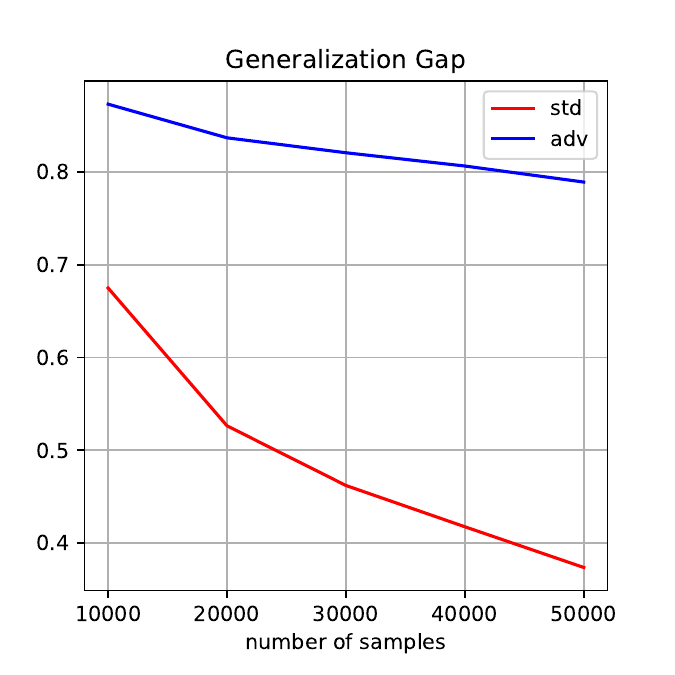}
			\end{minipage}%
		}
		\subfigure[]{
			\begin{minipage}[htp]{0.26\linewidth}
				\centering
				\includegraphics[width=1.5in]{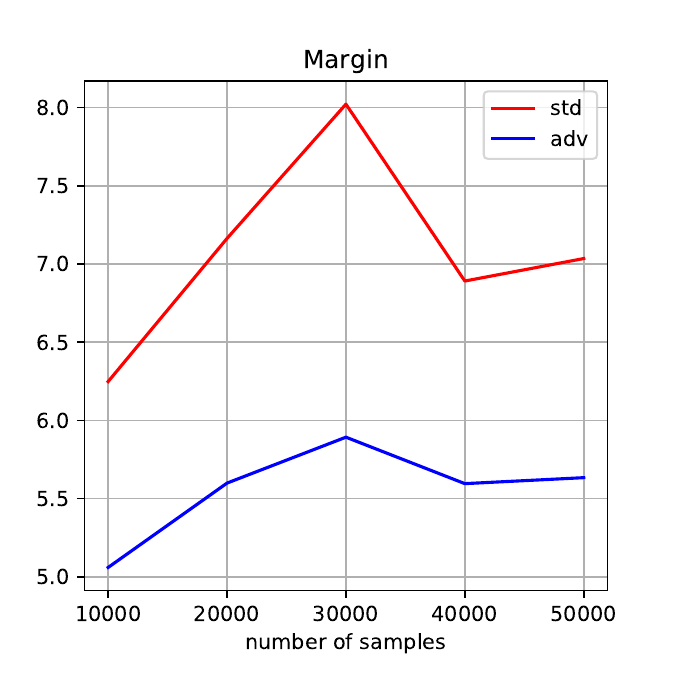}
			\end{minipage}
		}
		\subfigure[]{
			\begin{minipage}[htp]{0.26\linewidth}
				\centering
				\includegraphics[width=1.5in]{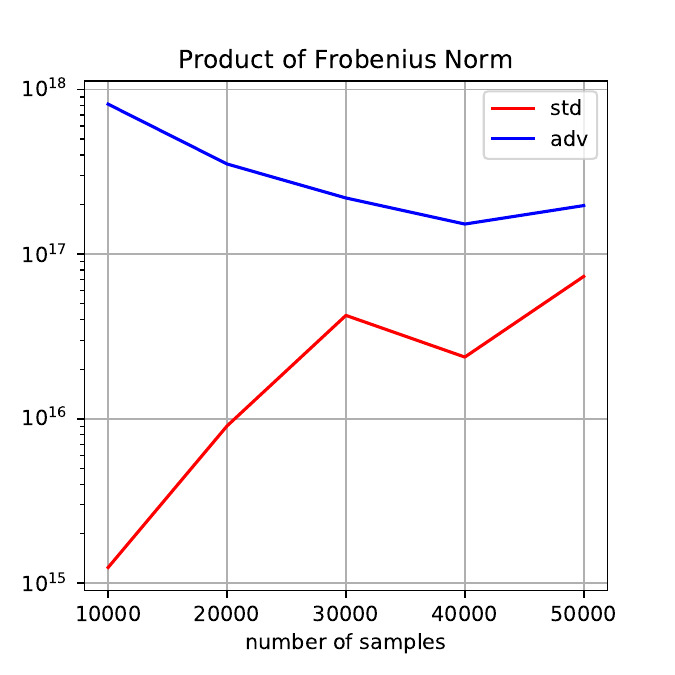}
			\end{minipage}
		}
		\subfigure[]{
			\begin{minipage}[htp]{0.26\linewidth}
				\centering
				\includegraphics[width=1.5in]{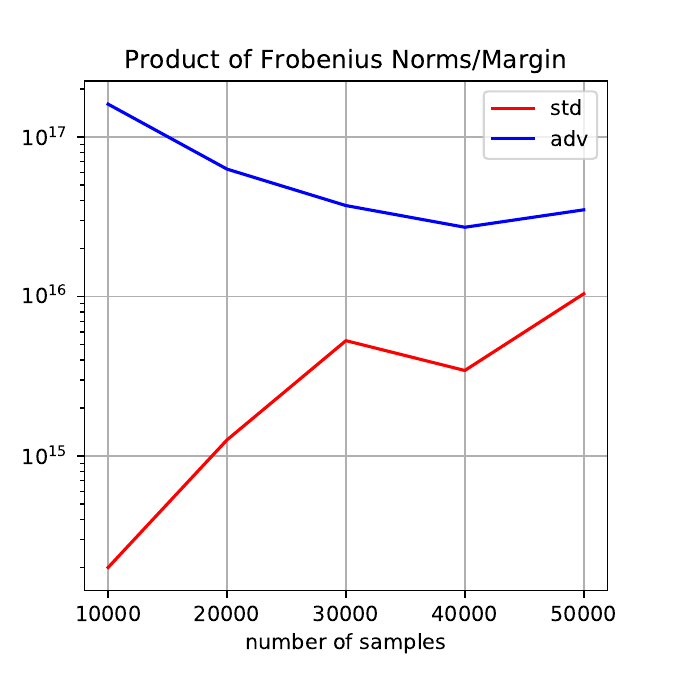}
			\end{minipage}
	}}
	
    \scalebox{0.9}{
			\begin{minipage}[htp]{0.01\linewidth}
			\centering
			\rotatebox{90}{\text{VGG-19}}
		\end{minipage}
		\subfigure[]{
			\begin{minipage}[htp]{0.26\linewidth}
				\centering
				\includegraphics[width=1.5in]{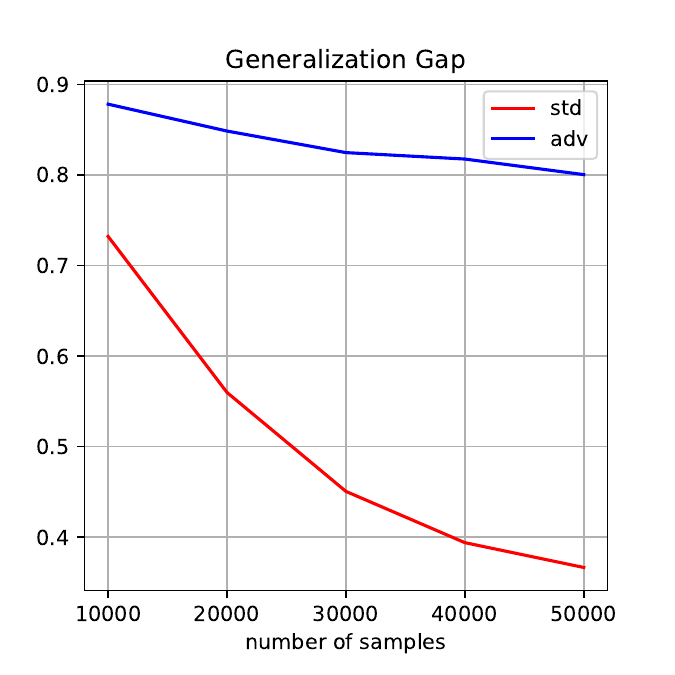}
			\end{minipage}%
		}
		\subfigure[]{
			\begin{minipage}[htp]{0.26\linewidth}
				\centering
				\includegraphics[width=1.5in]{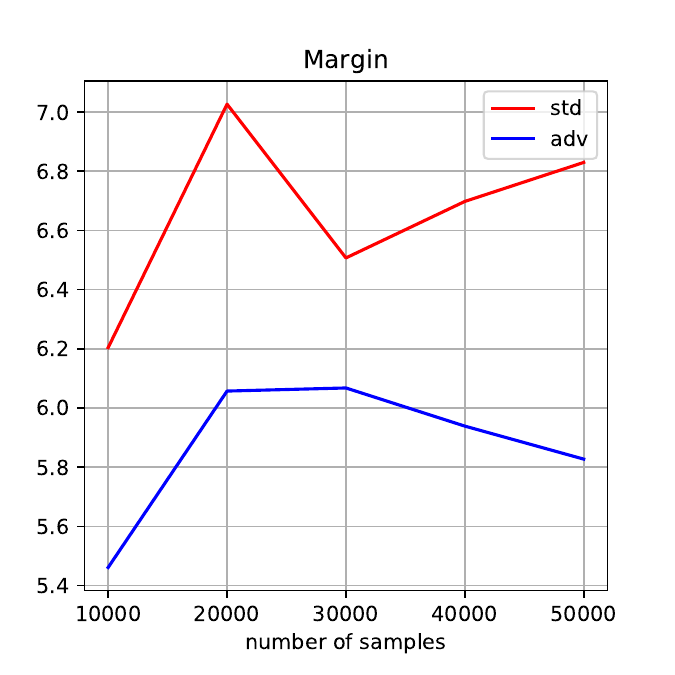}
			\end{minipage}
		}
		\subfigure[]{
			\begin{minipage}[htp]{0.26\linewidth}
				\centering
				\includegraphics[width=1.5in]{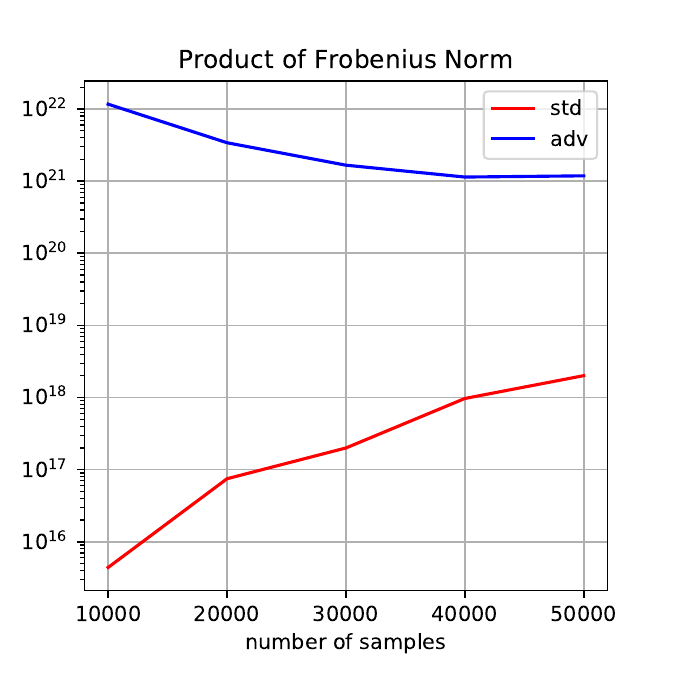}
			\end{minipage}
		}
		\subfigure[]{
			\begin{minipage}[htp]{0.26\linewidth}
				\centering
				\includegraphics[width=1.5in]{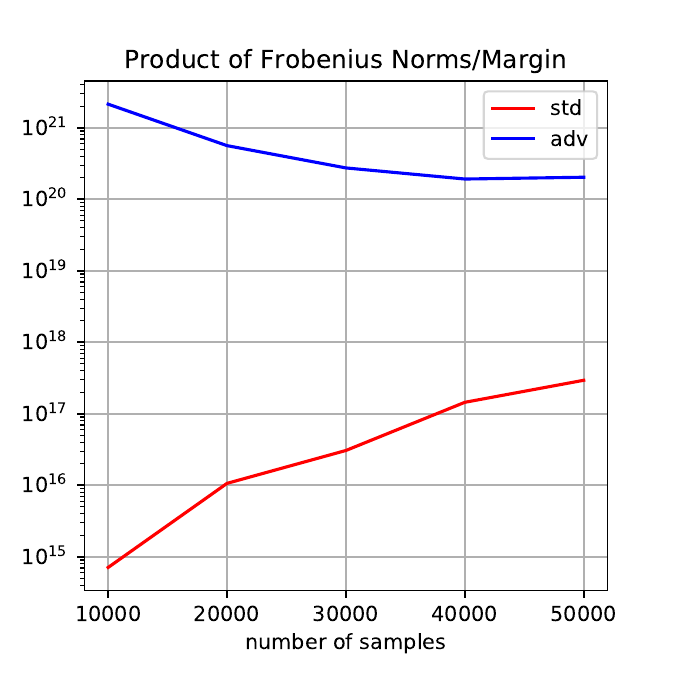}
			\end{minipage}
	}}
	\centering
	\caption{CIFAR-100 experimental results comparing standard training (red lines) and adversarial training (blue lines) for VGG-16 (top row) and VGG-19 (bottom row). The plots show: (a,e) Generalization gap, (b,f) Training set margin $\gamma$, (c,g) Product of layer-wise Frobenius norms $\prod_{j=1}^l\|W_j\|_F$, and (d,h) Ratio of Frobenius norm product to margin $\prod_{j=1}^l\|W_j\|_F/\gamma$.}
	\label{fig:cifar100}
\end{figure}

\paragraph{Product of Weight Norms.} Figure \ref{fig:cifar100} presents the training results for VGG-16 and VGG-19 architectures on CIFAR-100. Consistent with our CIFAR-10 experiments, adversarially trained models exhibit significantly larger weight norms compared to their standard-trained counterparts.

\begin{table}[ht]
        \caption{Performance comparison between standard and adversarial training on CIFAR-100 using VGG-16 and VGG-19 architectures. Results show clean accuracy for standard training and robust accuracy (against PGD attacks) for adversarial training.}
    \centering
    \begin{tabular}{cccccc}
    \toprule
    No. of Samples & 10000 &20000 &30000 &40000 &50000\\
    \midrule
VGG-16-STD &0.26&     0.44& 0.54&     0.60&      0.63\\
VGG-16-ADV &0.12 &0.15 &0.17   &  0.18 &0.19\\
VGG-19-STD& 0.32& 0.47&    0.53   &  0.58  &   0.62  \\
VGG-19-ADV&0.12& 0.16 &0.17 &0.19    & 0.21\\
\bottomrule
    \end{tabular}

    \label{tab:my_jabel}
\end{table}

\begin{figure}[htbp]
	
\scalebox{0.9}{

		\subfigure[]{
			\begin{minipage}[htp]{0.26\linewidth}
				\centering
				\includegraphics[width=1.5in]{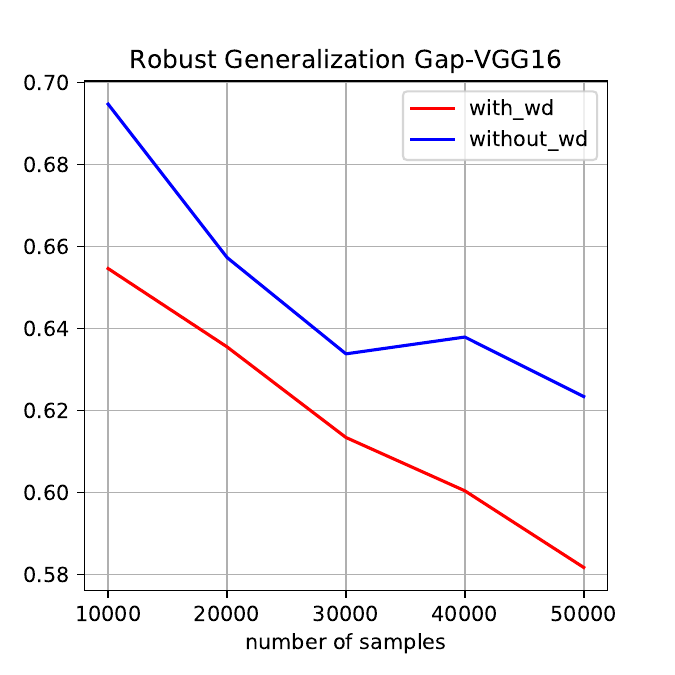}
			\end{minipage}%
		}
		\subfigure[]{
			\begin{minipage}[htp]{0.26\linewidth}
				\centering
				\includegraphics[width=1.5in]{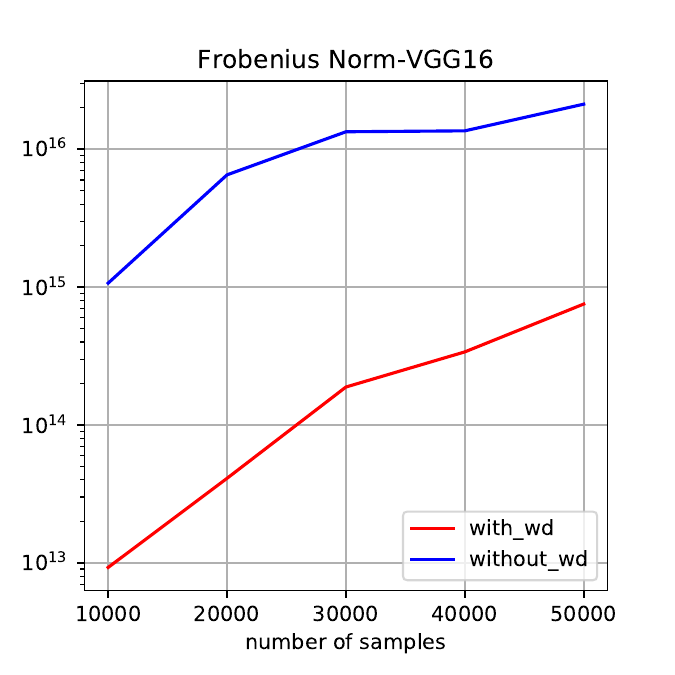}
			\end{minipage}
		}
		\subfigure[]{
			\begin{minipage}[htp]{0.26\linewidth}
				\centering
				\includegraphics[width=1.5in]{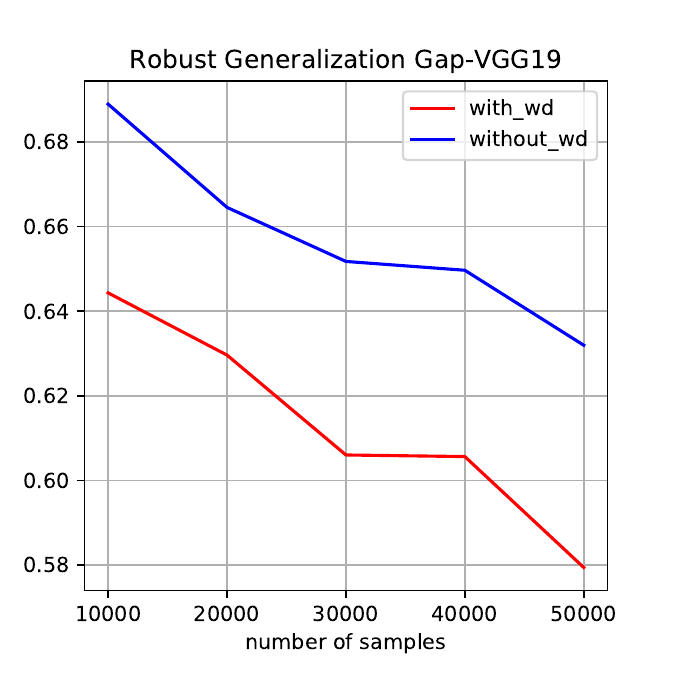}
			\end{minipage}
		}
		\subfigure[]{
			\begin{minipage}[htp]{0.26\linewidth}
				\centering
				\includegraphics[width=1.5in]{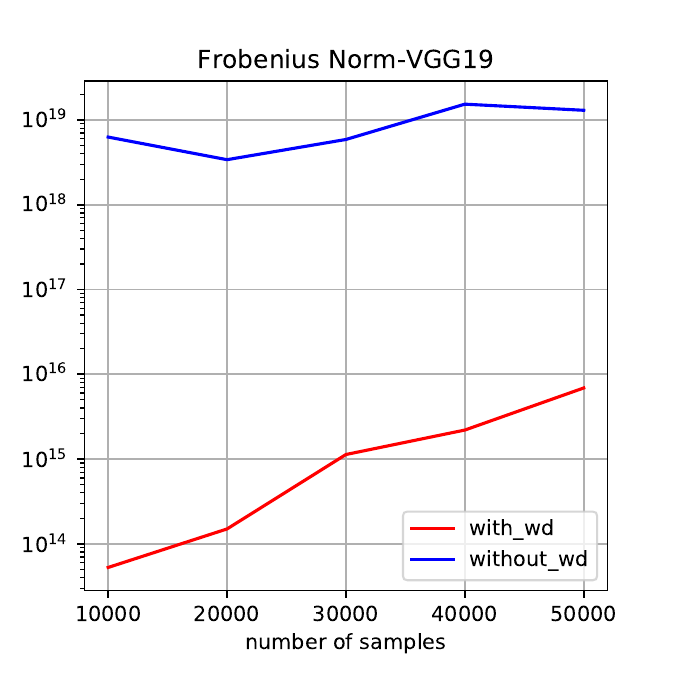}
			\end{minipage}
	}}
	
	\centering
	\vskip -0.15in
	\caption{Impact of weight decay on VGG architectures: Results for VGG-16 (a,b) and VGG-19 (c,d), comparing models trained with and without weight decay. The plots show: (a,c) Robust generalization gap and (b,d) Product of layer-wise Frobenius norms.}
	\label{fig:weightdecay}
	\vskip -0.1in
\end{figure}

\begin{figure}[htbp]	
\scalebox{0.9}{
		\subfigure[]{
			\begin{minipage}[htp]{0.32\linewidth}
				\centering
				\includegraphics[width=1.5in]{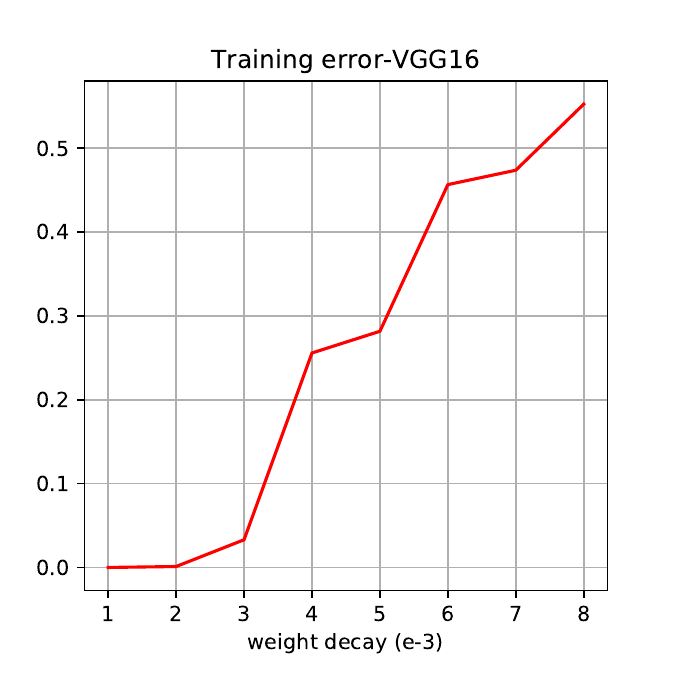}
			\end{minipage}
		}
		\subfigure[]{
			\begin{minipage}[htp]{0.32\linewidth}
				\centering
				\includegraphics[width=1.5in]{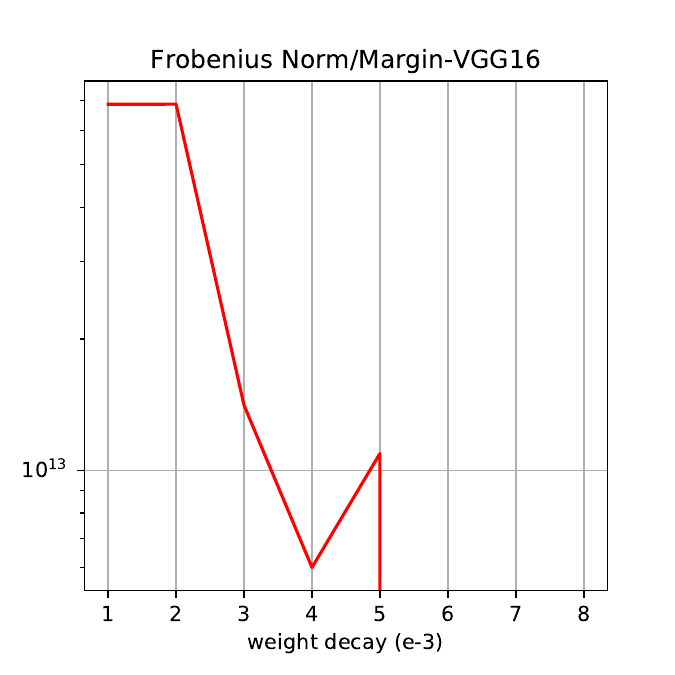}
			\end{minipage}
		}
		\subfigure[]{
			\begin{minipage}[htp]{0.32\linewidth}
				\centering
				\includegraphics[width=1.5in]{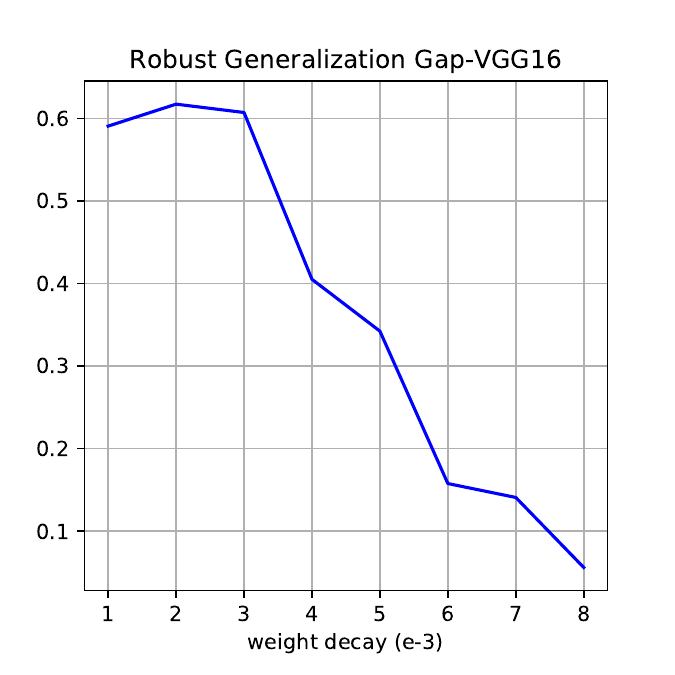}
			\end{minipage}
	}}
	
	\centering
	\vskip -0.15in
	\caption{Weight decay effects on model behavior for values ranging from $1\times 10^{-3}$ to $9\times 10^{-3}$: (a) Training error performance, (b) Frobenius norm of model weights, and (c) Robust generalization gap between training and test performance.}
	\label{fig:weightdecay2}
	\vskip -0.1in
\end{figure}

\subsection{Weight Decay}
\label{c4}
Theoretical upper bounds on adversarial Rademacher complexity suggest that adding weight regularization (weight decay) can improve generalization performance. We experimentally validate this theoretical insight in Figure \ref{fig:weightdecay}, comparing adversarial training with and without weight decay. The results in Figures \ref{fig:weightdecay}(a) and (c) demonstrate that incorporating weight decay reduces the robust generalization gap. Additionally, Figures \ref{fig:weightdecay}(b) and (d) show that adversarial training with weight decay yields smaller weight norm products. These experimental findings establish a clear empirical connection between the robust generalization gap and weight norm products, supporting our theoretical analysis.

In Figure \ref{fig:weightdecay2}, we examine the effect of weight decay values ranging from $1\times 10^{-3}$ to $9\times 10^{-3}$. The training error increases with higher weight decay values. At low weight decay (\(1 \times 10^{-3}\)), the model achieves minimal training error, indicating a high capacity to fit the training data. However, as weight decay increases, the model's flexibility is constrained, leading to higher training errors. This behavior aligns with the regularization effect of weight decay, which penalizes large weights and reduces the model's ability to overfit. At very high weight decay values (\(9 \times 10^{-3}\)), the model risks underfitting, as excessive regularization prevents it from capturing meaningful patterns in the data.  

The Frobenius norm of the model weights decreases monotonically with increasing weight decay. This trend reflects the direct impact of weight decay on the optimization process, where larger decay values impose stricter penalties on weight magnitudes. The reduction in the Frobenius norm indicates a simplification of the model, which is a key objective of regularization. However, excessively small weight magnitudes can lead to underfitting, highlighting the need for careful tuning of the weight decay parameter.  

The generalization gap, defined as the difference between training and test performance, demonstrates a non-linear relationship with weight decay. At low weight decay values, the gap is large, indicating poor generalization due to overfitting. As weight decay increases, the gap narrows, reaching a minimum at intermediate values (e.g., \(3 \times 10^{-3}\) to \(7 \times 10^{-3}\)). This reduction in the gap signifies improved generalization, as the model achieves a better balance between fitting the training data and maintaining performance on unseen data. At very high weight decay values, the gap may stabilize or slightly increase, as both training and test performance degrade due to underfitting.  

In conclusion, the results highlight the trade-offs associated with weight decay. Model performance deteriorates significantly within this range. At weight decay values of $6\times 10^{-3}$ and higher, the margin becomes negative, indicating high training error. Subsequently, the weight norm to margin ratio also becomes negative. At $9\times 10^{-3}$, the model fails to learn entirely, with both training and test errors reaching 90\%. Our analysis suggests the optimal weight decay range for minimizing weight norm lies between $1\times 10^{-3}$ and $5\times 10^{-3}$. The smallest weight norm observed was $6.01\times 10^{12}$ (with weight decay = $4\times 10^{-3}$). However, this value remains larger than the weight norm achieved through standard training ($1.90\times 10^{12}$). These findings emphasize the importance of selecting an appropriate weight decay value to achieve robust model performance.

\end{document}